\documentclass[10pt,twocolumn,letterpaper]{article}

\usepackage{cvpr}              %

\usepackage[a4paper,margin=1in]{geometry} %
\usepackage[table]{xcolor} %
\usepackage{graphicx} %
\usepackage{longtable} %
\usepackage{booktabs} %
\usepackage{adjustbox} %
\usepackage{makecell} %
\usepackage{pifont}%
\newcommand{\cmark}{\ding{51}}%
\newcommand{\xmark}{\ding{55}}%
\usepackage{float}
\usepackage{caption}
\usepackage{multirow}

\usepackage{fontawesome5} %

\definecolor{terramindblue}{HTML}{337FFE}
\definecolor{copgengreen}{HTML}{0DC657}
\definecolor{water}{HTML}{1a5bab}            %
\definecolor{trees}{HTML}{358221}            %
\definecolor{floodedVegetation}{HTML}{87d19e} %
\definecolor{crops}{HTML}{ffdb5c}           %
\definecolor{builtUp}{HTML}{fc5d79}         %
\definecolor{bareGround}{HTML}{e3e2c3}      %
\definecolor{snowIce}{HTML}{a8ebff}         %
\definecolor{clouds}{HTML}{616161}          %
\definecolor{rangeland}{HTML}{a59b8f}       %

\definecolor{cvprblue}{rgb}{0.21,0.49,0.74}
\usepackage[pagebackref,breaklinks,colorlinks,allcolors=cvprblue]{hyperref}

\def\paperID{*****} %
\def\confName{CVPR}
\def\confYear{2026}

\title{COP-GEN: Latent Diffusion Transformer\\for Copernicus Earth Observation Data}

\author{
Miguel Espinosa\\
School of Engineering\\
University of Edinburgh \\
{\tt\small miguel.espinosa@ed.ac.uk}
\and
Eva Gmelich Meijling \\
European Space Agency (ESA) \\
{\tt\small Eva.GmelichMeijling@esa.int}
\and
Valerio Marsocci \\
European Space Agency (ESA) \\
{\tt\small valerio.marsocci@esa.int}
\and
Elliot J. Crowley \\
School of Engineering\\
University of Edinburgh \\
{\tt\small  elliot.j.crowley@ed.ac.uk}
\and
Mikolaj Czerkawski\\
Asterisk Labs\\
{\tt\small miko@asterisk.coop}
}

\begin{document}
\maketitle

{\begin{figure*}
    \centering
    \includegraphics[width=\linewidth]{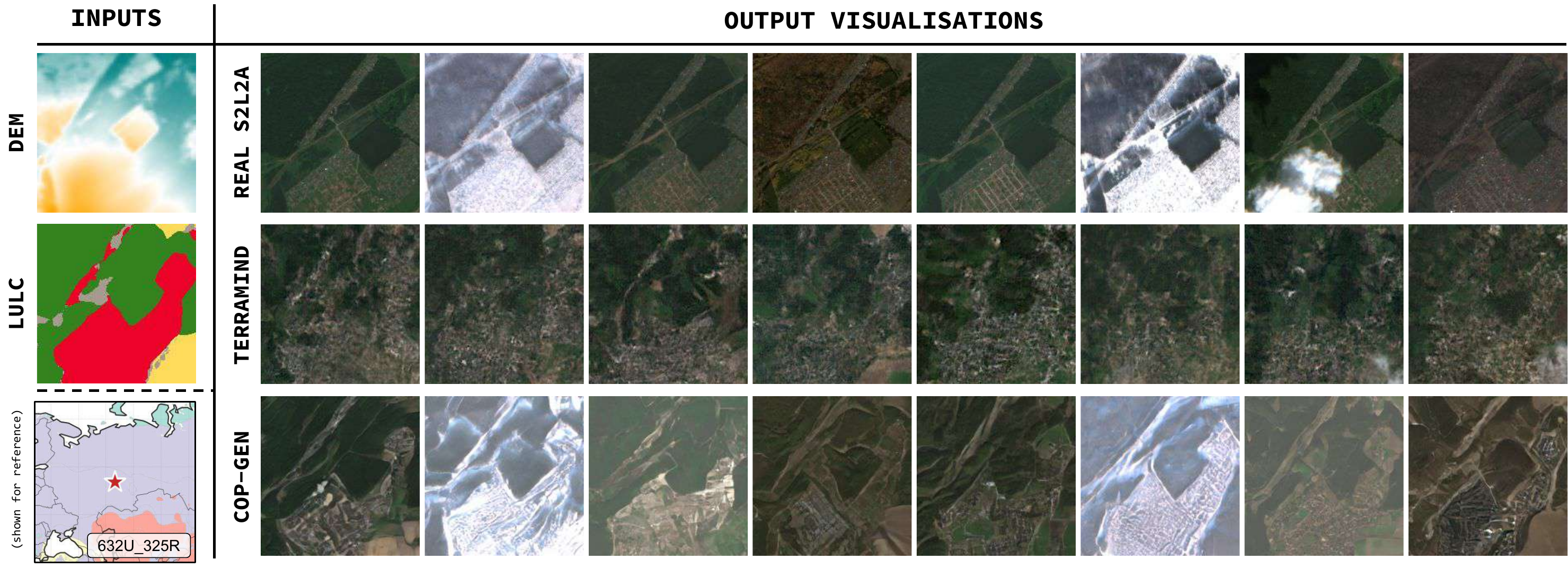}
    \caption{\textbf{Conditional generation of Sentinel-2 L2A imagery from topography (DEM) and land-cover (LULC) inputs.} We condition COP-GEN generations on DEM and LULC inputs (geographic location is provided solely for visual reference). COP-GEN produces diverse and physically consistent outputs, demonstrating variability in spectral appearance, illumination, and atmospheric conditions while preserving topographic and land-cover constraints.
    This highlights the model’s ability to capture the inherent one-to-many relationships of multimodal Earth Observation data.
    LULC classes are visualized using the following color scheme: 
    {\color{water}\textbf{Water}}, 
    {\color{trees}\textbf{Trees}}, 
    {\color{floodedVegetation}\textbf{Flooded vegetation}}, 
    {\color{crops}\textbf{Crops}}, 
    {\color{builtUp}\textbf{Built-up areas}}, 
    {\color{bareGround}\textbf{Bare ground}}, 
    {\color{snowIce}\textbf{Snow/ice}}, 
    {\color{clouds}\textbf{Clouds}}, and 
    {\color{rangeland}\textbf{Rangeland}}. 
    Additional qualitative results and visualisations are provided in the Supplementary Material.
    }
    \label{fig:teaser}
\end{figure*}
}

\begin{abstract}
    Earth observation applications increasingly rely on data from multiple sensors, including optical, radar, elevation, and land-cover products. Relationships between these modalities are fundamental for data integration but are inherently non-injective: identical conditioning information can correspond to multiple physically plausible observations, and should therefore be parametrised as conditional distributions. Deterministic models, by contrast, collapse toward conditional means and fail to represent the uncertainty and variability required for tasks such as data completion and cross-sensor translation.
    We introduce COP-GEN, a multimodal latent diffusion transformer that models the joint distribution of heterogeneous Earth Observation modalities at their native spatial resolutions. By parameterising cross-modal mappings as conditional distributions, COP-GEN enables flexible any-to-any conditional generation, including zero-shot modality translation, spectral band infilling, and generation under partial or missing inputs, without task-specific retraining.
    Experiments on a large-scale global multimodal dataset show that COP-GEN generates diverse yet physically consistent realisations while maintaining strong peak fidelity across optical, radar, and elevation modalities. Qualitative and quantitative analyses demonstrate that the model captures meaningful cross-modal structure and systematically adapts its output uncertainty as conditioning information increases.
    We further release a stochastic benchmark built from multi-temporal Sentinel-2 observations that enables distribution-level comparison of generative EO models. On this benchmark, COP-GEN covers 90\% of the real observation manifold and 63\% of its per-band reflectance range, while the strongest competing method collapses to 2.8\% and 18\%, respectively.
    These results highlight the practical importance of stochastic generative modeling for Earth observation and motivate evaluation protocols beyond single-reference, pointwise metrics.
    {\small \texttt{Website:} 
    \url{https://miquel-espinosa.github.io/cop-gen}}
\end{abstract}

\section{Introduction}
\label{sec:intro}

    \textit{``There is nothing permanent except change''} is an oft-repeated quote, but it does describe the state of our home planet, Earth, rather faithfully. In the field of Earth observation, the archives are filled with millions of observations, acquired at different times, locations, and by a heterogenous fleet of instruments spanning across decades, and yet, describing parts of the same underlying physical system.
    COP-GEN is one of many generative models that attempt to model the relationships between multiple Earth observation (EO) data sources (such as optical multi-spectral sensors or synthetic aperture radar), but it is the only model that successfully combines this many modalities into a single stochastic model. The work presented herein is as much about the architecture (and how it was designed and trained) as it is about how generative Earth observation models can be evaluated and understood.

    The name COP-GEN naturally echoes the name of the Copernicus Programme~\cite{copernicus_programme}, which represents one of the world’s largest and most influential initiatives for open-access EO, providing continuous global coverage through a diverse constellation of sensors and modalities (optical, radar, elevation, and more). Each year, Copernicus produces petabytes of freely available data, forming a rapidly expanding record of the Earth’s surface and atmosphere. This massive data influx has enabled transformative progress across domains such as climate monitoring, disaster response, and environmental management. However, the scale, heterogeneity, and complexity of this data also pose significant challenges in effectively integrating and interpreting multimodal observations at global scale. Generative models can fill this gap by integrating these diverse data sources into an interoperable framework that can flexibly map between the modalities even if their spatiotemporal contexts are disparate. Recent research~\cite{jia2025can} has shown that diffusion models in particular implicitly learn the statistics of the data distribution through their denoising objective. This acts as a powerful form of self-supervision and encourages the model to develop powerful representations of the input data, capturing highly abstract dependencies and supporting flexible conditional sampling.

    However, the key challenge in desigining generative models of EO data lies in the nature of cross-modal mappings themselves. Relationships between the modalities are often non-injective: a given set of conditioning variables, such as terrain elevation or land-cover class, can correspond to many physically plausible realisations of optical appearance, radar backscatter, or atmospheric conditions. Such mappings are inherently one-to-many, as shown in Figure~\ref{fig:teaser}. Therefore, modelling them with deterministic functions encourages models to inevitably regress toward conditional means, rather than capturing the full range of valid outcomes.

    Recent work in EO has been dominated by discriminative foundation models~\cite{2025_tessera,wang2025unifiedcopernicusfoundationmodel,waldmann2025panopticon,brown2025alphaearthfoundationsembeddingfield,xiong2024dofa,OlmoEarthpretrain,nedungadi2024mmearth,astruc2024omnisat,szwarcman2025prithvieo20versatilemultitemporalfoundation,tseng2025galileo,danish2025terrafmscalablefoundationmodel}
    that learn representations via self-supervised pretraining, oftentimes via masked autoencoding. While potentially effective for downstream classification or regression, these models typically inherit architectural assumptions from natural-image processing and lack mechanisms to explicitly represent multimodal variability. Consquently, cross-modal relationships that encode meaningful geophysical structure are difficult to model with deterministic masked autoencoding. This observation is also consistent with the recent work on the trade-offs between joint embeddings versus reconstruction targets in representation learning~\cite{assel2025jointembedding}.

    Stochastic generative modeling provides an alternative route. By learning joint probability distributions across sensors, generative models can estimate multiple physically plausible realisations of a scene conditioned on a subset of input modalities. This approach aligns with remote sensing practice, where environmental processes are dynamic, observations are most often incomplete, and many different outputs are valid.

    Despite their potential, unified multimodal generative architectures for EO remains an open problem. Existing approaches are limited in scope, as either restricted to few modalities~\cite{khanna2023diffusionsat,metaearth}, operate on reduced-resolution imagery~\cite{metaearth}, or constrain generation to fixed conditional settings~\cite{tang2024crs}.
    There is currently no multimodal generative framework capable of handling native-resolution data across diverse Copernicus modalities while supporting any-to-any conditional generation and producing diverse outputs that reflect distributions of possible scenarios.

    We address this gap with COP-GEN, a scalable multimodal latent diffusion transformer designed to model the joint distribution of Copernicus sensor data, including optical, radar, elevation, land cover, timestamp, and geolocation. We release the COP-GEN model and its implementation as open-source. COP-GEN encodes each modality into a unified sequence of latent tokens with modality-specific diffusion timesteps. This design allows COP-GEN to:
    \begin{itemize}
        \item integrate heterogeneous data sources at their native resolutions without agressive resampling,
        \item support flexible conditional and unconditional sampling within a single architecture, generating multiple plausible realisations of a scene, 
        \item capture natural variance of data in underdetermined settings (ill-posed inverse problems),
        \item perform zero-shot modality translation without task-specific retraining (generating any missing modality from any available subset)
    \end{itemize}

    In addition to COP-GEN architecture, we also highlight the importance of evaluation methodologies for generative EO. Traditional metrics based on single-reference comparisons are poorly aligned with stochastic modeling and risk penalising models that capture output variability. We complement quantitative metrics with qualitative assessments that reveal distributional properties, such as output variance, spatial plausibility, and spectral consistency.

{\begin{table*}[t]
    \centering
    \resizebox{\textwidth}{!}{%
    \begin{tabular}{ c l  l c l }
        \toprule
         \textbf{Year} & \textbf{Name} & \textbf{Modalities} & \textbf{Stochastic} & \textbf{Application} \\
         \midrule
         2023 & GYOS & OSM, ArcGIS World Imagery & \cmark & Map-to-Optical \\
         \midrule
         2023 & DiffusionSat & \begin{tabular}{l} 
            S2, High-res RGB, lat-lon, \\ 
            GSD, cloud cover, timestamp, text \end{tabular} & \cmark & General-purpose \\
         \midrule
         2025 & MESA & S2L2A, COP-DEM-30, text & \cmark & Text-to-Terrain \\
         \midrule
         \textbf{2025} & \textbf{COP-GEN-Beta} & \begin{tabular}{l}
              S2L2A (thumbnails), S2L1C (thumbnails) \\
              S1RTC (thumbnails), COP-DEM-30 (thumbnails)
         \end{tabular} & \cmark & General-purpose \\
         \midrule
         2025 & TerraMind & \begin{tabular}{l}
            S2L2A, S2RGB, S2L1C, S1RTC, S1GRD, NDVI, COP-DEM-30 \\
              LULC, lat-lon, image caption
         \end{tabular} & \xmark & General-purpose \\
         \midrule
         \textbf{2025} & \textbf{COP-GEN} & \begin{tabular}{l}
              S2L2A, S2L1C, S1RTC, COP-DEM-30 \\
              LULC, timestamp, lat-lon
         \end{tabular} & \cmark & General-purpose \\
         \bottomrule
    \end{tabular}
    }
    \caption{\textbf{Overview of recent generative models for medium-resolution (10\,m) Earth observation data.}
    The table compares representative generative and multimodal models by year, supported input modalities, whether they explicitly model stochasticity, and their primary application scope. 
    COP-GEN extends prior work by enabling stochastic generation across multiple Copernicus sensors within a unified general-purpose framework.
    }    
    \label{tab:sota}
\end{table*}
}

\section{Related Work}
\label{sec:rel_work}

\subsection{Generative Models and Diffusion Transformers}

Generative modeling has undergone rapid development over the past decade, progressing from early generative adversarial networks (GANs) \cite{goodfellow2014gan, karras2019stylegan2} to autoregressive models \cite{ramesh2021dalle, esser2021taming} and, more recently, diffusion models \cite{sohl2015deep, ho2020ddpm, dhariwal2021diffusion}. Diffusion models have become a dominant paradigm for high fidelity image generation due to their training stability, strong sample quality and scalability with data and compute. The vast majority of diffusion architectures rely on UNet-style backbones \cite{ronneberger2015unet}, adopted in powerful models such as DDPM \cite{ho2020ddpm}, Stable Diffusion \cite{rombach2022ldm} and Imagen \cite{saharia2022imagen}.

Despite their success, UNet-based architectures present several structural limitations. Their convolutional hierarchies typically tie them to fixed spatial and spectral (i.e. number of channels) resolutions, limiting their applicability to datasets with heterogeneous image sizes. Moreover, UNets encode each image within a monolithic hierarchical tensor representation, making it difficult to incorporate additional modalities or sensor streams without redesigning the entire backbone. Their ability to model long-range or cross-modal dependencies is inherently constrained, as attention mechanisms are typically inserted only into a subset of blocks and do not operate on a unified token space.

To address these limitations, recent work has shifted toward transformer-based diffusion architectures. Models such as DiT \cite{peebles2023dit}, UViT \cite{bao2023uvit}, MDT \cite{gao2023masked}, and UniDiffuser \cite{bao2023one} treat image patches as latent tokens and employ pure transformer backbones to model the denoising process. This formulation brings several advantages: transformers scale efficiently with increased depth and width, can integrate arbitrary streams of tokens corresponding to heterogeneous data sources, and remove the need for fixed spatial layouts through patchification. Furthermore, transformers naturally enable any-to-any conditioning via cross-attention, allowing a single architecture to flexibly generate from or condition on different combinations of modalities. These properties make transformer-based diffusion architectures particularly suitable for domains that require modeling the joint distribution of diverse and physically distinct data sources, such as Earth Observation.

\subsection{Generative Models for Earth Observation}

Generative modeling in Earth Observation (EO) has historically followed developments in natural-image computer vision, with early work focusing on task-specific applications such as cloud removal, pansharpening, and cross-modal translation. GAN-based approaches have been widely explored for cloud removal \cite{meraner2020cloud, jin2024hya}, super-resolution \cite{wang2018esrgan, dong2020remote}, pansharpening \cite{ma2020pan}, SAR-to-optical translation \cite{wang2019sar}, synthetic DEM generation \cite{liu2022generative}, and atmospheric correction \cite{paola2023correction}. While effective for narrow tasks, GANs often struggle with mode collapse and do not provide a principled framework for modeling the joint distribution of multimodal EO data.

In parallel, diffusion models have gained increased traction in RS, being applied to image generation~\cite{espinosa_2023_8_mapsat}, enhancement, interpretation~\cite{liu2024diffusion}, and terrain generation~\cite{mesa2025}.
Recent efforts~\cite{khanna2023diffusionsat,metaearth,zheng2024changen2,tang2024crs,toker2024satsynth,espinosa2025cop} have focused on developing diffusion-based generative foundation models for high-fidelity satellite image synthesis (Table~\ref{tab:sota}). For instance, DiffusionSat~\cite{khanna2023diffusionsat} generates data conditioned on semantic text and metadata, while MetaEarth~\cite{metaearth} enables arbitrary-sized image generation using a resolution-guided approach. CRS-Diff\cite{tang2024crs} further enhances scene generation controllability by incorporating multiple conditioning inputs. Beyond synthesis, numerous diffusion-based methods address image enhancement tasks, including denoising~\cite{he2023tdiffde, pang2024hir}, cloud removal~\cite{jing2023denoising, wang2024idf, zou2024diffcr}, and super-resolution~\cite{wang2025semantic, dong2024building, an2023efficient}, showcasing their versatility in RS.

Another line of research focuses on diffusion models for discriminative applications \cite{le2024detecting, jia2025can, liu2024diffusion}, though these often rely on labeled data and are limited to specific tasks, such as semantic segmentation~\cite{amit2021segdiff, kolbeinsson2024multi, zhou2024exploring, li2024mdfl, qu2024lds2ae} or change detection~\cite{wen2024gcd, tian2024swimdiff, zhang2023diffucd, jia2024siamese}. For example, SegDiff~\cite{amit2021segdiff} diffuse ground-truth masks, while others use class predictions or labeled guidance~\cite{kolbeinsson2024multi,li2024mdfl,qu2024lds2ae}. Although a few studies have explored diffusion as a label-free pretraining framework, they remain narrowly focused on a single application scenario, such as hyperspectral images segmentation~\cite{zhou2024exploring,sigger2024unveiling}, or change detection~\cite{bandara2022ddpm}. SatDiffuser is one of the few moving towards undestanding discriminative capabilities of diffusion-learned features \cite{jia2025can}. In contrast, our work provides a comprehensive investigation of the discriminative capabilities of diffusion-based generative models across multiple RS tasks. By moving beyond task-specific solutions and limited testing, our method advances the broader potential of diffusion-driven GFMs pretrained on global-scale data.

Among the most recent models, transformer-based multimodal foundation models have emerged to offer greater flexibility and scalability in handling diverse modalities and enabling cross-modal conditioning, especially for discrimantive purposes \cite{olmoearth2024, tseng2025galileo, astruc2024omnisat, xiong2024dofa}. Some of them are shaped for both generation and feaure extraction. TerraMind \cite{jakubik2025terramind} is a representative example of this, using a dual scale formulation that combines token-level and pixel-level inputs across nine modalities on co-registered 10\,m grids, including Sentinel-1, Sentinel-2, DEM, LULC and NDVI image captions, and discretized geographic coordinates. Following a two-stage pretraining process on all modalities, TerraMind first trains modality-specific tokenizers to map image-like modalities into sequences of discrete tokens by encoding each 16$\times$16 patch and discretizing it with finite scalar quantization (FSQ). The captions and geographic coordinates are treated as text and tokenized using a modified WordPiece tokenizer. In the second stage, TerraMind pretrains a symmetric transformer encoder-decoder jointly across modalities using a dual-scale early fusion strategy where it processes both token-level as pixel-level representations during masked token prediction, enabling cross modal generation. Since TerraMind achieves state-of-the-art performance on a range of multimodal EO tasks and is widely considered a strong baseline in this domain we include TerraMind in our experimental comparisons.

Nevertheless, while such models represent a significant step forward, several architectural constraints persist across existing EO GFMs. Transformer-based multimodal systems such as TerraMind typically adopt fixed-size patching on co-registered grids, which forces all modalities onto a common spatial grid. This process discards their heterogeneous native resolutions and can obscure physically meaningful relationships, particularly in cross-modal settings where each sensor captures distinct spatial structures. Diffusion models built on UNet-style backbones generally support a limited set of modalities and often rely on concatenation or modality-specific encoder branches. Such designs complicates the fusion and scaling of multiple modalities and sensors.

In contrast, COP-GEN is a transformer-based diffusion model, that offers a more flexible and physically grounded generative framework. By representing each modality as a sequence of latent tokens and processing them jointly through a unified transformer backbone, COP-GEN naturally enable cross-modal attention, scalable multimodal fusion, and any-to-any conditional generation. Crucially, tokenization allows each modality to be trained at its native spatial resolution, eliminating the need for aggressive resampling. To our knowledge, no prior EO generative model has combined a pure transformer diffusion backbone with tokenized multimodal inputs, native sensor resolutions, and unified modeling of the joint distribution across heterogeneous Copernicus modalities.

{\begin{figure*}
    \centering
    \includegraphics[width=\linewidth]{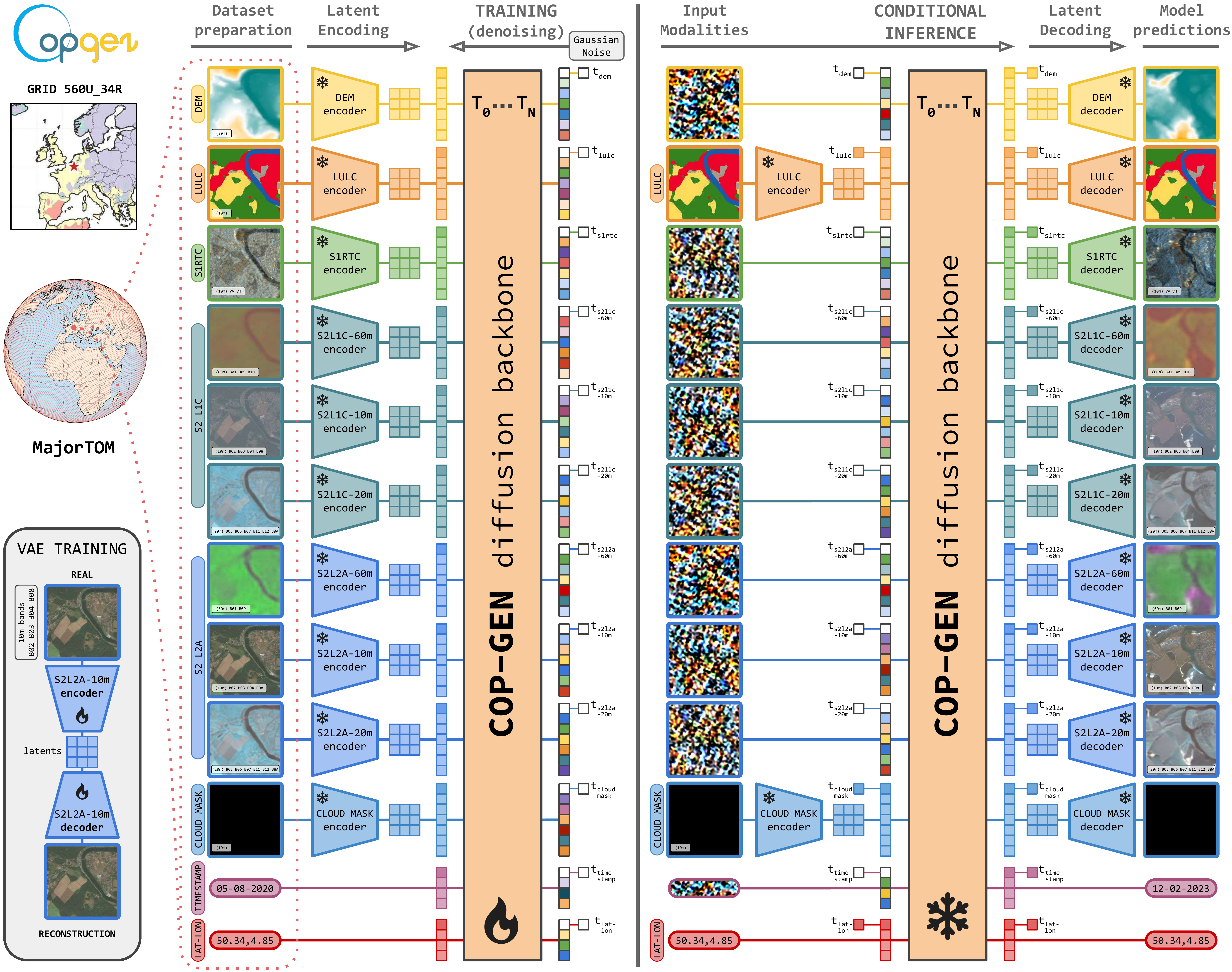}
    \caption{\textbf{COP-GEN architecture, training, and inference overview.} Multimodal inputs (optical, radar, elevation, land-cover, geolocation, and timestamps) are encoded into latent representations using modality-specific VAEs (or directly tokenized for scalar inputs). All tokens, augmented with modality-specific diffusion timestep embeddings, are processed by a shared transformer diffusion backbone. The model is trained to jointly predict noise for all modalities. At inference, modalities can be either sampled from noise or fixed at timestep zero, enabling both unconditional generation and flexible any-to-any conditional translation across modalities.}
    \label{fig:full-copgen-architecture}
\end{figure*}
}

\section{Methodology}
\label{sec:methodolgy}

COP-GEN is a unified multimodal generative model designed to approximate the joint distribution of heterogeneous EO modalities. The model combines modality-specific latent encoders with a shared transformer-based diffusion backbone, enabling flexible any-to-any conditional generation across sensors with different physical meanings and spatial resolutions (Figure~\ref{fig:full-copgen-architecture}).

\subsection{Dataset and Modalities}

We build a globally distributed, paired, multimodal dataset derived from MajorTOM~\cite{francis2024major}, covering optical, radar, elevation, semantic, temporal, and geolocation information. Land Use Land Cover (LULC) maps are sourced from~\cite{impact_observatory_lulc_2023}. It consists of 1,017,469 globally distributed samples, covering all major climate zones and land-cover types. Each data sample corresponds to a fixed geographic tile and includes the following modalities:

\begin{itemize}[leftmargin=2.5em]
    \item Sentinel-2 Level 2A (S2L2A)
    \item Sentinel-2 Level 1C (S2L1C)
    \item Sentinel-1 RTC (S1RTC)
    \item Digital Elevation Model (DEM)
    \item Land Use / Land Cover (LULC)
    \item Geographic location (latitude–longitude)
    \item Acquisition timestamp (mean of S1 and S2)
\end{itemize}

To account for the different native spatial and spectral resolutions, we group inputs by modality–resolution pairs and process each at its native scale. Table~\ref{tab:data_config} summarizes the input sources, how they are grouped, resolutions, patch sizes, and latent grid dimensions used throughout the model. This design avoids aggressive resampling and preserves physical spatial structures within each modality.

{\begin{table}[h]
\centering
\resizebox{\linewidth}{!}{%
\begin{tabular}{l l c c c}
\toprule
\textbf{Source} & \textbf{Bands / Data} & \textbf{Res} & \textbf{Patches} & \textbf{Latent} \\
\midrule

\multirow{3}{*}{S2L2A}
 & B4,3,2,8 & 10m & $192 \times 192$ & $24 \times 24$ \\
 & B5,6,7,8A,11,12 & 20m & $96 \times 96$ & $12 \times 12$ \\
 & B1,9 & 60m & $32 \times 32$ & $4 \times 4$ \\
\midrule

\multirow{4}{*}{S2L1C}
 & B3,2,8 & 10m & $192 \times 192$ & $24 \times 24$ \\
 & B5,6,7,8A,11,12 & 20m & $96 \times 96$ & $12 \times 12$ \\
 & B1,9,10 & 60m & $32 \times 32$ & $4 \times 4$ \\
 & cloud-mask & 10m & $192 \times 192$ & $24 \times 24$ \\
\midrule

S1RTC & VV,VH & 10m & $192 \times 192$ & $24 \times 24$ \\
\midrule

DEM & DEM & 30m & $64 \times 64$ & $8 \times 8$ \\
\midrule

LULC & LULC & 10m & $192 \times 192$ & $24 \times 24$ \\
\midrule

lat-lon & Unit Vector $(x, y, z)$ & - & - & $1 \times 3$ \\
\midrule
time & $\sin/\cos(\text{DoY}), \text{Year}$ & - & - & $1 \times 3$ \\

\bottomrule
\end{tabular}%
}
\caption{\textbf{Data configuration and tokenization}. Overview of input data sources, spectral bands, spatial resolutions, patch sizes, and corresponding latent grid dimensions used by the model. Multi-resolution Sentinel-2 imagery is patchified and encoded at resolution-specific latent scales, while geolocation and temporal metadata are embedded as global tokens.}
\label{tab:data_config}
\end{table}
}

\subsection{Latent Representation learning}

To enable efficient diffusion modeling, all high-dimensional spatial modalities are first encoded into compact latent representations using modality-specific variational autoencoders (VAEs).

\subsubsection{VAE Training}

For each modality–resolution group, we train an independent KL-regularized variational autoencoder (VAE) following the architecture of~\cite{rombach2022ldm}. The encoder–decoder consists of convolutional blocks with ResNet layers, group normalization, and Swish activations, progressively downsampling the input by a factor of 8 to produce an 8-channel latent embedding. Training uses a composite objective combining L1 and MSE reconstruction losses, LPIPS perceptual loss, KL divergence regularization (weight $10^{-6}$), and an adversarial loss implemented via a PatchGAN discriminator with hinge loss. To ensure stable reconstruction learning, the adversarial component is disabled during the first 50,000 training steps and introduced thereafter. Adaptive loss balancing based on gradient magnitudes is applied to stabilize optimization. Training is performed using AdamW with an initial learning rate of $5 \times 10^{-6}$, polynomial decay with exponent 0.95 (minimum learning rate $1 \times 10^{-6}$), gradient clipping with maximum norm 1.0, and exponential moving average of model weights ($\beta = 0.995$). Each VAE is trained for 500,000 steps. Once trained, all VAEs are frozen and used to pre-encode the entire dataset into latent space, substantially reducing the computational cost of training the diffusion backbone.

\subsubsection{Encoding of Geolocation and Time}

Geolocation and temporal information are incorporated as non-image tokens and do not require VAEs.

\textbf{Latitude-Longitude.}
We discretize the Earth into a fixed global 10 km grid and compute the geometric center of each tile. Coordinates are encoded as 3D Cartesian unit vectors $(x, y, z)$ on the sphere, ensuring rotational continuity and avoiding singularities at the poles. These embeddings are precomputed and stored alongside image latents.

\textbf{Timestamps.}
For each tile, we compute a representative acquisition date by taking the temporal midpoint between Sentinel-1 and Sentinel-2 observations. Dates are encoded using sine-cosine representations of day-of-year (to preserve circularity) and a normalized year scalar. This encoding captures both seasonal periodicity and long-term temporal trends.

\subsection{Unified Multimodal Diffusion Model}

COP-GEN models the joint distribution of all modalities using a single denoising diffusion process operating in latent space.

\subsubsection{Tokenization and Input Representation}

Each modality latent is tokenized using a modality-specific patch embedding. Image latents are patchified using $2\times2$ spatial patches. Scalar modalities (time, geolocation) use single-token embeddings.

Each modality is assigned its own diffusion timestep $t^{(i)}$, encoded via sinusoidal timestep embeddings and prepended as dedicated tokens. All modality tokens, including timestep tokens, are concatenated into a single sequence and augmented with learnable positional embeddings.

This unified token sequence allows the model to attend across modalities while preserving modality identity.

\subsubsection{Transformer Diffusion Backbone}

The denoising network is a U-shaped Vision Transformer (U-ViT \cite{bao2023uvit}) with 20 transformer layers, 1024-dimensional embeddings, and 16 attention heads. Long skip connections connect encoder and decoder blocks, preserving spatial detail across diffusion steps.

The model predicts noise for all modalities jointly using the standard DDPM $\epsilon$-prediction objective:

\[
\mathcal{L} = \mathbb{E}_{t, \mathbf{z}_0, \boldsymbol{\epsilon}} 
\sum_{i=1}^{M} \left\| \boldsymbol{\epsilon}^{(i)} - \hat{\boldsymbol{\epsilon}}_\theta^{(i)} \right\|_2^2,
\]

where $M$ is the number of modalities.

Training follows the Stable Diffusion noise schedule with 1000 timesteps. Optimization uses AdamW (learning rate $2\times10^{-4}$ weight decay 0.03), linear warmup, mixed precision, gradient checkpointing, Flash Attention, and EMA ($0.9999$). Training proceeds for 500,000 iterations with an effective batch size of 4096.

\subsection{Conditional and Unconditional Sampling}

A key feature of COP-GEN is independent timestep control per modality, enabling flexible generation scenarios without retraining.

\subsubsection{Joint Unconditional Generation}

In unconditional generation, all modalities are initialized from Gaussian noise and denoised jointly. This samples from the full joint distribution:

\[
p(\mathbf{z}^{(1)}, \dots, \mathbf{z}^{(M)}),
\]

producing consistent synthetic multimodal scenes.

\subsubsection{Any-to-Any Conditional Generation}

For conditional generation, modalities are partitioned into conditioning set $C$, and generation set $G$.

Modalities in $C$ are fixed at timestep $t=0$, while modalities in $G$ undergo the diffusion process from noise. During each denoising step, all modalities attend to one another through the shared transformer backbone, enabling cross-modal conditioning.

This mechanism supports arbitrary translation tasks such as: DEM to optical imagery, SAR and DEM to LULC, S2L1C to S2L2A (atmospheric correction), and partial conditioning with timestamps or geolocation.

Since conditioning is implemented purely through timestep control, COP-GEN supports zero-shot modality translation for any learned modality subset.

\subsection{Stochastic Benchmark}
\label{sec:benchmark}

    Standard reconstruction metrics measure peak single-sample fidelity but do not assess whether a model's \emph{distribution} of outputs matches the true distribution of plausible observations.
    Since COP-GEN is designed to capture this variability explicitly, we construct
    a dedicated benchmark that compares generated sample sets against real multi-temporal Sentinel-2 acquisitions at the same locations.

\subsubsection{Benchmark Dataset}

We draw 495 grid cells from the test set and assemble a multi-temporal reference set for each.
Per cell, we collect 16 near-cloud-free Sentinel-2 L2A acquisitions spanning 2018--2025, cropped to 1056$\times$1056 pixels at 10\,m resolution following the updated MajorTOM tiling convention, with all 12 spectral bands resampled to 10\,m via nearest-neighbour interpolation.
Evaluation is performed on the 192$\times$192 centre crop matching the model output footprint.

For each cell, we generate 33 COP-GEN samples (one per seed) and 16 TerraMind samples, all conditioned on the same DEM and LULC inputs.
To compare diversity on equal footing, we retain a deterministic 16-seed COP-GEN subsample per cell, selected via SHA-256 hashes of the cell IDs, yielding a balanced 16/16/16 split across real, COP-GEN, and TerraMind.
After discarding 6 cells whose footprint falls outside valid S2 coverage, 489 cells remain, giving 7{,}824 acquisitions per source.

\subsubsection{Evaluation Metrics}

Metrics are organised into two complementary streams.
All metrics are computed per cell and aggregated as mean $\pm$ std across the 489 cells.

\paragraph{Stream~1: Perceptual fidelity.}
We evaluate similarity in three embedding spaces:
\begin{enumerate}[leftmargin=2em]
    \item \textbf{Spectral vectors} (primary): global-mean-pooled 12-band reflectance vectors $(N, 12)$. Encoder-free and physically interpretable.
    \item \textbf{ResNet-50 (ImageNet)}: 2048-D features extracted from B04/B03/B02 RGB composites, capturing spatial texture and structure.
    \item \textbf{LPIPS (AlexNet)}: learned perceptual image patch similarity on RGB. Being a pairwise perceptual distance rather than a vector embedding, LPIPS is used only for the intra-set diversity statistic.
\end{enumerate}

In the spectral and ResNet-50 embedding spaces we compute all three of the following; for LPIPS only the intra-set distance applies:
\begin{itemize}[leftmargin=2em]
    \item \textbf{1-Nearest Neighbour accuracy} (1-NN)~\cite{lopez2017revisiting}.
    A two-sample test: we pool real and generated embeddings, label each sample by its source, and for every sample check whether its nearest neighbour in the combined pool shares the same label. Accuracy near 0.5 indicates that the real and generated sets are perfectly intermixed in feature space (ideal); values near 1.0 indicate they are trivially separable.

    \item \textbf{Precision} (realism) and \textbf{Recall} (diversity)~\cite{kynkaanniemi2019improved}.
    The union of $k$-nearest-neighbour balls around real samples empirically approximates the real-data manifold, and likewise for generated samples. \textbf{Precision} measures the fraction of generated samples inside the real manifold (how often the generator produces realistic outputs). \textbf{Recall} measures the fraction of real samples inside the generated manifold (how well the generator covers the range of real outcomes). A generator that collapses to the conditional mean can achieve high precision but near-zero recall, whereas a well-calibrated stochastic generator should score high on both. We use $k{=}5$ following the original work.

    \item \textbf{Mean intra-set pairwise distance}. The average pairwise distance between samples within a single set, measured in embedding space. It is an assumption-free diversity statistic: values near zero indicate mode collapse, while values close to the real intra-set distance indicate the generator matches the natural variability of the observations. We report it for real, COP-GEN, and TerraMind side-by-side to situate each generator against the empirical ground truth.
\end{itemize}

\paragraph{Stream~2: Physical consistency.}
Each image is reduced to a 12-dimensional \emph{spectral vector} by spatially averaging reflectance within each band. For a cell with $N$ images, this yields an $(N, 12)$ array per cell. This preserves
band co-occurrence (e.g.\ the relative magnitudes of NIR, Red, and SWIR
that characterise vegetation or water) while discarding spatial noise.
All Stream~2 metrics operate on these per-image spectral vectors.

\begin{itemize}[leftmargin=2em]
    \item \textbf{Maximum Mean Discrepancy (MMD)}.
    A kernel two-sample statistic measuring the distance between real and generated spectral-vector distributions. We use an RBF kernel with bandwidth set to half the median pairwise distance in the combined real + generated pool (median heuristic). MMD is sensitive to mean shift, variance mismatch, and higher-order moments, providing a single summary of distributional similarity. Lower is better.

    \item \textbf{1-D Wasserstein distance, per band}.
    For each of the 12 bands we form two 1-D distributions---the reflectance values across real samples and across generated samples---and compute the Wasserstein (earth-mover's) distance between them. We report the mean across bands. Unlike MMD this metric only checks marginals, so a generator that always outputs the conditional mean can score well even when its joint distribution is collapsed.

    \item \textbf{Spectral range coverage}.
    The fraction of each band's real $[\min, \max]$ reflectance range that is also spanned by the generated samples, averaged across bands. Unlike Wasserstein, which is dominated by the densest part of the distribution, range coverage is sensitive to the extremes and    therefore well suited to detecting conservative, mode-collapsed generators. Higher is better.
\end{itemize}

{\begin{figure*}
    \centering
    \includegraphics[width=\linewidth]{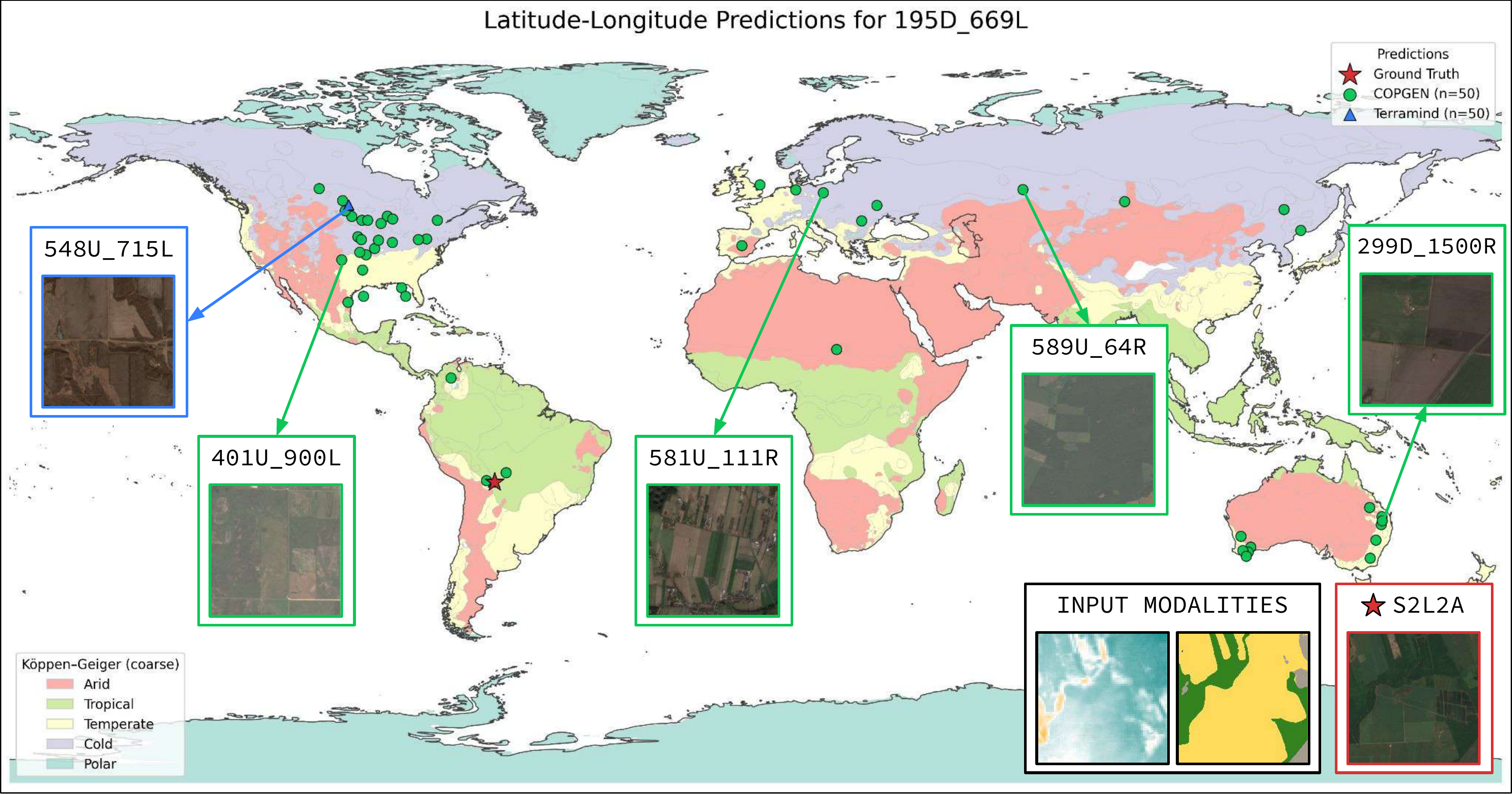}
    \caption{\textbf{Geospatial Distribution Analysis.} We predict latitude--longitude coordinates conditioned on DEM and LULC inputs ($n=50$ runs). TerraMind (\textcolor{terramindblue}{blue}) collapses to a few locations, whereas COP-GEN (\textcolor{copgengreen}{green}) predicts a distribution of plausible locations that share similar topographic and biome characteristics, accurately reflecting the non-injective nature of the mapping. A Köppen--Geiger climate classification basemap is overlaid to provide climatic context for the predicted locations. The ground-truth acquisition location is indicated by a red star ($\textcolor{red}{\bigstar}$), and real thumbnail visualisations of the predicted locations are shown for comparison.}
    \label{fig:lat_lon_195D}
\end{figure*}

}
{\begin{figure*}
    \centering
    \includegraphics[width=\linewidth]{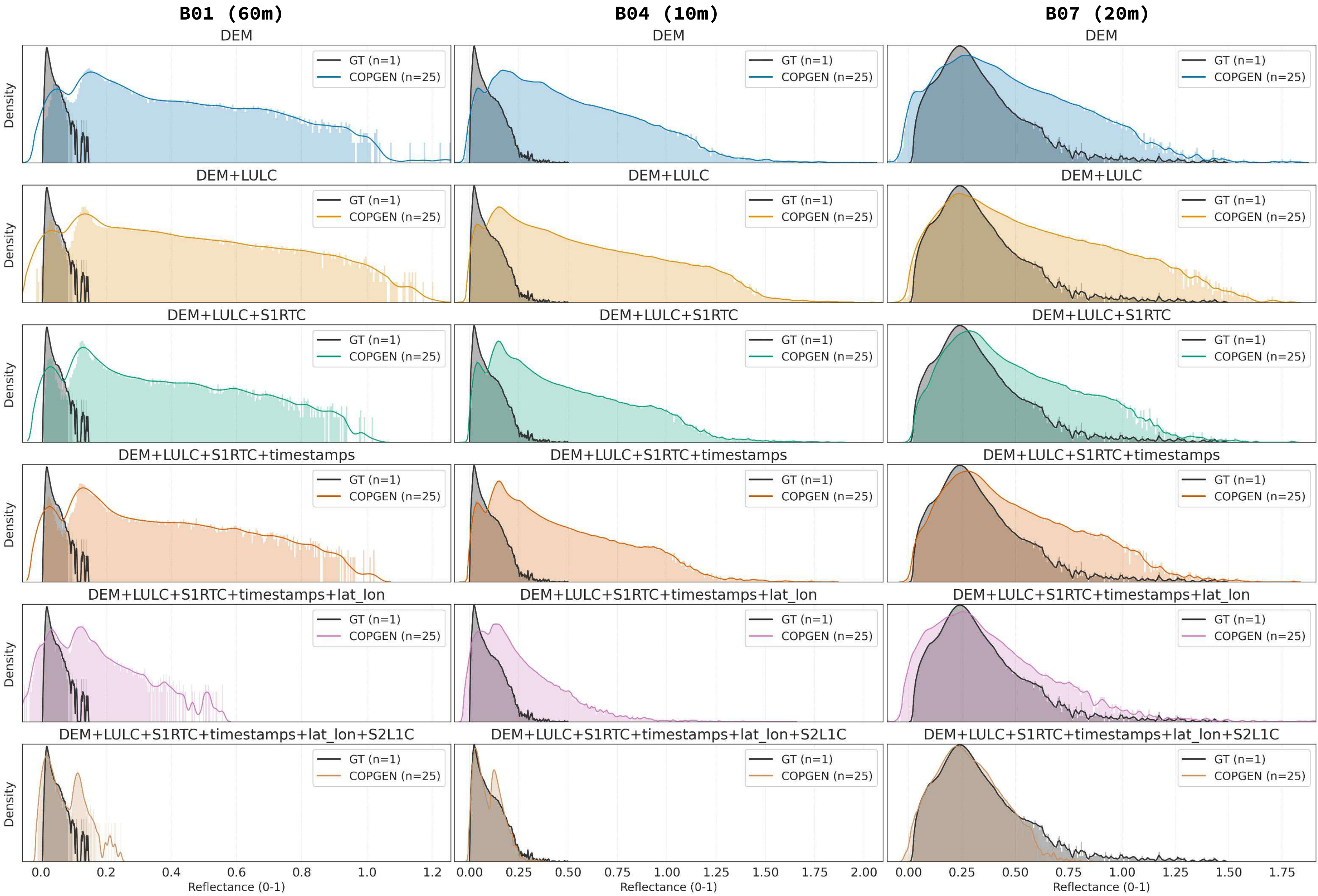}
    \caption{\textbf{Distribution spread narrowing by progressively increasing input conditioning.} As additional modalities are provided as input, the generated samples better align with the ground-truth (GT) distribution. For each conditioning setup, we generate 25 stochastic samples of Sentinel-2 L2A (S2L2A) imagery and report the predicted band distributions using histograms and kernel density estimates (KDEs). One spectral band is selected per S2L2A spatial resolution. The legend indicates the set of input modalities used for conditioning, always for a fixed geographic tile (215U\_1019R). Additional bands and visualisations are provided in the appendix.}
    \label{fig:distribution-teaser}
\end{figure*}
}
{\begin{figure*}
    \centering
    \includegraphics[width=\linewidth]{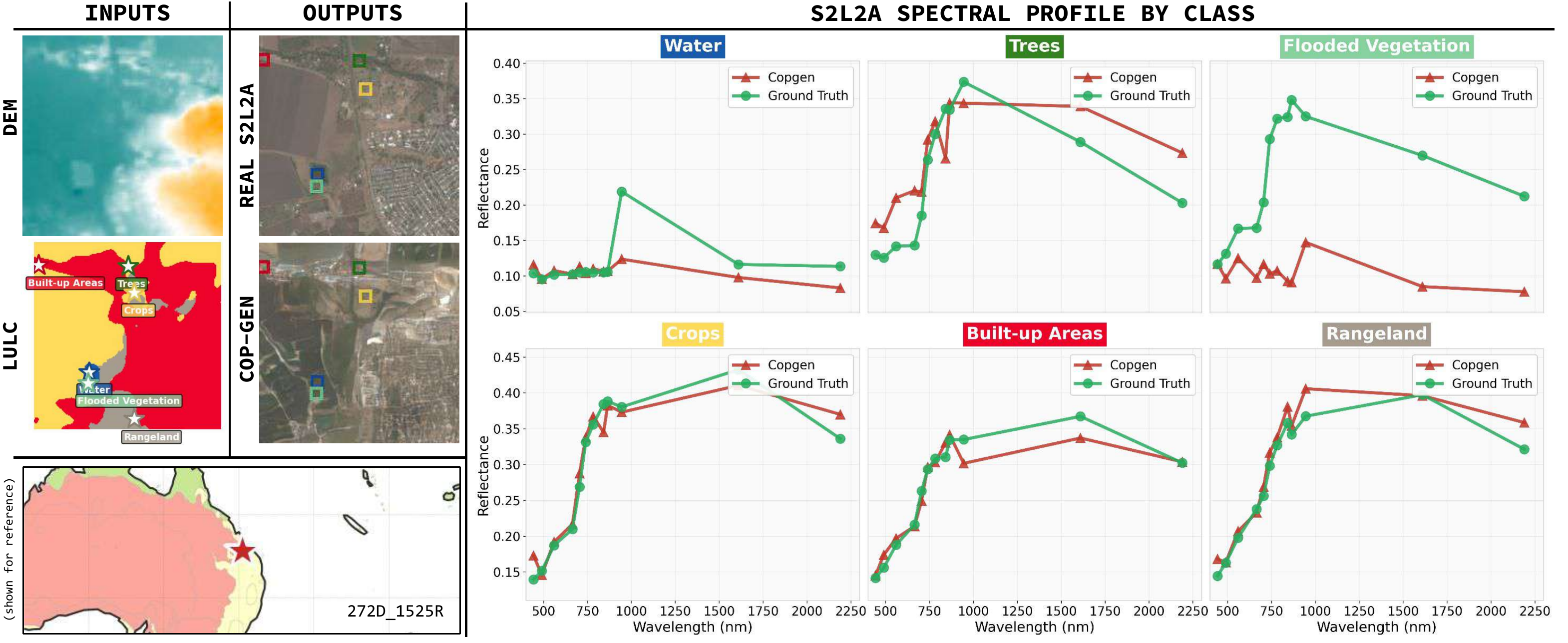}
    \caption{\textbf{Per-pixel spectral reflectance profiles across multiple LULC classes.} Conditioned on DEM and LULC inputs, COP-GEN generates multispectral S2L2A imagery that captures physically consistent spectral relationships. The plots compare spectral profiles from selected pixel locations in both real and generated images across the Sentinel-2 bands. The close alignment for trees, bare soil, water, crops, built-up areas, etc. demonstrates the model's ability to accurately reconstruct characteristic land-cover responses. Geographical location is provided for reference. Additional visualisations are provided in Supplementary Material.}
    \label{fig:272D_1525R_spectral_profile}
\end{figure*}
}
{\begin{figure*}
    \centering
    \includegraphics[width=\linewidth]{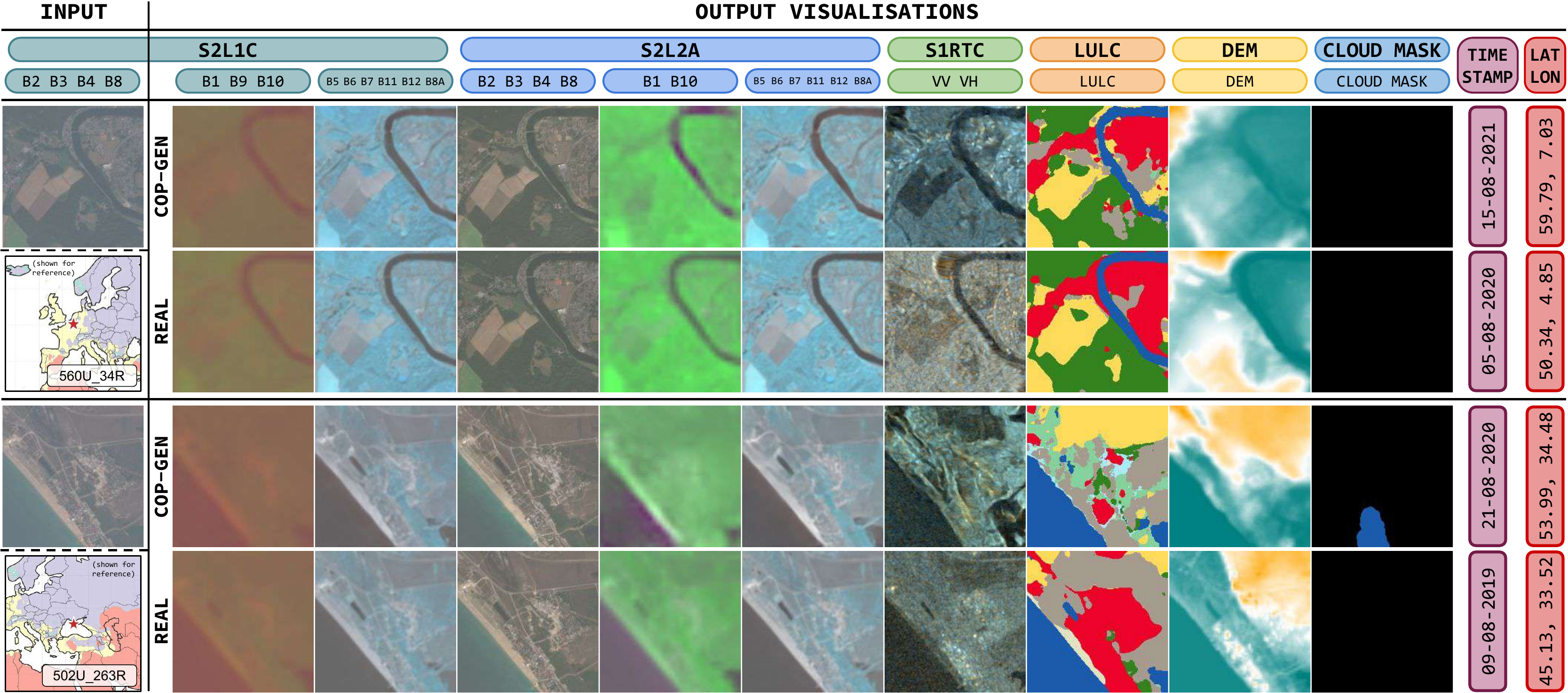}
    \caption{\textbf{Band infilling via resolution-specific latent groups.} By grouping Sentinel-2 bands according to resolution, COP-GEN treats spectral groups as independent modalities. Here, the model is conditioned only on the high-resolution visible band group (B2, B3, B4, B8) and successfully reconstructs the remaining spectral bands, auxiliary sensors (S1RTC, DEM), LULC maps, cloud mask, timestamp, and location. The high fidelity of the generated bands demonstrates effective cross-band and cross-modal inference. Geographical location thumbnail is provided for reference only.}
    \label{fig:band-infilling}
\end{figure*}
}

{\begin{table}[h]
\centering
\resizebox{\linewidth}{!}{%
\begin{tabular}{l l cc}
\toprule
& & \multicolumn{2}{c}{\textbf{Peak perf. (Best per Tile)}} \\
\cmidrule(lr){3-4}
\textbf{Target} & \textbf{Metric} & \textbf{COP-GEN} & \textbf{TerraMind} \\
\midrule

\multirow{2}{*}{\textbf{DEM}} 
 & MAE $\downarrow$  & \textbf{26.80} & 145.62 \\
 & SSIM $\uparrow$ & \textbf{0.45}  & 0.44 \\
\midrule

\multirow{2}{*}{\textbf{LULC}} 
 & Top-1 $\uparrow$ & \textbf{0.84} & 0.80 \\
 & mIoU $\uparrow$  & 0.42          & \textbf{0.55} \\
\midrule

\multirow{2}{*}{\textbf{S1RTC}} 
 & MAE $\downarrow$  & \textbf{2.63}  & 2.64 \\
 & PSNR $\uparrow$ & 16.83          & \textbf{19.65} \\
\midrule

\multirow{2}{*}{\textbf{S2L1C}} 
 & MAE $\downarrow$  & \textbf{0.02}  & 0.11 \\
 & PSNR $\uparrow$ & \textbf{21.16} & 12.77 \\
\midrule

\multirow{2}{*}{\textbf{S2L1C$^{\dagger}$}}
 & MAE $\downarrow$  & \textbf{0.05}  & 0.12 \\
 & PSNR $\uparrow$ & \textbf{13.92} & 12.68 \\
\midrule

\multirow{2}{*}{\textbf{S2L2A}} 
 & MAE $\downarrow$  & \textbf{0.02}  & 0.10 \\
 & PSNR $\uparrow$ & \textbf{22.47} & 17.46 \\
\midrule

\multirow{2}{*}{\textbf{S2L2A$^{\ddagger}$}} 
 & MAE $\downarrow$  & \textbf{0.06}  & 0.10 \\
 & PSNR $\uparrow$ & \textbf{14.40} & 16.18 \\
\midrule

\textbf{LatLon} 
 & Mean km $\downarrow$ & 98.35 & \textbf{94.25} \\

\bottomrule
\end{tabular}%
}
\caption{\textbf{Tile-Level Peak Capability Analysis.} We report the oracle performance (best generation selected per tile) to demonstrate the upper bound of generation quality. This metric isolates the model's peak capability from stochastic variance. \textbf{Bold} indicates the best result. $^{\dagger}$ indicates that S2L2A is not present among the input modalities. $^{\ddagger}$ indicates that S2L1C is not present among the input modalities. }
\label{tab:tile_analysis_best}
\end{table}
}
{\begin{table}[t]
\centering
\resizebox{\columnwidth}{!}{%
\begin{tabular}{l l l c c}
\toprule
& & & \textbf{COP-GEN} & \textbf{TerraM} \\
\textbf{Target} & \textbf{Removed} & \textbf{Metric} & \textbf{Best} & \textbf{Best} \\
\midrule

\multirow{5}{*}{\textbf{DEM}} 
  & \textit{w/o LatLon} & MAE $\downarrow$ & \textbf{47.44} & 140.01 \\
  & \textit{w/o LULC}   & MAE $\downarrow$ & \textbf{46.96} & 140.80 \\
  & \textit{w/o S1RTC}  & MAE $\downarrow$ & \textbf{54.85} & 140.86 \\
  & \textit{w/o S2L1C}  & MAE $\downarrow$ & \textbf{51.85} & 146.00 \\
  & \textit{w/o S2L2A}  & MAE $\downarrow$ & \textbf{45.78} & 146.71 \\
\midrule

\multirow{5}{*}{\textbf{LULC}} 
  & \textit{w/o LatLon} & Top-1 Acc $\uparrow$ & \textbf{0.81} & 0.80 \\
  & \textit{w/o DEM}    & Top-1 Acc $\uparrow$ & \textbf{0.80} & \textbf{0.80} \\
  & \textit{w/o S1RTC}  & Top-1 Acc $\uparrow$ & 0.79 & \textbf{0.80} \\
  & \textit{w/o S2L1C}  & Top-1 Acc $\uparrow$ & \textbf{0.80} & \textbf{0.80} \\
  & \textit{w/o S2L2A}  & Top-1 Acc $\uparrow$ & \textbf{0.80} & \textbf{0.80} \\
\midrule

\multirow{5}{*}{\textbf{S1RTC}} 
  & \textit{w/o LatLon} & MAE $\downarrow$ & 2.70 & \textbf{2.63} \\
  & \textit{w/o DEM}    & MAE $\downarrow$ & 2.75 & \textbf{2.63} \\
  & \textit{w/o LULC}   & MAE $\downarrow$ & 2.76 & \textbf{2.63} \\
  & \textit{w/o S2L1C}  & MAE $\downarrow$ & 2.70 & \textbf{2.64} \\
  & \textit{w/o S2L2A}  & MAE $\downarrow$ & 2.68 & \textbf{2.62} \\
\midrule

\multirow{5}{*}{\textbf{S2L1C}} 
  & \textit{w/o LatLon} & MAE $\downarrow$ & \textbf{0.02} & 0.11 \\
  & \textit{w/o DEM}    & MAE $\downarrow$ & \textbf{0.02} & 0.11 \\
  & \textit{w/o LULC}   & MAE $\downarrow$ & \textbf{0.02} & 0.11 \\
  & \textit{w/o S1RTC}  & MAE $\downarrow$ & \textbf{0.02} & 0.11 \\
  & \textit{w/o S2L2A}  & MAE $\downarrow$ & \textbf{0.06} & 0.12 \\
\midrule

\multirow{5}{*}{\textbf{S2L2A}} 
  & \textit{w/o LatLon} & MAE $\downarrow$ & \textbf{0.02} & 0.10 \\
  & \textit{w/o DEM}    & MAE $\downarrow$ & \textbf{0.02} & 0.10 \\
  & \textit{w/o LULC}   & MAE $\downarrow$ & \textbf{0.02} & 0.10 \\
  & \textit{w/o S1RTC}  & MAE $\downarrow$ & \textbf{0.02} & 0.10 \\
  & \textit{w/o S2L1C}  & MAE $\downarrow$ & \textbf{0.07} & 0.10 \\
\midrule

\multirow{5}{*}{\textbf{LatLon}} 
  & \textit{w/o DEM}    & Mean km $\downarrow$ & 210.54 & \textbf{90.67} \\
  & \textit{w/o LULC}   & Mean km $\downarrow$ & 188.70 & \textbf{95.50} \\
  & \textit{w/o S1RTC}  & Mean km $\downarrow$ & 173.09 & \textbf{78.23} \\
  & \textit{w/o S2L1C}  & Mean km $\downarrow$ & 193.43 & \textbf{138.83} \\
  & \textit{w/o S2L2A}  & Mean km $\downarrow$ & 182.45 & \textbf{77.41} \\

\bottomrule
\end{tabular}%
}
\caption{\textbf{Tile-Level Summarized Best Performance (Leave-One-Out).} We compare the peak performance of \textbf{COP-GEN} against TerraMind across all modalities. COP-GEN shows clear dominance in DEM reconstruction (MAE) and Optical bands (S2L1C/S2L2A), while TerraMind demonstrates stronger localization capabilities (LatLon).}
\label{tab:leave_one_out_summary}
\end{table}
}

\section{Results}
\label{sec:results}

Evaluating generative models in Earth Observation presents a unique challenge: the mapping between sparse input conditions and physical scenes is inherently one-to-many. For example, a specific Digital Elevation Model (DEM) and Land Cover (LULC) layout may may correspond to a wide range of physically plausible optical, radar, or atmospheric realizations.
However, standard EO benchmarks typically provide only a single snapshot of the Earth for each input condition.
As a result, conventional pointwise metrics such as MAE or PSNR systematically favor models that regress toward a conditional mean (often resulting in blurry, texture-less predictions), rather than those that faithfully represent the underlying data distribution.

To account for this mismatch, we adopt a Peak-Capability (oracle) evaluation protocol: the model's ability to generate a sample close to the ground truth within its distribution. 
For each test tile and conditioning setup, we generate multiple independent samples and report the best-matching generation per tile according to the target metric. This protocol decouples the model’s representational capacity from the stochastic variability and answers the question: does the model’s learned distribution contain high-fidelity realisations consistent with the ground truth? The same procedure is applied to competing methods ensuring a fair comparison.

Importantly, this evaluation does not claim that oracle selection is a deployable inference strategy. Rather, it provides an upper bound on generation quality and highlights differences in the support and diversity of the learned distributions. As discussed below, this distinction is crucial when comparing highly stochastic diffusion models against models optimized for deterministic reconstruction.

In line with this evaluation philosophy, qualitative experiments are designed to illustrate distributional support, multimodal plausibility, and conditional variability rather than to rank models according to single-reference metrics.

\subsection{Qualitative results}

We assess the physical plausibility and diversity of the generated data through specific qualitative experiments: visual variance, spatial localization, spectral consistency, and band infilling.

\subsubsection{Output Diversity Beyond Single-Image Metrics}

Figure~\ref{fig:teaser} illustrates that COP-GEN generates a broad and diverse set of high-frequency, realistic scenes, varying in illumination, atmospheric conditions, and spectral appearance, while consistently respecting the underlying terrain and land-cover structure. This behaviour reflects the expected stochasticity of diffusion models, aligned with the inherent ambiguity of the task.
In contrast, deterministic baselines (e.g. Terramind) tend to exhibit a narrow output distribution, producing visually similar samples across repeated generations. While this behavior yields strong MAE performance against a single ground truth snapshot, it limits the model’s ability to represent the diversity of valid physical outcomes.
Additional examples across multiple tiles are provided in Appendix \ref{sec:supplementary}.

\subsubsection{Spatial Distribution (Lat-Lon Distributions)}
We assess whether the learned distributions capture global spatial variability. Firstly, models predict latitude–longitude coordinates conditioned only on DEM and LULC, repeated 50 times per tile.
As shown in Figure~\ref{fig:lat_lon_195D}, COP-GEN generates a broad spatial distribution, covering geographically diverse regions that share similar geomorphological and land-cover characteristics. Predicted locations span multiple continents (e.g., agricultural regions in Europe and North America), consistent with the under-specified nature of the input conditions. 
On the other hand, TerraMind produces highly concentrated predictions clustered around a small number of locations, suggesting memorization or deterministic collapse. 

Secondly, (Appendix \ref{sec:supplementary}), we condition the models on a single homogeneous LULC class (e.g., water or trees). COP-GEN predictions align with known geographic priors, such as water bodies clustering along coastlines and river basins, demonstrating that the model internalises global geospatial semantics.

\subsubsection{Distribution Narrowing Under Increased Conditioning}
To verify that COP-GEN appropriately narrows its output distribution as conditioning becomes more informative, we progressively add input modalities when generating Sentinel-2 L2A imagery. Starting from DEM-only conditioning, we incrementally include LULC, S1 RTC, timestamps, latitude–longitude, and finally Sentinel-2 L1C.

Figure~\ref{fig:distribution-teaser} shows per-band reflectance histograms over 25 samples per setting. As additional modalities are introduced, the output distributions systematically narrow and shift toward the ground-truth histogram, indicating increased certainty and reduced ambiguity. Furthermore, even under sparse input conditioning, the generated distributions cover the true reflectance range, confirming that COP-GEN captures the support of the data distribution rather than collapsing prematurely. Additional tiles are shown in Appendix \ref{sec:supplementary}.

\subsubsection{Spectral Fidelity Across Land-Cover Classes}
Finally, we examine per-pixel spectral responses stratified by LULC classes. Figure~\ref{fig:272D_1525R_spectral_profile} compares ground truth spectra with generations from COP-GEN. COP-GEN closely follows the characteristic spectral signatures of different land-cover types across Sentinel-2 bands, while maintaining realistic variability. This indicates that the model captures physically meaningful relationships across modalities rather than superficial texture statistics. Further examples are provided in Appendix \ref{sec:supplementary}.

\subsubsection{Band-infilling}

We evaluate COP-GEN’s ability to perform band infilling by conditioning on a subset of Sentinel-2 L1C bands (B2, B3, B4, B8) and generating the remaining spectral bands. Since COP-GEN partitions Sentinel-2 data into resolution-specific latent groups, each group is treated as an independent modality, enabling the model to infer missing spectral bands without explicit supervision. Given only the visible and near-infrared bands, COP-GEN successfully reconstructs the remaining Sentinel-2 L1C and L2A bands, as well as auxiliary modalities including DEM, S1RTC, LULC, timestamp, and geolocation. As shown in Figure~\ref{fig:band-infilling}, the generated bands closely match the ground truth structure, demonstrating that COP-GEN learns coherent cross-band and cross-modal relationships and can flexibly infill missing spectral information under sparse conditioning.

\subsection{Quantitative Results}
\label{sec:quantitative}

Quantitative results are reported to assess peak generative fidelity across modalities, using an oracle evaluation protocol to account for the inherently one-to-many nature of EO mappings.

\subsubsection{Predicting Missing Modalities}
Table~\ref{tab:tile_analysis_best} reports tile-level peak performance across all target modalities. COP-GEN consistently outperforms TerraMind on DEM reconstruction, optical modalities (S2L1C and S2L2A), and SAR (S1 RTC) in terms of MAE and PSNR, demonstrating superior peak generative fidelity. Performance gains are particularly pronounced when predicting optical imagery without access to closely related optical inputs, highlighting COP-GEN’s ability to exploit cross-modal structure. Terramind achieves slightly higher performance on LatLon regression.

\subsubsection{Leave-One-Out Analysis}

Table~\ref{tab:leave_one_out_summary} presents a leave-one-out analysis that removes each conditioning modality in turn, revealing the physical couplings learned by the model. COP-GEN shows strong robustness when predicting DEM and optical targets. Optical generation degrades most noticeably when removing closely related optical inputs, as expected, while remaining stable under removal of weaker correlates. This confirms that the model utilizes strong spectral correlations between levels of processing.

Conversely, TerraMind’s localization performance benefits from the presence of multiple auxiliary modalities, but its generative quality for physical variables remains limited across settings. Full ablation results are provided in Appendix~E.

\subsection{Stochastic Benchmark}
\label{sec:stochastic}

The oracle evaluation protocol (Section~\ref{sec:quantitative}) measures the best single sample a model can produce.
We now ask the complementary question: \emph{does the model's output distribution match the true distribution of plausible observations?}
Following the setup in Section~\ref{sec:benchmark}, we compare 16 COP-GEN samples and 16 TerraMind samples per cell against 16 real Sentinel-2 L2A acquisitions spanning multiple years, across the 489 benchmark cells.

\subsubsection{Diversity Collapse}

The same pattern emerges across all three embedding spaces: TerraMind exhibits near-complete \emph{diversity collapse}, with its 16 per-cell samples being almost identical (thus occupying only a narrow slice of the real observation space), whereas COP-GEN produces samples that span most of the real distribution. Table~\ref{tab:stochastic_metrics} summarises the key metrics.

\begin{table}[t]
\centering
\resizebox{\linewidth}{!}{%
\begin{tabular}{llcc}
\toprule
\textbf{Stream} & \textbf{Metric} & \textbf{COP-GEN} & \textbf{TerraMind} \\
\midrule
\multirow{4}{*}{\rotatebox{90}{\scriptsize Spectral}}
& 1-NN accuracy $\downarrow$          & \textbf{0.911 $\pm$ 0.075} & 0.985 $\pm$ 0.027 \\
& Precision ($k{=}5$) $\uparrow$      & 0.289 $\pm$ 0.348          & \textbf{0.483 $\pm$ 0.469} \\
& Recall ($k{=}5$) $\uparrow$         & \textbf{0.900 $\pm$ 0.264} & 0.028 $\pm$ 0.080 \\
& Intra-set distance                   & \textbf{0.455 $\pm$ 0.155} & 0.050 $\pm$ 0.015 \\
\midrule
\multirow{4}{*}{\rotatebox{90}{\scriptsize RGB}}
& 1-NN accuracy $\downarrow$          & \textbf{0.982 $\pm$ 0.031} & 0.998 $\pm$ 0.009 \\
& Precision ($k{=}5$) $\uparrow$      & 0.086 $\pm$ 0.220          & \textbf{0.119 $\pm$ 0.286} \\
& Recall ($k{=}5$) $\uparrow$         & \textbf{0.726 $\pm$ 0.370} & 0.001 $\pm$ 0.013 \\
& Intra-set distance                   & \textbf{13.39 $\pm$ 1.53}  & 5.65 $\pm$ 0.76 \\
\midrule
\multirow{1}{*}{\scriptsize LPIPS}
& Intra-set distance                   & \textbf{0.470 $\pm$ 0.046} & 0.287 $\pm$ 0.058 \\
\midrule
\multirow{4}{*}{\rotatebox{90}{\scriptsize Physical}}
& MMD $\downarrow$                     & \textbf{0.589 $\pm$ 0.287} & 1.149 $\pm$ 0.423 \\
& Wasserstein (mean, 12 bands) $\downarrow$ & 0.143 $\pm$ 0.109    & \textbf{0.117 $\pm$ 0.146} \\
& Spectral coverage $\uparrow$        & \textbf{0.629 $\pm$ 0.291} & 0.180 $\pm$ 0.144 \\
\bottomrule
\end{tabular}%
}
\caption{Stochastic benchmark: COP-GEN vs.\ TerraMind across 489 cells, with 16 samples per cell for all three sources (real, COP-GEN, TerraMind). Mean $\pm$ std reported. For 1-NN accuracy and MMD, lower is better ($\downarrow$); for Precision, Recall, and Coverage, higher is better ($\uparrow$). Intra-set distance is a diversity measure (higher indicates more diverse sets). Best values in \textbf{bold}.}
\label{tab:stochastic_metrics}
\end{table}

\begin{itemize}[leftmargin=2em]
    \item \textbf{Recall.}
    COP-GEN achieves spectral recall of 0.900, meaning its generated samples cover 90\% of the real observation manifold. TerraMind achieves 0.028---its samples are effectively invisible to the real distribution. This pattern holds in RGB feature space (72.6\% vs.\ 0.1\%).

    \item \textbf{Intra-set diversity.} COP-GEN is 9.1$\times$ more diverse than TerraMind in spectral space (0.455 vs.\ 0.050), 2.4$\times$ in ResNet-50 RGB space, and 1.6$\times$ in LPIPS perceptual space. The real observations have spectral intra-set distance 0.214, placing them between COP-GEN and TerraMind.

    \item \textbf{Spectral coverage.}
    COP-GEN spans 63\% of the real per-band reflectance range; TerraMind spans only 18\%. This confirms that TerraMind collapses to a narrow spectral region near the conditional mean.

    \item \textbf{MMD.}
    COP-GEN's overall stochastic distance to real is approximately half that of TerraMind (0.589 vs.\ 1.149), indicating a closer match to the true observation distribution.
\end{itemize}

TerraMind's advantages lie in metrics that reward proximity to the distribution centre:
per-band Wasserstein distance (0.117 vs.\ 0.143) and spectral precision (0.483 vs.\ 0.289).
These reflect the fact that each TerraMind sample lands near the densest region of the real manifold---a consequence of its near-deterministic output, rather than from superior stochastic modelling.

\subsubsection{The Realism--Diversity Trade-off}
\label{sec:realism_diversity}

These results reveal a clear trade-off.
TerraMind achieves higher per-sample precision (each output is close to a plausible real image) but at the cost of near-zero recall (failing to represent the range of valid observations
).
COP-GEN achieves high recall and coverage at the cost of lower precision, with some samples falling outside the empirical real manifold.

Notably, COP-GEN's intra-set diversity (0.455) exceeds that of the real observations (0.214). This is partly expected, since 16 cloud-filtered acquisitions may not capture the full range of plausible surface states, but it also indicates that some COP-GEN samples explore beyond the physically realistic manifold (spectral precision = 0.289). Reducing over-exploration while maintaining high recall is a natural target for future work.

The same pattern holds across all three embedding representations (spectral, ResNet-50, LPIPS), indicating that TerraMind's diversity collapse is not an artefact of a particular feature space but a fundamental property of its near-deterministic generation process.

\section{Discussion}

COP-GEN complements existing EO foundation models by explicitly modeling the one-to-many nature of cross-modal mappings. By capturing distributions of physically plausible outcomes, COP-GEN is well-suited for tasks with sparse or incomplete inputs, band infilling, and scenario generation. In contrast, deterministic models can yield strong performance on single-image metrics and geolocation tasks, but often at the cost of reduced output diversity and bias toward conditional mean prediction. From this perspective, COP-GEN should be viewed as complementary rather than a competing paradigm: it models distributions of possibilities, while other approaches optimise deterministic reconstructions.

Architecturally, decoupling modalities (and spectral band groups) via resolution-aware tokenisation and independent diffusion timesteps enables COP-GEN to perform zero-shot any-to-any modality translation, band infilling, and flexible conditioning without retraining or task-specific heads.

Our results also highlight limitations of standard EO evaluation metrics for generative tasks. Conventional pointwise metrics and single-reference benchmarks are poorly aligned with stochastic generative modeling. Peak-capability evaluation and distributional analyses better reflect a model’s ability to represent variability and distributional support.

\paragraph{Limitations and Future Work.}
Despite these strengths, COP-GEN exhibits several limitations. First, geolocation (latitude-longitude) and timestamp conditioning currenctly have a limited influence on the generated outputs.
In practice, generations conditioned on identical input modalities but different lat–lon or timestamps often remain visually similar, even when the conditioning metadata differs substantially. This is most evident with snow appearing in near-equatorial or tropical regions despite geolocation conditioning. A likely contributing factor is imbalance in the training objective: lat–lon and timestamp are represented by far fewer tokens than spatial modalities, and thus may contribute weakly to the overall diffusion loss. Future work could address this through improved loss balancing, modality-specific weighting, or alternative conditioning mechanisms.

Second, the current diffusion training scheme focuses on joint unconditional generation, where all modalities are always generated together. While this encourages strong learning of the joint distribution, it may limit separability and marginal understanding of individual modalities. A promising direction is to introduce stochastic modality dropout during training, where only random subsets of modalities are generated at each step. Such a scheme could improve conditional robustness, disentangle modality dependencies, and yield cleaner marginal distributions.

In addition to the aforementioned limitations, several future work directions emerge from this work: (i) building hybrid systems that combine deterministic and stochastic models; (ii) exploration of distribution-aware metrics, temporal consistency measures, and multi-reference evaluation strategies to better capture generative performance in EO; (iii) extending COP-GEN to explicitly model temporal sequences would enable simulation of Earth system dynamics rather than static snapshots; (iv) scaling to higher resolutions and additional sensors.

\section{Conclusion}
\label{sec:conclusion}

This paper introduces COP-GEN, a stochastic, multimodal latent diffusion transformer designed to learn the joint distribution of heterogeneous Earth observation data at scale. By integrating optical, radar, elevation, land-cover, temporal, and geolocation information within a unified probabilistic framework, COP-GEN addresses a fundamental limitation of existing Earth observation models: their inability to represent the inherently non-injective and uncertain nature of cross-sensor mappings.

In contrast to deterministic approaches, COP-GEN captures one-to-many physical mappings that are intrinsic to remote sensing. Resolution-aware tokenisation, modality-specific latent encoders, and independent diffusion timesteps enable flexible any-to-any conditional generation, zero-shot modality translation, and spectral band infilling without task-specific retraining. This design preserves native sensor characteristics and avoids the aggressive resampling
commonly required by prior multimodal architectures.

Experimental results demonstrate strong peak generative fidelity across optical, radar, and elevation modalities, while qualitative analyses show that COP-GEN produces diverse, physically plausible realisations that respect topography, land cover, and spectral signatures. The model systematically narrows its output distributions as conditioning information increases, indicating that it learns to modulate conditional variability in a physically meaningful way rather than collapsing toward conditional means.

Beyond the proposed model, this work highlights a broader evaluation consideration: single-reference, pointwise metrics are poorly suited to stochastic generative modeling. Such metrics favor deterministic solutions and fail to capture a model’s ability to represent the variability inherent in underdetermined Earth Observation tasks. Our stochastic benchmark (Section~\ref{sec:stochastic}) makes this concrete: a deterministic-style generator can win per-sample precision (0.483 spectral) and per-band Wasserstein while achieving near-zero recall (0.028) and only 18\% spectral range coverage — a collapse that pointwise metrics hide entirely but that COP-GEN's stochastic design avoids (0.900 recall, 63\% coverage). Peak-capability and distribution-level analyses, consistent across independent spectral, ResNet-50, and LPIPS representations, provide a more informative assessment of generative performance and are therefore essential in future EO benchmarks.

Future work will focus on extending COP-GEN to explicitly model temporal sequences, enabling generative simulation of Earth system dynamics rather than static snapshots. Additional directions include scaling to higher spatial resolutions and additional sensors, developing distribution-aware evaluation metrics tailored to Earth observation, and exploring hybrid systems that combine deterministic predictors with stochastic generative components for distribution-aware inference.

By modeling conditional data distributions and their inherent variability, COP-GEN establishes a principled generative framework for multimodal Earth observation that better aligns with the physical reality of remote sensing data.

\section*{Acknowledgements}

Funding for this research is provided in part by the SENSE~-~Centre for Satellite Data in Environmental Science CDT studentship (NE/T00939X/1), and an EPSRC New Investigator Award (EP/X020703/1). This work used JASMIN, the UK’s collaborative data analysis environment~\url{https://jasmin.ac.uk} \cite{jasmin}.

\subsection*{Declaration of Generative AI}

During the preparation of this manuscript, the authors used generative AI tools to improve wording and clarity. All scientific content and interpretations were developed by the authors, who thoroughly reviewed and edited the manuscript and take full responsibility for its content.

{
    \small
    \bibliographystyle{ieeenat_fullname}
    \bibliography{main}

@String(CVPR= {IEEE Conf. Comput. Vis. Pattern Recog.})

@String(ECCV= {Eur. Conf. Comput. Vis.})

@String(ICLR = {Int. Conf. Learn. Represent.})

@String(AAAI = {AAAI})

@String(CVPR  = {CVPR})

@String(ECCV  = {ECCV})

@String(ICLR  = {ICLR})

@article{goodfellow2014gan,
  title={Generative adversarial nets},
  author={Goodfellow, Ian and Pouget-Abadie, Jean and Mirza, Mehdi and Xu, Bing and Warde-Farley, David and Ozair, Sherjil and Courville, Aaron and Bengio, Yoshua},
  journal={Advances in Neural Information Processing Systems},
  volume={27},
  year={2014}
}

@inproceedings{karras2019stylegan2,
  title={Analyzing and Improving the Image Quality of {StyleGAN}},
  author={Karras, Tero and Laine, Samuli and Aila, Timo},
  booktitle={IEEE/CVF Conference on Computer Vision and Pattern Recognition},
  year={2020}
}

@inproceedings{ramesh2021dalle,
  title={Zero-shot text-to-image generation},
  author={Ramesh, Aditya and Pavlov, Pavel and Goh, Gabriel and others},
  booktitle={International Conference on Machine Learning},
  year={2021}
}

@inproceedings{esser2021taming,
  title={Taming Transformers for High-Resolution Image Synthesis},
  author={Esser, Patrick and Rombach, Robin and Ommer, Björn},
  booktitle={IEEE/CVF Conference on Computer Vision and Pattern Recognition},
  year={2021}
}

@article{sohl2015deep,
  title={Deep Unsupervised Learning using Nonequilibrium Thermodynamics},
  author={Sohl-Dickstein, Jascha and Weiss, Eric and Maheswaranathan, Niru and Ganguli, Surya},
  journal={International Conference on Machine Learning},
  year={2015}
}

@inproceedings{ho2020ddpm,
  title={Denoising Diffusion Probabilistic Models},
  author={Ho, Jonathan and Jain, Ajay and Abbeel, Pieter},
  booktitle={Advances in Neural Information Processing Systems},
  year={2020}
}

@article{an2023efficient,
  title={Efficient remote sensing image super-resolution via lightweight diffusion models},
  author={An, Tai and Xue, Bin and Huo, Chunlei and Xiang, Shiming and Pan, Chunhong},
  journal={IEEE Geoscience and Remote Sensing Letters},
  year={2023},
  publisher={IEEE}
}

@article{jing2023denoising,
  title={Denoising diffusion probabilistic feature-based network for cloud removal in Sentinel-2 imagery},
  author={Jing, Ran and Duan, Fuzhou and Lu, Fengxian and Zhang, Miao and Zhao, Wenji},
  journal={Remote Sensing},
  volume={15},
  number={9},
  pages={2217},
  year={2023},
  publisher={MDPI}
}

@inproceedings{dhariwal2021diffusion,
  title={Diffusion Models Beat {GANs} on Image Synthesis},
  author={Dhariwal, Prafulla and Nichol, Alex},
  booktitle={Advances in Neural Information Processing Systems},
  year={2021}
}

@article{ronneberger2015unet,
  title={U-Net: Convolutional Networks for Biomedical Image Segmentation},
  author={Ronneberger, Olaf and Fischer, Philipp and Brox, Thomas},
  journal={International Conference on Medical Image Computing and Computer-Assisted Intervention},
  year={2015}
}

@inproceedings{rombach2022ldm,
  title={High-Resolution Image Synthesis with Latent Diffusion Models},
  author={Rombach, Robin and Blattmann, Andreas and Lorenz, Dominik and Esser, Patrick and Ommer, Björn},
  booktitle={IEEE/CVF Conference on Computer Vision and Pattern Recognition},
  year={2022}
}

@misc{saharia2022imagen,
      title={Photorealistic Text-to-Image Diffusion Models with Deep Language Understanding}, 
      author={Chitwan Saharia and William Chan and Saurabh Saxena and Lala Li and Jay Whang and Emily Denton and Seyed Kamyar Seyed Ghasemipour and Burcu Karagol Ayan and S. Sara Mahdavi and Rapha Gontijo Lopes and Tim Salimans and Jonathan Ho and David J Fleet and Mohammad Norouzi},
      year={2022},
      eprint={2205.11487},
      archivePrefix={arXiv},
      primaryClass={cs.CV},
      url={https://arxiv.org/abs/2205.11487}, 
}

@inproceedings{peebles2023dit,
  title={Scalable diffusion models with transformers},
  author={Peebles, William and Xie, Saining},
  booktitle={Proceedings of the IEEE/CVF international conference on computer vision},
  pages={4195--4205},
  year={2023}
}

@inproceedings{bao2023uvit,
  title={All are worth words: A vit backbone for diffusion models},
  author={Bao, Fan and Nie, Shen and Xue, Kaiwen and Cao, Yue and Li, Chongxuan and Su, Hang and Zhu, Jun},
  booktitle={Proceedings of the IEEE/CVF conference on computer vision and pattern recognition},
  pages={22669--22679},
  year={2023}
}

@inproceedings{gao2023masked,
  title={Masked diffusion transformer is a strong image synthesizer},
  author={Gao, Shanghua and Zhou, Pan and Cheng, Ming-Ming and Yan, Shuicheng},
  booktitle={Proceedings of the IEEE/CVF international conference on computer vision},
  pages={23164--23173},
  year={2023}
}

@inproceedings{bao2023one,
  title={One transformer fits all distributions in multi-modal diffusion at scale},
  author={Bao, Fan and Nie, Shen and Xue, Kaiwen and Li, Chongxuan and Pu, Shi and Wang, Yaole and Yue, Gang and Cao, Yue and Su, Hang and Zhu, Jun},
  booktitle={International Conference on Machine Learning},
  pages={1692--1717},
  year={2023},
  organization={PMLR}
}

@article{meraner2020cloud,
  title={Cloud removal in Sentinel-2 imagery using a deep residual neural network and SAR-optical data fusion},
  author={Meraner, Andrea and Ebel, Patrick and Zhu, Xiao Xiang and Schmitt, Michael},
  journal={ISPRS Journal of Photogrammetry and Remote Sensing},
  volume={166},
  pages={333--346},
  year={2020},
  publisher={Elsevier}
}

@article{liu2022generative,
  title={A generative adversarial network for pixel-scale lunar DEM generation from high-resolution monocular imagery and low-resolution DEM},
  author={Liu, Yang and Wang, Yexin and Di, Kaichang and Peng, Man and Wan, Wenhui and Liu, Zhaoqin},
  journal={Remote Sensing},
  volume={14},
  number={21},
  pages={5420},
  year={2022},
  publisher={MDPI}
}

@article{paola2023correction,
  title={Correction of banding errors in satellite images with generative adversarial networks (gan)},
  author={Paola, Z{\'a}rate L and Jes{\'u}s, L{\'o}pez S and Christian, Arroyo H and Sonia, Rinc{\'o}n U},
  journal={IEEE Access},
  volume={11},
  pages={51960--51970},
  year={2023},
  publisher={IEEE}
}

@article{jin2024hya,
  title={Hya-gan: remote sensing image cloud removal based on hybrid attention generation adversarial network},
  author={Jin, Minghao and Wang, Pengwei and Li, Yusong},
  journal={International Journal of Remote Sensing},
  volume={45},
  number={6},
  pages={1755--1773},
  year={2024},
  publisher={Taylor \& Francis}
}

@inproceedings{wang2018esrgan,
  title={Esrgan: Enhanced super-resolution generative adversarial networks},
  author={Wang, Xintao and Yu, Ke and Wu, Shixiang and Gu, Jinjin and Liu, Yihao and Dong, Chao and Qiao, Yu and Change Loy, Chen},
  booktitle={Proceedings of the European conference on computer vision (ECCV) workshops},
  pages={0--0},
  year={2018}
}

@inproceedings{dong2020remote,
  title={Remote sensing image super-resolution via enhanced back-projection networks},
  author={Dong, Xiaoyu and Xi, Zhihong and Sun, Xu and Yang, Lina},
  booktitle={IGARSS 2020-2020 IEEE International Geoscience and Remote Sensing Symposium},
  pages={1480--1483},
  year={2020},
  organization={IEEE}
}

@article{ma2020pan,
  title={Pan-GAN: An unsupervised pan-sharpening method for remote sensing image fusion},
  author={Ma, Jiayi and Yu, Wei and Chen, Chen and Liang, Pengwei and Guo, Xiaojie and Jiang, Junjun},
  journal={Information Fusion},
  volume={62},
  pages={110--120},
  year={2020},
  publisher={Elsevier}
}

@article{wang2019sar,
  title={SAR-to-optical image translation using supervised cycle-consistent adversarial networks},
  author={Wang, Lei and Xu, Xin and Yu, Yue and Yang, Rui and Gui, Rong and Xu, Zhaozhuo and Pu, Fangling},
  journal={Ieee Access},
  volume={7},
  pages={129136--129149},
  year={2019},
  publisher={IEEE}
}

@article{khanna2023diffusionsat,
  title={Diffusionsat: A generative foundation model for satellite imagery},
  author={Khanna, Samar and Liu, Patrick and Zhou, Linqi and Meng, Chenlin and Rombach, Robin and Burke, Marshall and Lobell, David and Ermon, Stefano},
  journal={arXiv preprint arXiv:2312.03606},
  year={2023}
}

@article{jia2025can,
  title={Can Generative Geospatial Diffusion Models Excel as Discriminative Geospatial Foundation Models?},
  author={Jia, Yuru and Marsocci, Valerio and Gong, Ziyang and Yang, Xue and Vergauwen, Maarten and Nascetti, Andrea},
  journal={arXiv preprint arXiv:2503.07890},
  year={2025}
}

@article{jakubik2025terramind,
  title={Terramind: Large-scale generative multimodality for earth observation},
  author={Jakubik, Johannes and Yang, Felix and Blumenstiel, Benedikt and Scheurer, Erik and Sedona, Rocco and Maurogiovanni, Stefano and Bosmans, Jente and Dionelis, Nikolaos and Marsocci, Valerio and Kopp, Niklas and others},
  journal={arXiv preprint arXiv:2504.11171},
  year={2025}
}

@article{olmoearth2024,
  title={OlmoEarth: Stable Latent Image Modeling for Multimodal Earth Observation},
  author={Herzog, Henry and Bastani, Favyen and Zhang, Yawen and Tseng, Gabriel and Redmon, Joseph and Sablon, Hadrien and Park, Ryan and Morrison, Jacob and Buraczynski, Alexandra and Farley, Karen and others},
  journal={arXiv preprint arXiv:2511.13655},
  year={2025}
}

@misc{copernicus_programme,
  title        = {Copernicus: {E}uropes Eyes on {E}arth},
  author       = {{European Commission}},
  year         = {2025},
  url          = {https://www.copernicus.eu},
  note         = {Accessed: 2024-12-30},
  organization = {European Union}
}

@inproceedings{
assel2025jointembedding,
title={Joint\nobreakdash-Embedding vs Reconstruction: Provable Benefits of Latent Space Prediction for Self\nobreakdash-Supervised Learning},
author={Hugues Van Assel and Mark Ibrahim and Tommaso Biancalani and Aviv Regev and Randall Balestriero},
booktitle={The Thirty-ninth Annual Conference on Neural Information Processing Systems},
year={2025},
url={https://openreview.net/forum?id=UOaLsgn5wb}
}

@misc{2025_tessera,
  author = {Zhengpeng Feng and Sadiq Jaffer and Jovana Knezevic and Silja Sormunen and Robin Young and Madeline Lisaius and Markus Immitzer and James Ball and Clement Atzberger and David A. Coomes and Anil Madhavapeddy and Andrew Blake and Srinivasan Keshav},
  doi = {10.48550/arXiv.2506.20380},
  month = {jun},
  title = {TESSERA: Temporal Embeddings of Surface Spectra for Earth Representation and Analysis},
  url = {http://arxiv.org/abs/2506.20380},
  year = {2025}}

@misc{wang2025unifiedcopernicusfoundationmodel,
      title={Towards a Unified Copernicus Foundation Model for Earth Vision}, 
      author={Yi Wang and Zhitong Xiong and Chenying Liu and Adam J. Stewart and Thomas Dujardin and Nikolaos Ioannis Bountos and Angelos Zavras and Franziska Gerken and Ioannis Papoutsis and Laura Leal-Taixé and Xiao Xiang Zhu},
      year={2025},
      eprint={2503.11849},
      archivePrefix={arXiv},
      primaryClass={cs.CV},
      url={https://arxiv.org/abs/2503.11849}, 
}

@inproceedings{waldmann2025panopticon,
  title={Panopticon: Advancing any-sensor foundation models for earth observation},
  author={Waldmann, Leonard and Shah, Ando and Wang, Yi and Lehmann, Nils and Stewart, Adam and Xiong, Zhitong and Zhu, Xiao Xiang and Bauer, Stefan and Chuang, John},
  booktitle={Proceedings of the Computer Vision and Pattern Recognition Conference},
  pages={2204--2214},
  year={2025}
}

@misc{brown2025alphaearthfoundationsembeddingfield,
title={AlphaEarth Foundations: An embedding field model for accurate and efficient global mapping from sparse label data}, 
author={Christopher F. Brown and Michal R. Kazmierski and Valerie J. Pasquarella and William J. Rucklidge and Masha Samsikova and Chenhui Zhang and Evan Shelhamer and Estefania Lahera and Olivia Wiles and Simon Ilyushchenko and Noel Gorelick and Lihui Lydia Zhang and Sophia Alj and Emily Schechter and Sean Askay and Oliver Guinan and Rebecca Moore and Alexis Boukouvalas and Pushmeet Kohli},
year={2025},
eprint={2507.22291},
archivePrefix={arXiv},
primaryClass={cs.CV},
}

@article{xiong2024dofa,
  title={Neural Plasticity-Inspired Foundation Model for Observing the {Earth} Crossing Modalities},
  author={Xiong, Zhitong and Wang, Yi and Zhang, Fahong and Stewart, Adam J and Hanna, Jo{\"e}lle and Borth, Damian and Papoutsis, Ioannis and Saux, Bertrand Le and Camps-Valls, Gustau and Zhu, Xiao Xiang},
  journal={arXiv preprint arXiv:2403.15356},
  year={2024}
}

@misc{OlmoEarthpretrain,
  author = {Zhang, Y. and Tseng, G. and Redmon, J. and Herzog, H. and Bastani, F. and Sablon, H. and Park, R. and Morrison, J. and Buraczynski, A. and Farley, K. and Hansen, J. and Howe, A. and Johnson, P. and Otterlee, M. and Pitelka, H. and Ratner, R. and Schmitt, T. and Wilhelm, C. and Wood, S. and Jacobi, M. and Kerner, H. and Shelhamer, E. and Farhadi, A. and Krishna, R. and Beukema, P.},
  title = {{OlmoEarth}: Earth Observation Foundation Model}, 
  howpublished = {\url{https://www.pjreddie.com/static/papers/OlmoEarth.pdf}},
  year = {2025},
  note = {Accessed: 2025-11-11}
}

@inproceedings{nedungadi2024mmearth,
  title={MMEarth: Exploring multi-modal pretext tasks for geospatial representation learning},
  author={Nedungadi, Vishal and Kariryaa, Ankit and Oehmcke, Stefan and Belongie, Serge and Igel, Christian and Lang, Nico},
  booktitle={European Conference on Computer Vision},
  pages={164--182},
  year={2024},
  organization={Springer}
}

@article{astruc2024omnisat,
  title={Omni{S}at: {S}elf-Supervised Modality Fusion for {E}arth Observation},
  author={Astruc, Guillaume and Gonthier, Nicolas and Mallet, Clement and Landrieu, Loic},
  journal={ECCV},
  year={2024}
}

@misc{szwarcman2025prithvieo20versatilemultitemporalfoundation,
      title={Prithvi-EO-2.0: A Versatile Multi-Temporal Foundation Model for Earth Observation Applications}, 
      author={Daniela Szwarcman et al.},
      year={2025},
      eprint={2412.02732},
      archivePrefix={arXiv},
      primaryClass={cs.CV},
      url={https://arxiv.org/abs/2412.02732}, 
}

@inproceedings{tseng2025galileo,
title={Galileo: Learning Global \& Local Features of Many Remote Sensing Modalities},
author={Gabriel Tseng and Anthony Fuller and Marlena Reil and Henry Herzog and Patrick Beukema and Favyen Bastani and James R Green and Evan Shelhamer and Hannah Kerner and David Rolnick},
booktitle={Forty-second International Conference on Machine Learning},
year={2025},
url={https://openreview.net/forum?id=gqZO3eSZRy}
}

@article{danish2025terrafmscalablefoundationmodel,
      title={TerraFM: A Scalable Foundation Model for Unified Multisensor Earth Observation}, 
      author={Muhammad Sohail Danish and Muhammad Akhtar Munir and Syed Roshaan Ali Shah and Muhammad Haris Khan and Rao Muhammad Anwer and Jorma Laaksonen and Fahad Shahbaz Khan and Salman Khan},
      year={2025},
      eprint={2506.06281},
      archivePrefix={arXiv},
      primaryClass={cs.CV},
      url={https://arxiv.org/abs/2506.06281}, 
}

@inproceedings{lopez2017revisiting,
  title     = {Revisiting Classifier Two-Sample Tests},
  author    = {Lopez-Paz, David and Oquab, Maxime},
  booktitle = {International Conference on Learning Representations (ICLR)},
  year      = {2017},
}

@inproceedings{kynkaanniemi2019improved,
  title     = {Improved Precision and Recall Metric for Assessing Generative Models},
  author    = {Kynk{\"a}{\"a}nniemi, Tuomas and Karras, Tero and Laine, Samuli and Lehtinen, Jaakko and Aila, Timo},
  booktitle = {Advances in Neural Information Processing Systems (NeurIPS)},
  year      = {2019},
}

@article{liu2024diffusion,
  title={Diffusion models meet remote sensing: Principles, methods, and perspectives},
  author={Liu, Yidan and Yue, Jun and Xia, Shaobo and Ghamisi, Pedram and Xie, Weiying and Fang, Leyuan},
  journal={arXiv preprint arXiv:2404.08926},
  year={2024}
}

@ARTICLE{metaearth,
  author={Yu, Zhiping and Liu, Chenyang and Liu, Liqin and Shi, Zhenwei and Zou, Zhengxia},
  journal={IEEE Transactions on Pattern Analysis and Machine Intelligence}, 
  title={MetaEarth: A Generative Foundation Model for Global-Scale Remote Sensing Image Generation}, 
  year={2025},
  volume={47},
  number={3},
  pages={1764-1781},
}

@article{zheng2024changen2,
  title={Changen2: Multi-temporal remote sensing generative change foundation model},
  author={Zheng, Zhuo and Ermon, Stefano and Kim, Dongjun and Zhang, Liangpei and Zhong, Yanfei},
  journal={IEEE Transactions on Pattern Analysis and Machine Intelligence},
  year={2024},
  publisher={IEEE}
}

@article{tang2024crs,
  title={Crs-diff: Controllable remote sensing image generation with diffusion model},
  author={Tang, Datao and Cao, Xiangyong and Hou, Xingsong and Jiang, Zhongyuan and Liu, Junmin and Meng, Deyu},
  journal={IEEE Transactions on Geoscience and Remote Sensing},
  year={2024},
  publisher={IEEE}
}

@inproceedings{toker2024satsynth,
  title={Satsynth: Augmenting image-mask pairs through diffusion models for aerial semantic segmentation},
  author={Toker, Aysim and Eisenberger, Marvin and Cremers, Daniel and Leal-Taix{\'e}, Laura},
  booktitle={Proceedings of the IEEE/CVF Conference on Computer Vision and Pattern Recognition},
  pages={27695--27705},
  year={2024}
}

@inproceedings{espinosa2025cop,
  title={COP-GEN-Beta: Unified Generative Modelling of COPernicus Imagery Thumbnails},
  author={Espinosa, Miguel and Marsocci, Valerio and Jia, Yuru and Crowley, Elliot and Czerkawski, Mikolaj},
  booktitle={Proceedings of the Computer Vision and Pattern Recognition Conference},
  year={2025}
}

@article{he2023tdiffde,
  title={Tdiffde: A truncated diffusion model for remote sensing hyperspectral image denoising},
  author={He, Jiang and Li, Yajie and Yuan, Qiangqiang and others},
  journal={arXiv preprint arXiv:2311.13622},
  year={2023}
}

@inproceedings{pang2024hir,
  title={HIR-Diff: Unsupervised Hyperspectral Image Restoration Via Improved Diffusion Models},
  author={Pang, Li and Rui, Xiangyu and Cui, Long and Wang, Hongzhong and Meng, Deyu and Cao, Xiangyong},
  booktitle={Proceedings of the IEEE/CVF Conference on Computer Vision and Pattern Recognition},
  pages={3005--3014},
  year={2024}
}

@article{wang2024idf,
  title={IDF-CR: Iterative Diffusion Process for Divide-and-Conquer Cloud Removal in Remote-sensing Images},
  author={Wang, Meilin and Song, Yexing and Wei, Pengxu and Xian, Xiaoyu and Shi, Yukai and Lin, Liang},
  journal={IEEE Transactions on Geoscience and Remote Sensing},
  year={2024},
  publisher={IEEE}
}

@article{zou2024diffcr,
  title={DiffCR: A Fast Conditional Diffusion Framework for Cloud Removal From Optical Satellite Images},
  author={Zou, Xuechao and Li, Kai and Xing, Junliang and Zhang, Yu and Wang, Shiying and Jin, Lei and Tao, Pin},
  journal={IEEE Transactions on Geoscience and Remote Sensing},
  volume={62},
  pages={1--14},
  year={2024},
  publisher={IEEE}
}

@article{wang2025semantic,
  title={Semantic guided large scale factor remote sensing image super-resolution with generative diffusion prior},
  author={Wang, Ce and Sun, Wanjie},
  journal={ISPRS Journal of Photogrammetry and Remote Sensing},
  volume={220},
  pages={125--138},
  year={2025},
  publisher={Elsevier}
}

@inproceedings{dong2024building,
  title={Building Bridges across Spatial and Temporal Resolutions: Reference-Based Super-Resolution via Change Priors and Conditional Diffusion Model},
  author={Dong, Runmin and Yuan, Shuai and Luo, Bin and Chen, Mengxuan and Zhang, Jinxiao and Zhang, Lixian and Li, Weijia and Zheng, Juepeng and Fu, Haohuan},
  booktitle={Proceedings of the IEEE/CVF Conference on Computer Vision and Pattern Recognition},
  pages={27684--27694},
  year={2024}
}

@inproceedings{le2024detecting,
  title={Detecting Out-Of-Distribution Earth Observation Images with Diffusion Models},
  author={Le Bellier, Georges and Audebert, Nicolas},
  booktitle={Proceedings of the IEEE/CVF Conference on Computer Vision and Pattern Recognition},
  pages={481--491},
  year={2024}
}

@article{amit2021segdiff,
  title={Segdiff: Image segmentation with diffusion probabilistic models},
  author={Amit, Tomer and Shaharbany, Tal and Nachmani, Eliya and Wolf, Lior},
  journal={arXiv preprint arXiv:2112.00390},
  year={2021}
}

@inproceedings{kolbeinsson2024multi,
  title={Multi-class segmentation from aerial views using recursive noise diffusion},
  author={Kolbeinsson, Benedikt and Mikolajczyk, Krystian},
  booktitle={Proceedings of the IEEE/CVF Winter Conference on Applications of Computer Vision},
  pages={8439--8449},
  year={2024}
}

@article{zhou2024exploring,
  title={Exploring Multi-Timestep Multi-Stage Diffusion Features for Hyperspectral Image Classification},
  author={Zhou, Jingyi and Sheng, Jiamu and Ye, Peng and Fan, Jiayuan and He, Tong and Wang, Bin and Chen, Tao},
  journal={IEEE Transactions on Geoscience and Remote Sensing},
  year={2024},
  publisher={IEEE}
}

@inproceedings{li2024mdfl,
  title={MDFL: Multi-domain diffusion-driven feature learning},
  author={Li, Daixun and Xie, Weiying and Zhang, Jiaqing and Li, Yunsong},
  booktitle={Proceedings of the AAAI conference on artificial intelligence},
  year={2024}
}

@inproceedings{qu2024lds2ae,
  title={LDS2AE: Local Diffusion Shared-Specific Autoencoder for Multimodal Remote Sensing Image Classification with Arbitrary Missing Modalities},
  author={Qu, Jiahui and Yang, Yuanbo and Dong, Wenqian and Yang, Yufei},
  booktitle={Proceedings of the AAAI Conference on Artificial Intelligence},
  volume={38},
  number={13},
  pages={14731--14739},
  year={2024}
}

@article{wen2024gcd,
  title={GCD-DDPM: A generative change detection model based on difference-feature guided DDPM},
  author={Wen, Yihan and Ma, Xianping and Zhang, Xiaokang and Pun, Man-On},
  journal={IEEE Transactions on Geoscience and Remote Sensing},
  year={2024},
  publisher={IEEE}
}

@article{tian2024swimdiff,
  title={SwiMDiff: Scene-wide Matching Contrastive Learning with Diffusion Constraint for Remote Sensing Image},
  author={Tian, Jiayuan and Lei, Jie and Zhang, Jiaqing and Xie, Weiying and Li, Yunsong},
  journal={IEEE Transactions on Geoscience and Remote Sensing},
  year={2024},
  publisher={IEEE}
}

@article{zhang2023diffucd,
  title={Diffucd: Unsupervised hyperspectral image change detection with semantic correlation diffusion model},
  author={Zhang, Xiangrong and Tian, Shunli and Wang, Guanchun and Zhou, Huiyu and Jiao, Licheng},
  journal={arXiv preprint arXiv:2305.12410},
  year={2023}
}

@article{jia2024siamese,
  title={Siamese Meets Diffusion Network: SMDNet for Enhanced Change Detection in High-Resolution RS Imagery},
  author={Jia, Jia and Lee, Geunho and Wang, Zhibo and Zhi, Lyu and He, Yuchu},
  journal={IEEE Journal of Selected Topics in Applied Earth Observations and Remote Sensing},
  year={2024},
  publisher={IEEE}
}

@article{sigger2024unveiling,
  title={Unveiling the potential of diffusion model-based framework with transformer for hyperspectral image classification},
  author={Sigger, Neetu and Vien, Quoc-Tuan and Nguyen, Sinh Van and Tozzi, Gianluca and Nguyen, Tuan Thanh},
  journal={Scientific Reports},
  volume={14},
  number={1},
  pages={8438},
  year={2024},
  publisher={Nature Publishing Group UK London}
}

@article{bandara2022ddpm,
  title={DDPM-CD: Denoising diffusion probabilistic models as feature extractors for change detection},
  author={Bandara, Wele Gedara Chaminda and Nair, Nithin Gopalakrishnan and Patel, Vishal M},
  journal={arXiv preprint arXiv:2206.11892},
  year={2022}
}

@inproceedings{jasmin,
  title={Storing and manipulating environmental big data with JASMIN},
  author={Lawrence, Bryan N. and Bennett, Victoria L. and Churchill, James and Juckes, Martin and Kershaw, Philip and Pascoe, Stephen and Pepler, Sam and Pritchard, Matthew and Stephens, Ag},
  booktitle={IEEE Big Data},
  year={2013},
  month={October},
  pages={1--5},
  organization={IEEE},
  address={San Francisco}
}

@inproceedings{francis2024major,
  title={Major tom: Expandable datasets for earth observation},
  author={Francis, Alistair and Czerkawski, Mikolaj},
  booktitle={IGARSS 2024-2024 IEEE International Geoscience and Remote Sensing Symposium},
  pages={2935--2940},
  year={2024},
  organization={IEEE}
}

@misc{impact_observatory_lulc_2023,
  author       = {{Impact Observatory} and {Microsoft} and {Esri}},
  title        = {10m Annual Land Use Land Cover (9-class) V2},
  year         = {2023},
  howpublished = {Microsoft Planetary Computer},
  url          = {https://planetarycomputer.microsoft.com/dataset/io-lulc-annual-v02},
  note         = {Accessed: 2026-05-11. Temporal extent: 2017--2023. 
                  Derived from ESA Sentinel-2 imagery at 10\,m resolution. 
                  Licensed under CC BY 4.0.
                  STAC collection: \url{https://planetarycomputer.microsoft.com/api/stac/v1/collections/io-lulc-annual-v02}},
  license      = {CC BY 4.0},
}

@article{espinosa_2023_8_mapsat,
  	author = {Miguel Espinosa and Elliot J. Crowley},
  	title = {Generate Your Own Scotland: Satellite Image Generation Conditioned on Maps},
  	year = {2023},
  	month = {Aug},
  	journal = {NeurIPS 2023 Workshop on Diffusion Models},
  	institution = {University of Edinburgh},
  	url = {https://arxiv.org/abs/2308.16648},
}

@inproceedings{mesa2025,
title={MESA: Text-Driven Terrain Generation Using Latent Diffusion and Global Copernicus Data},
author={Paul Borne--Pons and Mikolaj Czerkawski and Rosalie Martin and Romain Rouffet},
year={2025},
booktitle={MORSE Workshop at CVPR 2025},
eprint={2504.07210},
url={https://arxiv.org/abs/2504.07210},}
}

\clearpage

\onecolumn

\appendix
\section{Supplementary Material}
\label{sec:supplementary}

We provide additional qualitative, quantitative, and architectural results that support and extend the findings presented in the main paper. The material is organised thematically to allow readers to quickly locate specific analyses, visualisations, or ablations of interest.

\paragraph{Output diversity and conditional variability.}
Additional qualitative visualisations illustrating output diversity and conditional variability for DEM--LULC to Sentinel-2 generation are shown in Figures~\ref{fig:DEM_LULC_to_S2L2A_v1}, and~\ref{fig:DEM_LULC_to_S2L2A_v2}. These examples demonstrate how COP-GEN generates diverse yet physically plausible optical scenes under identical conditioning, highlighting its ability to model one-to-many relationships.

\paragraph{Spectral fidelity across land-cover classes.}
Extended qualitative results analysing spectral fidelity across different land-cover classes are shown in Figures~\ref{fig:560U_34R_spectral_profile}, \ref{fig:502U_263R_spectral_profile}, and \ref{fig:256U_1125L_spectral_profile}. These figures report per-band spectral profiles stratified by land-cover category, illustrating how COP-GEN preserves characteristic spectral signatures while maintaining realistic variability across samples.

\paragraph{Extended quantitative evaluation.}
Complete quantitative results are provided for all experimental settings, including both full-context and leave-one-out evaluations. Tile-level and experiment-level metrics are reported in Tables~\ref{tab:tile_level_detailed}, \ref{tab:experiment_level_detailed}, \ref{tab:leave_one_out_full_tile_level}, and \ref{tab:leave_one_out_full_experiment_level}. These tables complement the main paper by exposing the full distribution of performance across modalities, conditioning setups, and evaluation protocols.

\paragraph{Model size and parameter count.}
A comparison of backbone parameter counts between COP-GEN and TerraMind (Base) is provided in Table~\ref{tab:model_parameters}. Despite supporting stochastic joint generation across heterogeneous modalities, native sensor resolutions, and any-to-any conditional sampling, COP-GEN remains comparable in scale to existing multimodal Earth observation foundation models. This highlights that the increased generative flexibility of COP-GEN arises primarily from architectural design choices.

\paragraph{Global spatial uncertainty and geolocation distributions.}
Global spatial uncertainty and multimodal geolocation predictions are visualised through latitude--longitude distributions in Figures~\ref{fig:lat-lon-215U_1019L}, \ref{fig:lat-lon-95U_112R}, \ref{fig:lat-lon-250U_409R}, and \ref{fig:lat-lon-211D_500R}. These figures illustrate the breadth and multimodality of spatial predictions inferred from under-specified conditioning, reflecting learned global geospatial priors.

\paragraph{Distributional behaviour under increasing conditioning.}
Per-band reflectance distributions under progressively stronger conditioning are shown in Figures~\ref{fig:195D_669L_distribution}, \ref{fig:248U_978R_distribution}, \ref{fig:250U_409R_distribution}, and \ref{fig:143D_1481R_distribution}. These analyses demonstrate how COP-GEN systematically narrows its output distributions as additional modalities are provided, indicating uncertainty reduction consistent with physical constraints.

\paragraph{Class-conditional geolocation priors.}
Class-conditional geolocation priors inferred from individual land-cover classes are visualised in Figures~\ref{fig:lat_lon_comparison_trees} (trees), \ref{fig:lat_lon_comparison_snow_ice} (snow/ice), \ref{fig:lat_lon_comparison_rangeland} (rangeland), \ref{fig:lat_lon_comparison_flooded_vegetation} (flooded vegetation), \ref{fig:lat_lon_comparison_crops} (crops), \ref{fig:lat_lon_comparison_clouds} (clouds), \ref{fig:lat_lon_comparison_built_area} (built-up area), and \ref{fig:lat_lon_comparison_bare_ground} (bare ground). These figures show that COP-GEN captures meaningful global geographic priors associated with different land-cover types.

\paragraph{}Together, these supplementary results provide deeper insight into the distributional behaviour, physical plausibility, and scalability of COP-GEN beyond what can be presented in the main paper.

{\begin{figure*}
    \vspace{-2cm}
    \centering
    \includegraphics[width=\linewidth]{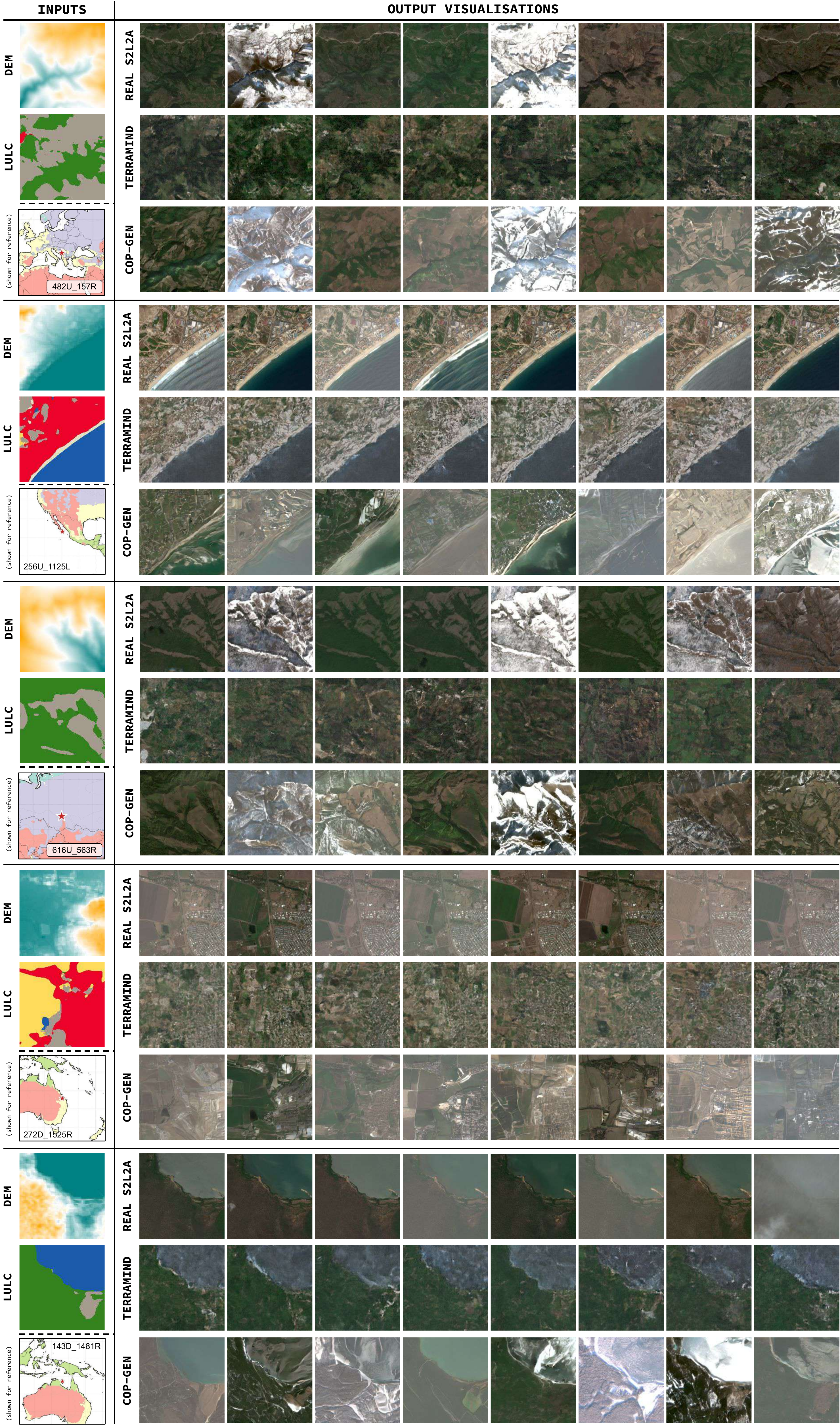}
    \caption{\textbf{Multiple conditional Sentinel-2 L2A generations from DEM and LULC inputs.} Generated outputs per scene demonstrates COP-GEN’s ability to model multimodal variability in spectral response, illumination, and atmospheric conditions, while maintaining consistency with underlying terrain and land-cover information.}
    \label{fig:DEM_LULC_to_S2L2A_v2}
\end{figure*}
}
{\begin{figure*}
    \vspace{-2cm}
    \centering
    \includegraphics[width=\linewidth]{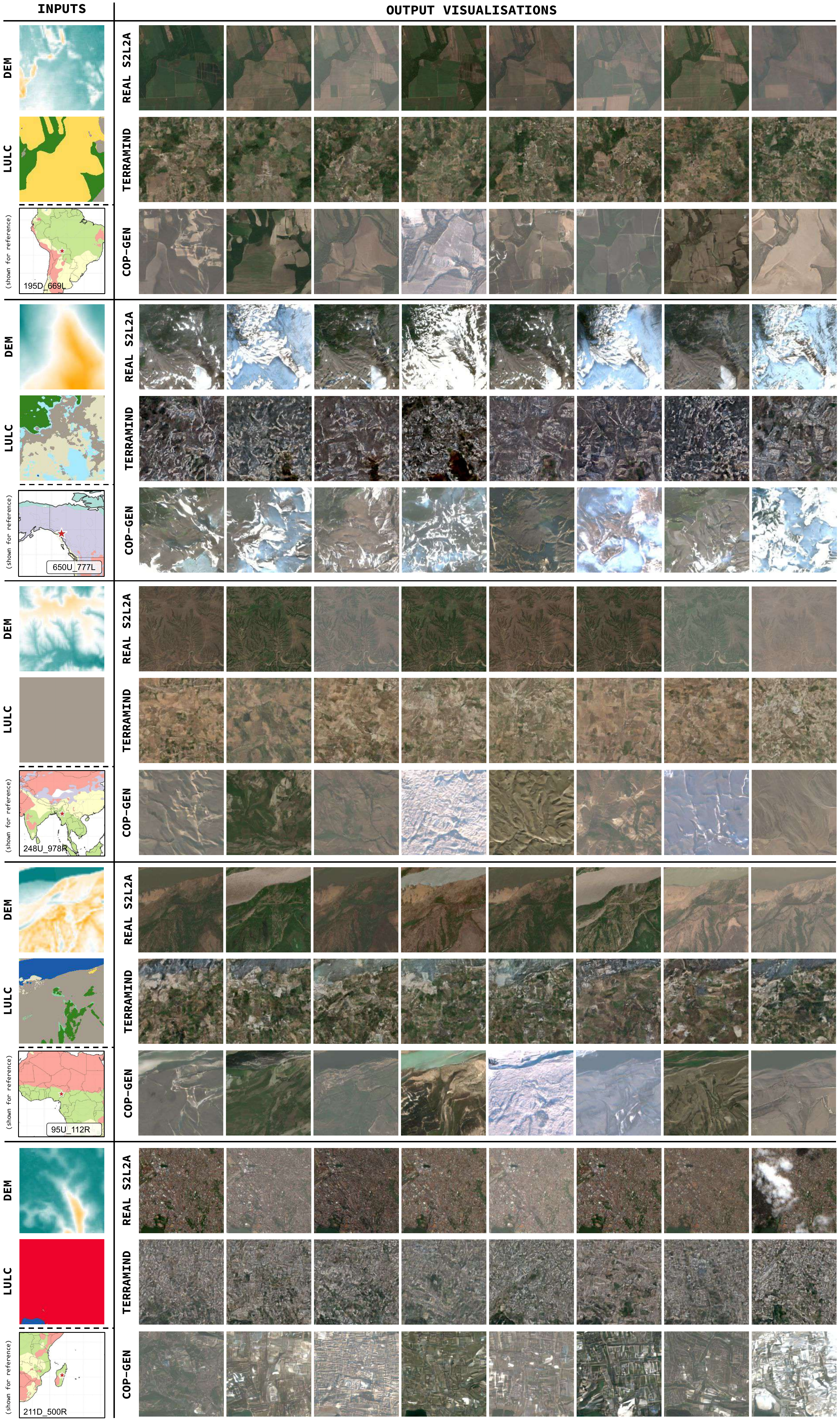}
    \caption{\textbf{Conditional generation of Sentinel-2 L2A imagery from DEM and LULC inputs.} For each location, a grid of generated examples illustrates the diversity of COP-GEN outputs under varying spectral, illumination, and atmospheric conditions, while respecting topographic and land-cover constraints. This demonstrates the model’s ability to represent one-to-many relationships in multimodal Earth observation.}
    \label{fig:DEM_LULC_to_S2L2A_v1}
\end{figure*}
}

{\begin{figure*}
    \centering
    \includegraphics[width=\linewidth]{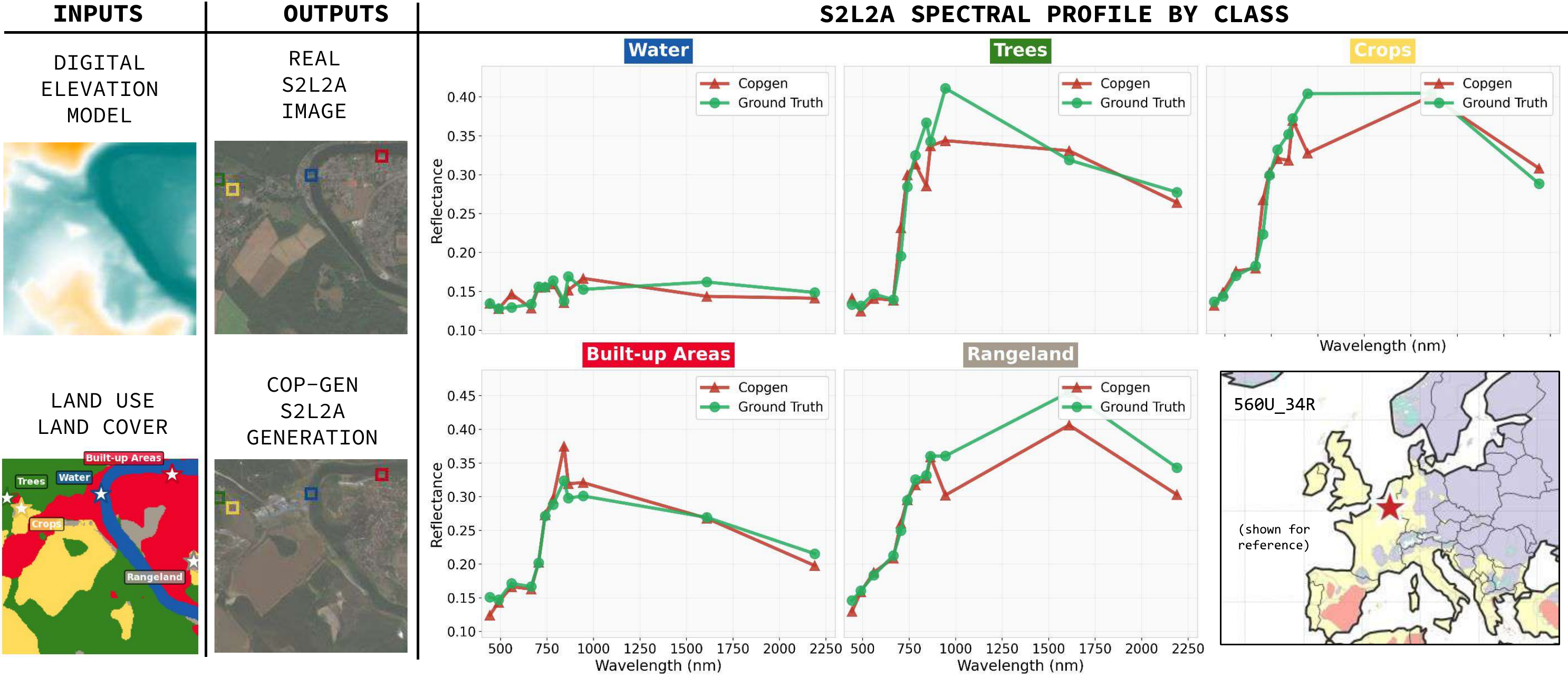}
    \caption{\textbf{Per-pixel spectral reflectance profiles across multiple LULC classes (560U\_34R tile).} The COP-GEN model reproduces characteristic Sentinel-2 band signatures for LULC classes, indicating robust learning of physical spectral patterns.}
    \label{fig:560U_34R_spectral_profile}
\end{figure*}
}
{\begin{figure*}
    \centering
    \includegraphics[width=\linewidth]{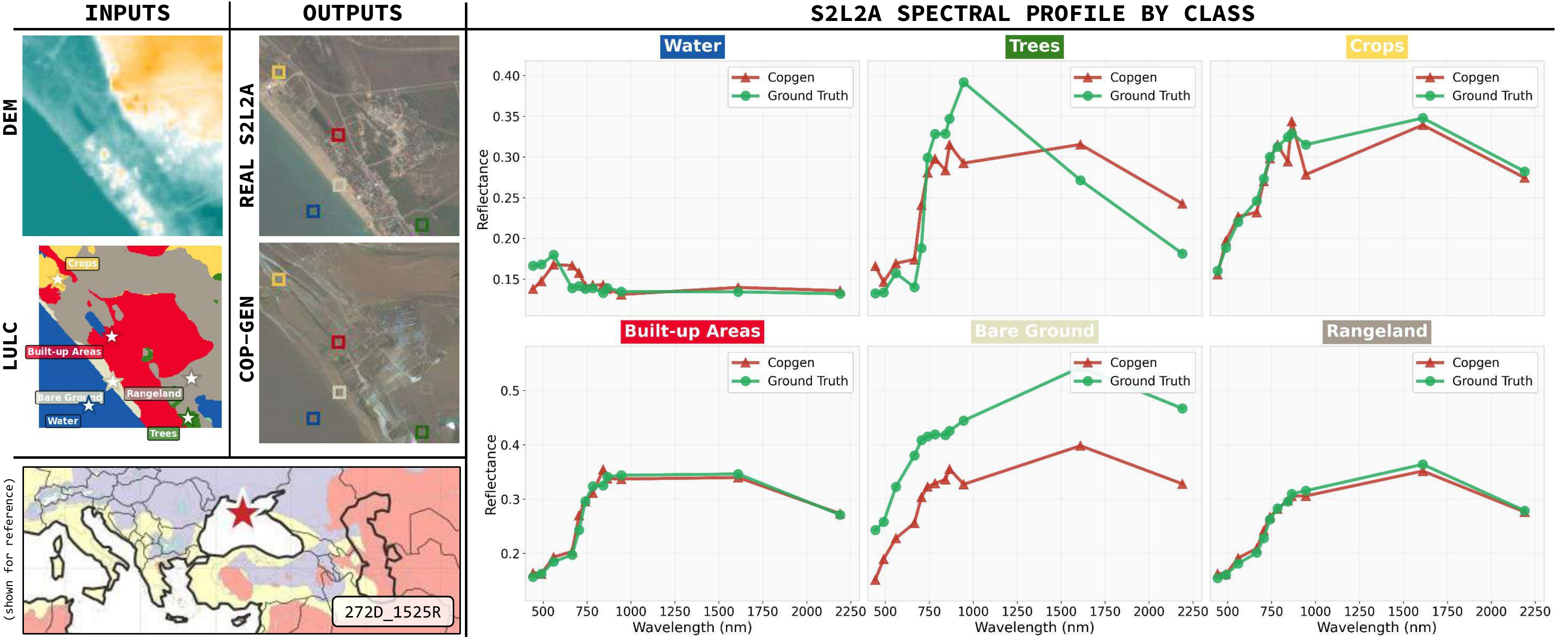}
    \caption{\textbf{Per-pixel spectral reflectance profiles across multiple LULC classes (502U\_263R tile).} COP-GEN closely matches Sentinel-2 band responses, indicating physically meaningful spectral signatures across land-cover types.}
    \label{fig:502U_263R_spectral_profile}
\end{figure*} 
}
{\begin{figure*}
    \centering
    \includegraphics[width=\linewidth]{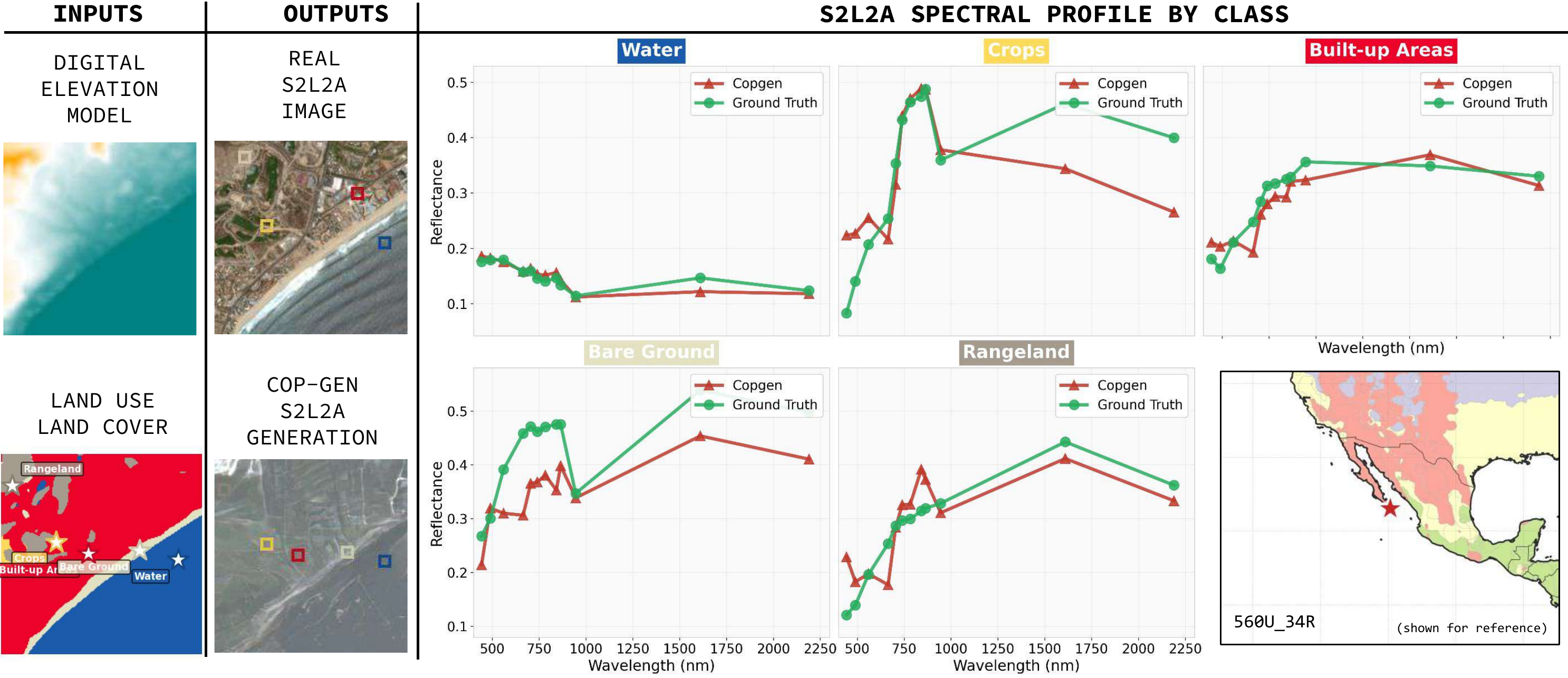}
    \caption{\textbf{Per-pixel spectral reflectance profiles across multiple LULC classes (256U\_1125L tile).} COP-GEN preserves characteristic Sentinel-2 spectral responses, illustrating consistent physical fidelity across land-cover types.}
    \label{fig:256U_1125L_spectral_profile}
\end{figure*}
}

{\begin{table*}[t]
\centering
\resizebox{\textwidth}{!}{%
\begin{tabular}{l l l cc cc}
\toprule
& & & \multicolumn{2}{c}{\textbf{COP-GEN}} & \multicolumn{2}{c}{\textbf{TerraMind}} \\
\cmidrule(lr){4-5} \cmidrule(lr){6-7}
\textbf{Target} & \textbf{Condition (Input)} & \textbf{Metric} & \textbf{Mean $\pm$ Std} & \textbf{Best} & \textbf{Mean $\pm$ Std} & \textbf{Best} \\
\midrule

\multirow{3}{*}{\textbf{DEM}} 
 & \multirow{3}{*}{Full Context} 
   & MAE $\downarrow$  & 301.79 {\scriptsize $\,\pm$ 260.49} & \textbf{26.80} & \textbf{159.96} {\scriptsize $\,\pm$ 515.97} & 145.62 \\
 & & RMSE $\downarrow$ & 304.05 {\scriptsize $\,\pm$ 261.26} & \textbf{30.72} & \textbf{161.66} {\scriptsize $\,\pm$ 516.68} & 147.54 \\
 & & SSIM $\uparrow$ & 0.18 {\scriptsize $\,\pm$ 0.15}      & \textbf{0.45}  & \textbf{0.37} {\scriptsize $\,\pm$ 0.31}      & 0.44 \\
\midrule

\multirow{4}{*}{\textbf{LULC}} 
 & \multirow{4}{*}{Full Context} 
   & Top-1 Acc $\uparrow$ & 0.52 {\scriptsize $\,\pm$ 0.18} & \textbf{0.84} & \textbf{0.78} {\scriptsize $\,\pm$ 0.30} & 0.80 \\
 & & Mean IoU $\uparrow$  & 0.19 {\scriptsize $\,\pm$ 0.10} & 0.42          & \textbf{0.46} {\scriptsize $\,\pm$ 0.28} & \textbf{0.55} \\
 & & Fw. IoU $\uparrow$   & 0.46 {\scriptsize $\,\pm$ 0.19} & \textbf{0.77} & \textbf{0.72} {\scriptsize $\,\pm$ 0.31} & 0.75 \\
 & & Mean F1 $\uparrow$   & 0.24 {\scriptsize $\,\pm$ 0.11} & 0.49          & \textbf{0.51} {\scriptsize $\,\pm$ 0.28} & \textbf{0.61} \\
\midrule

\multirow{3}{*}{\textbf{S1RTC}} 
 & \multirow{3}{*}{Full Context} 
   & MAE $\downarrow$  & 3.48 {\scriptsize $\,\pm$ 0.95}  & \textbf{2.63} & \textbf{2.74} {\scriptsize $\,\pm$ 1.25} & 2.64 \\
 & & SSIM $\uparrow$ & 0.10 {\scriptsize $\,\pm$ 0.03}  & 0.12          & \textbf{0.17} {\scriptsize $\,\pm$ 0.05} & \textbf{0.18} \\
 & & PSNR $\uparrow$ & 14.87 {\scriptsize $\,\pm$ 1.75} & 16.83         & \textbf{19.37} {\scriptsize $\,\pm$ 2.33} & \textbf{19.65} \\
\midrule

\multirow{6}{*}{\textbf{S2L1C}} 
 & \multirow{3}{*}{Full Context} 
   & MAE $\downarrow$  & \textbf{0.04} {\scriptsize $\,\pm$ 0.03} & \textbf{0.02} & 0.12 {\scriptsize $\,\pm$ 0.12} & 0.11 \\
 & & SSIM $\uparrow$ & 0.39 {\scriptsize $\,\pm$ 0.12}          & \textbf{0.48} & 0.39 {\scriptsize $\,\pm$ 0.16} & 0.43 \\
 & & PSNR $\uparrow$ & \textbf{17.22} {\scriptsize $\,\pm$ 4.19} & \textbf{21.16} & 12.39 {\scriptsize $\,\pm$ 3.00} & 12.77 \\
 \cmidrule(l){2-7}
 & \multirow{3}{*}{\textit{w/o S2L2A}} 
   & MAE $\downarrow$  & \textbf{0.12} {\scriptsize $\,\pm$ 0.10} & \textbf{0.05} & 0.13 {\scriptsize $\,\pm$ 0.11} & 0.12 \\
 & & SSIM $\uparrow$ & 0.10 {\scriptsize $\,\pm$ 0.05}          & 0.19          & \textbf{0.22} {\scriptsize $\,\pm$ 0.10} & \textbf{0.26} \\
 & & PSNR $\uparrow$ & 9.52 {\scriptsize $\,\pm$ 2.85}          & \textbf{13.92} & \textbf{12.13} {\scriptsize $\,\pm$ 2.67} & 12.68 \\
\midrule

\multirow{6}{*}{\textbf{S2L2A}} 
 & \multirow{3}{*}{Full Context} 
   & MAE $\downarrow$  & \textbf{0.06} {\scriptsize $\,\pm$ 0.04} & \textbf{0.02} & 0.11 {\scriptsize $\,\pm$ 0.15} & 0.10 \\
 & & SSIM $\uparrow$ & 0.43 {\scriptsize $\,\pm$ 0.12}          & \textbf{0.53} & 0.43 {\scriptsize $\,\pm$ 0.15} & 0.47 \\
 & & PSNR $\uparrow$ & \textbf{17.08} {\scriptsize $\,\pm$ 4.66} & \textbf{22.47} & 16.62 {\scriptsize $\,\pm$ 4.94} & 17.46 \\
 \cmidrule(l){2-7}
 & \multirow{3}{*}{\textit{w/o S2L1C}} 
   & MAE $\downarrow$  & 0.15 {\scriptsize $\,\pm$ 0.11}          & \textbf{0.06} & \textbf{0.11} {\scriptsize $\,\pm$ 0.13} & 0.10 \\
 & & SSIM $\uparrow$ & 0.12 {\scriptsize $\,\pm$ 0.06}          & 0.20          & \textbf{0.30} {\scriptsize $\,\pm$ 0.11} & \textbf{0.35} \\
 & & PSNR $\uparrow$ & 9.50 {\scriptsize $\,\pm$ 2.92}          & 14.40         & \textbf{15.16} {\scriptsize $\,\pm$ 4.09} & \textbf{16.18} \\
\midrule

\textbf{LatLon} 
 & Full Context 
   & Mean km $\downarrow$ & 810.48 {\scriptsize $\,\pm$ 1332.12} & 98.35 & \textbf{483.85} {\scriptsize $\,\pm$ 1270.39} & \textbf{94.25} \\

\bottomrule
\end{tabular}%
}
\caption{\textbf{Tile-Level Cross-Modal Generation Analysis.} We sample 36 times per tile. We report the performance of COP-GEN and TerraMind across all tiles. \textit{Mean} denotes the average performance across all seeds and tiles (measuring stability), while \textit{Best} denotes the oracle selection of the best generation per tile (measuring peak capability). The ``Condition'' column indicates the available input modalities. \textbf{Bold} indicates the best result between the two models for a given metric/setting.}
\label{tab:tile_level_detailed}
\end{table*}
}
{\begin{table*}[t]
\centering
\resizebox{\textwidth}{!}{%
\begin{tabular}{llcccc}
\toprule
\textbf{Model} & \textbf{Conditioning Variant} & \textbf{MAE} $\downarrow$ & \textbf{RMSE} $\downarrow$ & \textbf{SSIM} $\uparrow$ & \textbf{PSNR} $\uparrow$ \\
\midrule
\multicolumn{6}{l}{\textit{\textbf{Target: Digital Elevation Model (DEM)}}} \\
COP-GEN    & All-minus-target & 301.79 {\scriptsize $\,\pm$ 9.48} & \textbf{483.21} {\scriptsize $\,\pm$ 20.54} & 0.18 {\scriptsize $\,\pm$ 0.01} & 5.79 {\scriptsize $\,\pm$ 0.10} \\
TerraMind & All-minus-target & \textbf{159.96} {\scriptsize $\,\pm$ 1.57} & 542.53 {\scriptsize $\,\pm$ 12.51} & \textbf{0.37} {\scriptsize $\,\pm$ 0.00} & \textbf{7.83} {\scriptsize $\,\pm$ 0.05} \\
\midrule
\multicolumn{6}{l}{\textit{\textbf{Target: Sentinel-1 (S1RTC)}}} \\
COP-GEN    & All-minus-target & 3.48 {\scriptsize $\,\pm$ 0.05} & 4.67 {\scriptsize $\,\pm$ 0.16} & 0.10 {\scriptsize $\,\pm$ 0.00} & 14.24 {\scriptsize $\,\pm$ 0.08} \\
TerraMind & All-minus-target & \textbf{2.74} {\scriptsize $\,\pm$ 0.00} & \textbf{3.63} {\scriptsize $\,\pm$ 0.01} & \textbf{0.17} {\scriptsize $\,\pm$ 0.00} & \textbf{18.28} {\scriptsize $\,\pm$ 0.02} \\
\midrule
\multicolumn{6}{l}{\textit{\textbf{Target: Sentinel-2 L1C (S2L1C)}}} \\
COP-GEN    & All-minus-target & \textbf{0.04} {\scriptsize $\,\pm$ 0.00} & \textbf{0.06} {\scriptsize $\,\pm$ 0.00} & \textbf{0.39} {\scriptsize $\,\pm$ 0.00} & \textbf{13.01} {\scriptsize $\,\pm$ 0.27} \\
TerraMind & All-minus-target & 0.12 {\scriptsize $\,\pm$ 0.00} & 0.21 {\scriptsize $\,\pm$ 0.01} & \textbf{0.39} {\scriptsize $\,\pm$ 0.00} & 11.21 {\scriptsize $\,\pm$ 0.09} \\
\cmidrule(l){2-6}
COP-GEN    & No S2L2A         & \textbf{0.12} {\scriptsize $\,\pm$ 0.00} & \textbf{0.19} {\scriptsize $\,\pm$ 0.00} & 0.10 {\scriptsize $\,\pm$ 0.00} & 8.09 {\scriptsize $\,\pm$ 0.09} \\
TerraMind & No S2L2A         & 0.13 {\scriptsize $\,\pm$ 0.00} & 0.20 {\scriptsize $\,\pm$ 0.00} & \textbf{0.22} {\scriptsize $\,\pm$ 0.00} & \textbf{11.19} {\scriptsize $\,\pm$ 0.01} \\
\midrule
\multicolumn{6}{l}{\textit{\textbf{Target: Sentinel-2 L2A (S2L2A)}}} \\
COP-GEN    & All-minus-target & \textbf{0.06} {\scriptsize $\,\pm$ 0.00} & \textbf{0.08} {\scriptsize $\,\pm$ 0.00} & \textbf{0.43} {\scriptsize $\,\pm$ 0.00} & 12.07 {\scriptsize $\,\pm$ 0.21} \\
TerraMind & All-minus-target & 0.11 {\scriptsize $\,\pm$ 0.01} & 0.21 {\scriptsize $\,\pm$ 0.02} & \textbf{0.43} {\scriptsize $\,\pm$ 0.00} & \textbf{13.04} {\scriptsize $\,\pm$ 0.36} \\
\cmidrule(l){2-6}
COP-GEN    & No S2L1C         & 0.15 {\scriptsize $\,\pm$ 0.00} & 0.23 {\scriptsize $\,\pm$ 0.00} & 0.12 {\scriptsize $\,\pm$ 0.00} & 7.98 {\scriptsize $\,\pm$ 0.10} \\
TerraMind & No S2L1C         & \textbf{0.11} {\scriptsize $\,\pm$ 0.01} & \textbf{0.20} {\scriptsize $\,\pm$ 0.01} & \textbf{0.30} {\scriptsize $\,\pm$ 0.00} & \textbf{12.66} {\scriptsize $\,\pm$ 0.18} \\
\midrule
\midrule
& & \textbf{Top-1 Acc} $\uparrow$ & \textbf{Top-3 Acc} $\uparrow$ & \textbf{Mean IoU} $\uparrow$ & \textbf{Mean F1} $\uparrow$ \\
\midrule
\multicolumn{6}{l}{\textit{\textbf{Target: Land Use / Land Cover (LULC)}}} \\
COP-GEN    & All-minus-target & 0.52 {\scriptsize $\,\pm$ 0.01} & 0.55 {\scriptsize $\,\pm$ 0.01} & 0.27 {\scriptsize $\,\pm$ 0.01} & 0.38 {\scriptsize $\,\pm$ 0.01} \\
TerraMind & All-minus-target & \textbf{0.78} {\scriptsize $\,\pm$ 0.00} & \textbf{0.89} {\scriptsize $\,\pm$ 0.00} & \textbf{0.49} {\scriptsize $\,\pm$ 0.00} & \textbf{0.58} {\scriptsize $\,\pm$ 0.00} \\
\midrule
\midrule
& & \textbf{Median km} $\downarrow$ & \textbf{Mean km} $\downarrow$ & \textbf{Std km} & \textbf{RMSE km} $\downarrow$ \\
\midrule
\multicolumn{6}{l}{\textit{\textbf{Target: Geolocation (Lat/Lon)}}} \\
COP-GEN    & All-minus-target & 414.98 {\scriptsize $\,\pm$ 13.97} & 810.48 {\scriptsize $\,\pm$ 38.89} & 1715.53 {\scriptsize $\,\pm$ 156.32} & 1897.83 {\scriptsize $\,\pm$ 155.35} \\
TerraMind & All-minus-target & \textbf{44.34} {\scriptsize $\,\pm$ 5.91} & \textbf{492.94} {\scriptsize $\,\pm$ 58.35} & 1805.32 {\scriptsize $\,\pm$ 225.04} & \textbf{1871.51} {\scriptsize $\,\pm$ 231.69} \\
\bottomrule
\end{tabular}
}
\caption{\textbf{Experiment-Level Cross-Modal Generation Performance.} We generate a single sample per tile, and obtain a final aggregated test metric. We run 36 times and report mean and std. Comparison of COP-GEN vs. TerraMind. For distinct modalities, we report domain-specific metrics. \textit{All-minus-target} implies all available modalities are used as conditioning. \textbf{Bold} indicates the best result.}
\label{tab:experiment_level_detailed}
\end{table*}
}
{\begin{table*}[t]
\centering
\resizebox{0.75\textwidth}{!}{%
\begin{tabular}{l l l cc cc}
\toprule
& & & \multicolumn{2}{c}{\textbf{COP-GEN}} & \multicolumn{2}{c}{\textbf{TerraMind}} \\
\cmidrule(lr){4-5} \cmidrule(lr){6-7}
\textbf{Target} & \textbf{Removed Modality} & \textbf{Metric} & \textbf{Mean $\pm$ Std} & \textbf{Best} & \textbf{Mean $\pm$ Std} & \textbf{Best} \\
\midrule

\multirow{10}{*}{\textbf{DEM}} 
 & \multirow{2}{*}{\textit{w/o LatLon}} 
   & MAE $\downarrow$  & 308.65 {\scriptsize $\,\pm$ 280.81} & \textbf{47.44} & \textbf{164.01} {\scriptsize $\,\pm$ 546.39} & 140.01 \\
 & & SSIM $\uparrow$ & 0.18 {\scriptsize $\,\pm$ 0.16}      & 0.41           & \textbf{0.40} {\scriptsize $\,\pm$ 0.28}      & \textbf{0.53} \\
 \cmidrule(lr){2-7}
 & \multirow{2}{*}{\textit{w/o LULC}} 
   & MAE $\downarrow$  & 305.07 {\scriptsize $\,\pm$ 271.09} & \textbf{46.96} & \textbf{163.98} {\scriptsize $\,\pm$ 545.79} & 140.80 \\
 & & SSIM $\uparrow$ & 0.17 {\scriptsize $\,\pm$ 0.16}      & 0.40           & \textbf{0.40} {\scriptsize $\,\pm$ 0.28}      & \textbf{0.53} \\
 \cmidrule(lr){2-7}
 & \multirow{2}{*}{\textit{w/o S1RTC}} 
   & MAE $\downarrow$  & 318.17 {\scriptsize $\,\pm$ 281.35} & \textbf{54.85} & \textbf{164.12} {\scriptsize $\,\pm$ 541.42} & 140.86 \\
 & & SSIM $\uparrow$ & 0.18 {\scriptsize $\,\pm$ 0.15}      & 0.37           & \textbf{0.40} {\scriptsize $\,\pm$ 0.28}      & \textbf{0.53} \\
 \cmidrule(lr){2-7}
 & \multirow{2}{*}{\textit{w/o S2L1C}} 
   & MAE $\downarrow$  & 434.98 {\scriptsize $\,\pm$ 406.69} & \textbf{51.85} & \textbf{172.71} {\scriptsize $\,\pm$ 539.85} & 146.00 \\
 & & SSIM $\uparrow$ & 0.17 {\scriptsize $\,\pm$ 0.15}      & 0.39           & \textbf{0.39} {\scriptsize $\,\pm$ 0.28}      & \textbf{0.52} \\
 \cmidrule(lr){2-7}
 & \multirow{2}{*}{\textit{w/o S2L2A}} 
   & MAE $\downarrow$  & 313.25 {\scriptsize $\,\pm$ 285.84} & \textbf{45.78} & \textbf{173.84} {\scriptsize $\,\pm$ 546.43} & 146.71 \\
 & & SSIM $\uparrow$ & 0.18 {\scriptsize $\,\pm$ 0.16}      & 0.40           & \textbf{0.39} {\scriptsize $\,\pm$ 0.28}      & \textbf{0.51} \\
\midrule

\multirow{10}{*}{\textbf{LULC}} 
 & \multirow{2}{*}{\textit{w/o LatLon}} 
   & mIoU $\uparrow$     & 0.19 {\scriptsize $\,\pm$ 0.10} & 0.37 & \textbf{0.45} {\scriptsize $\,\pm$ 0.26} & \textbf{0.54} \\
 & & Top-1 Acc $\uparrow$ & 0.53 {\scriptsize $\,\pm$ 0.18} & \textbf{0.81} & \textbf{0.78} {\scriptsize $\,\pm$ 0.29} & 0.80 \\
 \cmidrule(lr){2-7}
 & \multirow{2}{*textit{w/o DEM}} 
   & mIoU $\uparrow$     & 0.19 {\scriptsize $\,\pm$ 0.10} & 0.38 & \textbf{0.45} {\scriptsize $\,\pm$ 0.26} & \textbf{0.55} \\
 & & Top-1 Acc $\uparrow$ & 0.52 {\scriptsize $\,\pm$ 0.18} & \textbf{0.80} & \textbf{0.78} {\scriptsize $\,\pm$ 0.30} & \textbf{0.80} \\
 \cmidrule(lr){2-7}
 & \multirow{2}{*}{\textit{w/o S1RTC}} 
   & mIoU $\uparrow$     & 0.18 {\scriptsize $\,\pm$ 0.10} & 0.37 & \textbf{0.45} {\scriptsize $\,\pm$ 0.26} & \textbf{0.54} \\
 & & Top-1 Acc $\uparrow$ & 0.52 {\scriptsize $\,\pm$ 0.18} & 0.79 & \textbf{0.78} {\scriptsize $\,\pm$ 0.29} & \textbf{0.80} \\
 \cmidrule(lr){2-7}
 & \multirow{2}{*}{\textit{w/o S2L1C}} 
   & mIoU $\uparrow$     & 0.19 {\scriptsize $\,\pm$ 0.10} & 0.38 & \textbf{0.46} {\scriptsize $\,\pm$ 0.26} & \textbf{0.55} \\
 & & Top-1 Acc $\uparrow$ & 0.54 {\scriptsize $\,\pm$ 0.18} & \textbf{0.80} & \textbf{0.78} {\scriptsize $\,\pm$ 0.29} & \textbf{0.80} \\
 \cmidrule(lr){2-7}
 & \multirow{2}{*}{\textit{w/o S2L2A}} 
   & mIoU $\uparrow$     & 0.19 {\scriptsize $\,\pm$ 0.10} & 0.37 & \textbf{0.46} {\scriptsize $\,\pm$ 0.27} & \textbf{0.55} \\
 & & Top-1 Acc $\uparrow$ & 0.54 {\scriptsize $\,\pm$ 0.18} & \textbf{0.80} & \textbf{0.78} {\scriptsize $\,\pm$ 0.30} & \textbf{0.80} \\
\midrule

\multirow{10}{*}{\textbf{S1RTC}} 
 & \multirow{2}{*}{\textit{w/o LatLon}} 
   & MAE $\downarrow$  & 3.45 {\scriptsize $\,\pm$ 0.94} & 2.70 & \textbf{2.73} {\scriptsize $\,\pm$ 1.20} & \textbf{2.63} \\
 & & PSNR $\uparrow$ & 14.91 {\scriptsize $\,\pm$ 1.76} & 16.61 & \textbf{19.61} {\scriptsize $\,\pm$ 2.27} & \textbf{20.05} \\
 \cmidrule(lr){2-7}
 & \multirow{2}{*}{\textit{w/o DEM}} 
   & MAE $\downarrow$  & 3.58 {\scriptsize $\,\pm$ 0.93} & 2.75 & \textbf{2.72} {\scriptsize $\,\pm$ 1.20} & \textbf{2.63} \\
 & & PSNR $\uparrow$ & 14.61 {\scriptsize $\,\pm$ 1.74} & 16.43 & \textbf{19.63} {\scriptsize $\,\pm$ 2.26} & \textbf{20.06} \\
 \cmidrule(lr){2-7}
 & \multirow{2}{*}{\textit{w/o LULC}} 
   & MAE $\downarrow$  & 3.54 {\scriptsize $\,\pm$ 0.91} & 2.76 & \textbf{2.73} {\scriptsize $\,\pm$ 1.21} & \textbf{2.63} \\
 & & PSNR $\uparrow$ & 14.68 {\scriptsize $\,\pm$ 1.72} & 16.41 & \textbf{19.62} {\scriptsize $\,\pm$ 2.28} & \textbf{20.06} \\
 \cmidrule(lr){2-7}
 & \multirow{2}{*}{\textit{w/o S2L1C}} 
   & MAE $\downarrow$  & 3.46 {\scriptsize $\,\pm$ 0.91} & 2.70 & \textbf{2.74} {\scriptsize $\,\pm$ 1.24} & \textbf{2.64} \\
 & & PSNR $\uparrow$ & 14.90 {\scriptsize $\,\pm$ 1.73} & 16.58 & \textbf{19.61} {\scriptsize $\,\pm$ 2.30} & \textbf{20.05} \\
 \cmidrule(lr){2-7}
 & \multirow{2}{*}{\textit{w/o S2L2A}} 
   & MAE $\downarrow$  & 3.44 {\scriptsize $\,\pm$ 0.89} & 2.68 & \textbf{2.72} {\scriptsize $\,\pm$ 1.18} & \textbf{2.62} \\
 & & PSNR $\uparrow$ & 14.97 {\scriptsize $\,\pm$ 1.69} & 16.65 & \textbf{19.63} {\scriptsize $\,\pm$ 2.26} & \textbf{20.07} \\
\midrule

\multirow{10}{*}{\textbf{S2L1C}} 
 & \multirow{2}{*}{\textit{w/o LatLon}} 
   & MAE $\downarrow$  & \textbf{0.04} {\scriptsize $\,\pm$ 0.03} & \textbf{0.02} & 0.12 {\scriptsize $\,\pm$ 0.12} & 0.11 \\
 & & PSNR $\uparrow$ & \textbf{17.18} {\scriptsize $\,\pm$ 4.34} & \textbf{20.85} & 12.61 {\scriptsize $\,\pm$ 3.02} & 13.13 \\
 \cmidrule(lr){2-7}
 & \multirow{2}{*}{\textit{w/o DEM}} 
   & MAE $\downarrow$  & \textbf{0.04} {\scriptsize $\,\pm$ 0.03} & \textbf{0.02} & 0.12 {\scriptsize $\,\pm$ 0.12} & 0.11 \\
 & & PSNR $\uparrow$ & \textbf{17.01} {\scriptsize $\,\pm$ 4.25} & \textbf{20.83} & 12.62 {\scriptsize $\,\pm$ 3.02} & 13.14 \\
 \cmidrule(lr){2-7}
 & \multirow{2}{*}{\textit{w/o LULC}} 
   & MAE $\downarrow$  & \textbf{0.04} {\scriptsize $\,\pm$ 0.03} & \textbf{0.02} & 0.12 {\scriptsize $\,\pm$ 0.12} & 0.11 \\
 & & PSNR $\uparrow$ & \textbf{17.11} {\scriptsize $\,\pm$ 4.31} & \textbf{20.91} & 12.62 {\scriptsize $\,\pm$ 3.02} & 13.14 \\
 \cmidrule(lr){2-7}
 & \multirow{2}{*}{\textit{w/o S1RTC}} 
   & MAE $\downarrow$  & \textbf{0.04} {\scriptsize $\,\pm$ 0.03} & \textbf{0.02} & 0.12 {\scriptsize $\,\pm$ 0.12} & 0.11 \\
 & & PSNR $\uparrow$ & \textbf{17.09} {\scriptsize $\,\pm$ 4.30} & \textbf{20.86} & 12.62 {\scriptsize $\,\pm$ 3.02} & 13.14 \\
 \cmidrule(lr){2-7}
 & \multirow{2}{*}{\textit{w/o S2L2A}} 
   & MAE $\downarrow$  & \textbf{0.12} {\scriptsize $\,\pm$ 0.10} & \textbf{0.06} & \textbf{0.12} {\scriptsize $\,\pm$ 0.11} & 0.12 \\
 & & PSNR $\uparrow$ & 9.45 {\scriptsize $\,\pm$ 2.88} & \textbf{13.24} & \textbf{12.35} {\scriptsize $\,\pm$ 2.68} & 12.99 \\
\midrule

\multirow{10}{*}{\textbf{S2L2A}} 
 & \multirow{2}{*}{\textit{w/o LatLon}} 
   & MAE $\downarrow$  & \textbf{0.06} {\scriptsize $\,\pm$ 0.04} & \textbf{0.02} & 0.11 {\scriptsize $\,\pm$ 0.15} & 0.10 \\
 & & PSNR $\uparrow$ & \textbf{16.99} {\scriptsize $\,\pm$ 4.91} & \textbf{22.02} & 16.94 {\scriptsize $\,\pm$ 4.94} & 17.88 \\
 \cmidrule(lr){2-7}
 & \multirow{2}{*}{\textit{w/o DEM}} 
   & MAE $\downarrow$  & \textbf{0.06} {\scriptsize $\,\pm$ 0.04} & \textbf{0.02} & 0.11 {\scriptsize $\,\pm$ 0.15} & 0.10 \\
 & & PSNR $\uparrow$ & \textbf{17.01} {\scriptsize $\,\pm$ 4.84} & \textbf{22.07} & 16.94 {\scriptsize $\,\pm$ 4.96} & 17.87 \\
 \cmidrule(lr){2-7}
 & \multirow{2}{*}{\textit{w/o LULC}} 
   & MAE $\downarrow$  & \textbf{0.06} {\scriptsize $\,\pm$ 0.04} & \textbf{0.02} & 0.11 {\scriptsize $\,\pm$ 0.15} & 0.10 \\
 & & PSNR $\uparrow$ & \textbf{17.03} {\scriptsize $\,\pm$ 4.79} & \textbf{22.15} & 16.90 {\scriptsize $\,\pm$ 4.97} & 17.83 \\
 \cmidrule(lr){2-7}
 & \multirow{2}{*}{\textit{w/o S1RTC}} 
   & MAE $\downarrow$  & \textbf{0.06} {\scriptsize $\,\pm$ 0.04} & \textbf{0.02} & 0.11 {\scriptsize $\,\pm$ 0.15} & 0.10 \\
 & & PSNR $\uparrow$ & \textbf{16.99} {\scriptsize $\,\pm$ 4.70} & \textbf{22.14} & 16.89 {\scriptsize $\,\pm$ 4.97} & 17.82 \\
 \cmidrule(lr){2-7}
 & \multirow{2}{*}{\textit{w/o S2L1C}} 
   & MAE $\downarrow$  & 0.15 {\scriptsize $\,\pm$ 0.11}          & \textbf{0.07} & \textbf{0.11} {\scriptsize $\,\pm$ 0.13} & 0.10 \\
 & & PSNR $\uparrow$ & 9.50 {\scriptsize $\,\pm$ 2.92}          & 13.75         & \textbf{15.40} {\scriptsize $\,\pm$ 4.08} & \textbf{16.46} \\
\midrule

\multirow{5}{*}{\textbf{LatLon}} 
 & \textit{w/o DEM}   & Mean km $\downarrow$ & 1402.1 {\scriptsize $\,\pm$ 1786.6} & 210.5 & \textbf{468.16} {\scriptsize $\,\pm$ 1199.3} & \textbf{90.67} \\
 & \textit{w/o LULC}  & Mean km $\downarrow$ & 1034.2 {\scriptsize $\,\pm$ 1620.2} & 188.7 & \textbf{454.33} {\scriptsize $\,\pm$ 1229.9} & \textbf{95.50} \\
 & \textit{w/o S1RTC} & Mean km $\downarrow$ & 1060.8 {\scriptsize $\,\pm$ 1611.6} & 173.1 & \textbf{476.16} {\scriptsize $\,\pm$ 1193.4} & \textbf{78.23} \\
 & \textit{w/o S2L1C} & Mean km $\downarrow$ & 1122.8 {\scriptsize $\,\pm$ 1914.4} & 193.4 & \textbf{472.48} {\scriptsize $\,\pm$ 1380.5} & \textbf{138.83} \\
 & \textit{w/o S2L2A} & Mean km $\downarrow$ & 1000.5 {\scriptsize $\,\pm$ 1574.1} & 182.4 & \textbf{380.31} {\scriptsize $\,\pm$ 1137.2} & \textbf{77.41} \\

\bottomrule
\end{tabular}%
}
\caption{\textbf{Extended Leave-One-Out Ablation Analysis (Tile-Level).} We analyze the impact of removing individual modalities on generation performance across all combinations. We report MAE and PSNR for all continuous bands. COP-GEN demonstrates superior peak performance (``Best'' column) across DEM and optical tasks, significantly outperforming TerraMind in MAE when correlated modalities are present.}
\label{tab:leave_one_out_full_tile_level}
\end{table*}
}
{\begin{table*}[t]
\centering
\resizebox{\textwidth}{!}{%
\begin{tabular}{lcccccccc}
\toprule
& \multicolumn{2}{c}{\textbf{MAE} $\downarrow$} & \multicolumn{2}{c}{\textbf{RMSE} $\downarrow$} & \multicolumn{2}{c}{\textbf{SSIM} $\uparrow$} & \multicolumn{2}{c}{\textbf{PSNR} $\uparrow$} \\
\cmidrule(lr){2-3} \cmidrule(lr){4-5} \cmidrule(lr){6-7} \cmidrule(lr){8-9}
\textbf{Removed} & \textbf{COP-GEN} & \textbf{Terra.} & \textbf{COP-GEN} & \textbf{Terra.} & \textbf{COP-GEN} & \textbf{Terra.} & \textbf{COP-GEN} & \textbf{Terra.} \\
\midrule
\multicolumn{9}{l}{\textit{\textbf{Target: Digital Elevation Model (DEM)}}} \\
Lat/Lon & 308.65 {\scriptsize $\,\pm$ 14.85} & \textbf{164.01} {\scriptsize $\,\pm$ 4.38} & \textbf{500.39} {\scriptsize $\,\pm$ 30.78} & 574.52 {\scriptsize $\,\pm$ 35.96} & 0.18 {\scriptsize $\,\pm$ 0.01} & \textbf{0.40} {\scriptsize $\,\pm$ 0.03} & 5.73 {\scriptsize $\,\pm$ 0.09} & \textbf{7.83} {\scriptsize $\,\pm$ 0.05} \\
LULC & 305.07 {\scriptsize $\,\pm$ 10.36} & \textbf{163.98} {\scriptsize $\,\pm$ 4.70} & \textbf{493.47} {\scriptsize $\,\pm$ 24.48} & 573.99 {\scriptsize $\,\pm$ 37.49} & 0.17 {\scriptsize $\,\pm$ 0.01} & \textbf{0.40} {\scriptsize $\,\pm$ 0.03} & 5.78 {\scriptsize $\,\pm$ 0.10} & \textbf{7.82} {\scriptsize $\,\pm$ 0.05} \\
S1RTC & 318.17 {\scriptsize $\,\pm$ 10.19} & \textbf{164.12} {\scriptsize $\,\pm$ 3.75} & \textbf{513.10} {\scriptsize $\,\pm$ 17.68} & 569.83 {\scriptsize $\,\pm$ 32.49} & 0.18 {\scriptsize $\,\pm$ 0.00} & \textbf{0.40} {\scriptsize $\,\pm$ 0.03} & 5.71 {\scriptsize $\,\pm$ 0.11} & \textbf{7.74} {\scriptsize $\,\pm$ 0.05} \\
S2L1C & 434.98 {\scriptsize $\,\pm$ 16.66} & \textbf{172.71} {\scriptsize $\,\pm$ 3.38} & 762.06 {\scriptsize $\,\pm$ 38.21} & \textbf{571.05} {\scriptsize $\,\pm$ 31.62} & 0.17 {\scriptsize $\,\pm$ 0.01} & \textbf{0.39} {\scriptsize $\,\pm$ 0.02} & 5.59 {\scriptsize $\,\pm$ 0.10} & \textbf{7.59} {\scriptsize $\,\pm$ 0.05} \\
S2L2A & 313.25 {\scriptsize $\,\pm$ 10.16} & \textbf{173.84} {\scriptsize $\,\pm$ 4.97} & \textbf{507.14} {\scriptsize $\,\pm$ 18.63} & 577.96 {\scriptsize $\,\pm$ 39.14} & 0.18 {\scriptsize $\,\pm$ 0.01} & \textbf{0.39} {\scriptsize $\,\pm$ 0.03} & 5.75 {\scriptsize $\,\pm$ 0.09} & \textbf{7.58} {\scriptsize $\,\pm$ 0.03} \\
\midrule
\multicolumn{9}{l}{\textit{\textbf{Target: Sentinel-1 (S1RTC)}}} \\
Lat/Lon & 3.45 {\scriptsize $\,\pm$ 0.04} & \textbf{2.73} {\scriptsize $\,\pm$ 0.01} & 4.57 {\scriptsize $\,\pm$ 0.10} & \textbf{3.61} {\scriptsize $\,\pm$ 0.02} & 0.10 {\scriptsize $\,\pm$ 0.00} & \textbf{0.18} {\scriptsize $\,\pm$ 0.01} & 14.29 {\scriptsize $\,\pm$ 0.08} & \textbf{18.58} {\scriptsize $\,\pm$ 0.29} \\
DEM & 3.58 {\scriptsize $\,\pm$ 0.05} & \textbf{2.72} {\scriptsize $\,\pm$ 0.01} & 4.77 {\scriptsize $\,\pm$ 0.13} & \textbf{3.60} {\scriptsize $\,\pm$ 0.02} & 0.09 {\scriptsize $\,\pm$ 0.00} & \textbf{0.18} {\scriptsize $\,\pm$ 0.01} & 14.01 {\scriptsize $\,\pm$ 0.08} & \textbf{18.60} {\scriptsize $\,\pm$ 0.28} \\
LULC & 3.54 {\scriptsize $\,\pm$ 0.04} & \textbf{2.73} {\scriptsize $\,\pm$ 0.01} & 4.73 {\scriptsize $\,\pm$ 0.14} & \textbf{3.61} {\scriptsize $\,\pm$ 0.02} & 0.09 {\scriptsize $\,\pm$ 0.00} & \textbf{0.18} {\scriptsize $\,\pm$ 0.01} & 14.09 {\scriptsize $\,\pm$ 0.07} & \textbf{18.58} {\scriptsize $\,\pm$ 0.30} \\
S2L1C & 3.46 {\scriptsize $\,\pm$ 0.05} & \textbf{2.74} {\scriptsize $\,\pm$ 0.01} & 4.69 {\scriptsize $\,\pm$ 0.17} & \textbf{3.63} {\scriptsize $\,\pm$ 0.01} & 0.10 {\scriptsize $\,\pm$ 0.00} & \textbf{0.18} {\scriptsize $\,\pm$ 0.01} & 14.30 {\scriptsize $\,\pm$ 0.06} & \textbf{18.54} {\scriptsize $\,\pm$ 0.27} \\
S2L2A & 3.44 {\scriptsize $\,\pm$ 0.06} & \textbf{2.72} {\scriptsize $\,\pm$ 0.01} & 4.68 {\scriptsize $\,\pm$ 0.24} & \textbf{3.60} {\scriptsize $\,\pm$ 0.02} & 0.10 {\scriptsize $\,\pm$ 0.00} & \textbf{0.18} {\scriptsize $\,\pm$ 0.01} & 14.37 {\scriptsize $\,\pm$ 0.07} & \textbf{18.61} {\scriptsize $\,\pm$ 0.28} \\
\midrule
\multicolumn{9}{l}{\textit{\textbf{Target: Sentinel-2 L1C (S2L1C)}}} \\
Lat/Lon & \textbf{0.04} {\scriptsize $\,\pm$ 0.00} & 0.12 {\scriptsize $\,\pm$ 0.00} & \textbf{0.06} {\scriptsize $\,\pm$ 0.00} & 0.21 {\scriptsize $\,\pm$ 0.00} & 0.39 {\scriptsize $\,\pm$ 0.01} & \textbf{0.40} {\scriptsize $\,\pm$ 0.01} & \textbf{12.86} {\scriptsize $\,\pm$ 0.25} & 11.36 {\scriptsize $\,\pm$ 0.17} \\
DEM & \textbf{0.04} {\scriptsize $\,\pm$ 0.00} & 0.12 {\scriptsize $\,\pm$ 0.00} & \textbf{0.06} {\scriptsize $\,\pm$ 0.00} & 0.21 {\scriptsize $\,\pm$ 0.00} & 0.39 {\scriptsize $\,\pm$ 0.00} & \textbf{0.40} {\scriptsize $\,\pm$ 0.01} & \textbf{12.86} {\scriptsize $\,\pm$ 0.28} & 11.37 {\scriptsize $\,\pm$ 0.17} \\
LULC & \textbf{0.04} {\scriptsize $\,\pm$ 0.00} & 0.12 {\scriptsize $\,\pm$ 0.00} & \textbf{0.06} {\scriptsize $\,\pm$ 0.00} & 0.21 {\scriptsize $\,\pm$ 0.00} & 0.39 {\scriptsize $\,\pm$ 0.00} & \textbf{0.40} {\scriptsize $\,\pm$ 0.01} & \textbf{12.85} {\scriptsize $\,\pm$ 0.26} & 11.37 {\scriptsize $\,\pm$ 0.17} \\
S1RTC & \textbf{0.04} {\scriptsize $\,\pm$ 0.00} & 0.12 {\scriptsize $\,\pm$ 0.00} & \textbf{0.06} {\scriptsize $\,\pm$ 0.00} & 0.21 {\scriptsize $\,\pm$ 0.00} & 0.39 {\scriptsize $\,\pm$ 0.00} & \textbf{0.40} {\scriptsize $\,\pm$ 0.01} & \textbf{12.83} {\scriptsize $\,\pm$ 0.27} & 11.37 {\scriptsize $\,\pm$ 0.17} \\
S2L2A & \textbf{0.12} {\scriptsize $\,\pm$ 0.00} & \textbf{0.12} {\scriptsize $\,\pm$ 0.00} & \textbf{0.19} {\scriptsize $\,\pm$ 0.00} & 0.20 {\scriptsize $\,\pm$ 0.00} & 0.10 {\scriptsize $\,\pm$ 0.00} & \textbf{0.22} {\scriptsize $\,\pm$ 0.01} & 8.04 {\scriptsize $\,\pm$ 0.10} & \textbf{11.37} {\scriptsize $\,\pm$ 0.18} \\
\midrule
\multicolumn{9}{l}{\textit{\textbf{Target: Sentinel-2 L2A (S2L2A)}}} \\
Lat/Lon & \textbf{0.06} {\scriptsize $\,\pm$ 0.00} & 0.11 {\scriptsize $\,\pm$ 0.00} & \textbf{0.08} {\scriptsize $\,\pm$ 0.00} & 0.21 {\scriptsize $\,\pm$ 0.00} & 0.43 {\scriptsize $\,\pm$ 0.01} & \textbf{0.44} {\scriptsize $\,\pm$ 0.01} & 11.94 {\scriptsize $\,\pm$ 0.19} & \textbf{13.21} {\scriptsize $\,\pm$ 0.21} \\
DEM & \textbf{0.06} {\scriptsize $\,\pm$ 0.00} & 0.11 {\scriptsize $\,\pm$ 0.00} & \textbf{0.08} {\scriptsize $\,\pm$ 0.00} & 0.21 {\scriptsize $\,\pm$ 0.00} & 0.43 {\scriptsize $\,\pm$ 0.00} & \textbf{0.44} {\scriptsize $\,\pm$ 0.01} & 11.99 {\scriptsize $\,\pm$ 0.19} & \textbf{13.21} {\scriptsize $\,\pm$ 0.21} \\
LULC & \textbf{0.06} {\scriptsize $\,\pm$ 0.00} & 0.11 {\scriptsize $\,\pm$ 0.00} & \textbf{0.08} {\scriptsize $\,\pm$ 0.00} & 0.21 {\scriptsize $\,\pm$ 0.00} & 0.43 {\scriptsize $\,\pm$ 0.00} & \textbf{0.44} {\scriptsize $\,\pm$ 0.01} & 12.00 {\scriptsize $\,\pm$ 0.19} & \textbf{13.18} {\scriptsize $\,\pm$ 0.22} \\
S1RTC & \textbf{0.06} {\scriptsize $\,\pm$ 0.00} & 0.11 {\scriptsize $\,\pm$ 0.00} & \textbf{0.08} {\scriptsize $\,\pm$ 0.00} & 0.21 {\scriptsize $\,\pm$ 0.00} & 0.43 {\scriptsize $\,\pm$ 0.00} & \textbf{0.45} {\scriptsize $\,\pm$ 0.01} & 12.03 {\scriptsize $\,\pm$ 0.17} & \textbf{13.17} {\scriptsize $\,\pm$ 0.21} \\
S2L1C & 0.15 {\scriptsize $\,\pm$ 0.00} & \textbf{0.11} {\scriptsize $\,\pm$ 0.00} & 0.22 {\scriptsize $\,\pm$ 0.00} & \textbf{0.20} {\scriptsize $\,\pm$ 0.00} & 0.12 {\scriptsize $\,\pm$ 0.00} & \textbf{0.31} {\scriptsize $\,\pm$ 0.01} & 7.97 {\scriptsize $\,\pm$ 0.12} & \textbf{12.82} {\scriptsize $\,\pm$ 0.20} \\
\midrule
\midrule
& \multicolumn{2}{c}{\textbf{Top-1 Acc} $\uparrow$} & \multicolumn{2}{c}{\textbf{Top-3 Acc} $\uparrow$} & \multicolumn{2}{c}{\textbf{Mean IoU} $\uparrow$} & \multicolumn{2}{c}{\textbf{Mean F1} $\uparrow$} \\
\cmidrule(lr){2-3} \cmidrule(lr){4-5} \cmidrule(lr){6-7} \cmidrule(lr){8-9}
\textbf{Removed} & \textbf{COP-GEN} & \textbf{Terra.} & \textbf{COP-GEN} & \textbf{Terra.} & \textbf{COP-GEN} & \textbf{Terra.} & \textbf{COP-GEN} & \textbf{Terra.} \\
\midrule
\multicolumn{9}{l}{\textit{\textbf{Target: Land Use / Land Cover (LULC)}}} \\
Lat/Lon & 0.53 {\scriptsize $\,\pm$ 0.01} & \textbf{0.78} {\scriptsize $\,\pm$ 0.00} & 0.56 {\scriptsize $\,\pm$ 0.01} & \textbf{0.89} {\scriptsize $\,\pm$ 0.00} & 0.27 {\scriptsize $\,\pm$ 0.01} & \textbf{0.49} {\scriptsize $\,\pm$ 0.00} & 0.39 {\scriptsize $\,\pm$ 0.01} & \textbf{0.58} {\scriptsize $\,\pm$ 0.00} \\
DEM & 0.52 {\scriptsize $\,\pm$ 0.01} & \textbf{0.78} {\scriptsize $\,\pm$ 0.00} & 0.55 {\scriptsize $\,\pm$ 0.01} & \textbf{0.89} {\scriptsize $\,\pm$ 0.00} & 0.27 {\scriptsize $\,\pm$ 0.01} & \textbf{0.49} {\scriptsize $\,\pm$ 0.00} & 0.38 {\scriptsize $\,\pm$ 0.01} & \textbf{0.58} {\scriptsize $\,\pm$ 0.00} \\
S1RTC & 0.52 {\scriptsize $\,\pm$ 0.01} & \textbf{0.78} {\scriptsize $\,\pm$ 0.00} & 0.54 {\scriptsize $\,\pm$ 0.01} & \textbf{0.89} {\scriptsize $\,\pm$ 0.00} & 0.26 {\scriptsize $\,\pm$ 0.01} & \textbf{0.49} {\scriptsize $\,\pm$ 0.00} & 0.38 {\scriptsize $\,\pm$ 0.01} & \textbf{0.58} {\scriptsize $\,\pm$ 0.00} \\
S2L1C & 0.54 {\scriptsize $\,\pm$ 0.01} & \textbf{0.78} {\scriptsize $\,\pm$ 0.00} & 0.57 {\scriptsize $\,\pm$ 0.01} & \textbf{0.90} {\scriptsize $\,\pm$ 0.00} & 0.28 {\scriptsize $\,\pm$ 0.01} & \textbf{0.49} {\scriptsize $\,\pm$ 0.00} & 0.39 {\scriptsize $\,\pm$ 0.01} & \textbf{0.58} {\scriptsize $\,\pm$ 0.00} \\
S2L2A & 0.54 {\scriptsize $\,\pm$ 0.01} & \textbf{0.78} {\scriptsize $\,\pm$ 0.00} & 0.57 {\scriptsize $\,\pm$ 0.01} & \textbf{0.89} {\scriptsize $\,\pm$ 0.00} & 0.28 {\scriptsize $\,\pm$ 0.01} & \textbf{0.49} {\scriptsize $\,\pm$ 0.00} & 0.39 {\scriptsize $\,\pm$ 0.01} & \textbf{0.58} {\scriptsize $\,\pm$ 0.00} \\
\midrule
\midrule
& \multicolumn{2}{c}{\textbf{Median km} $\downarrow$} & \multicolumn{2}{c}{\textbf{Mean km} $\downarrow$} & \multicolumn{2}{c}{\textbf{Std km}} & \multicolumn{2}{c}{\textbf{RMSE km} $\downarrow$} \\
\cmidrule(lr){2-3} \cmidrule(lr){4-5} \cmidrule(lr){6-7} \cmidrule(lr){8-9}
\textbf{Removed} & \textbf{COP-GEN} & \textbf{Terra.} & \textbf{COP-GEN} & \textbf{Terra.} & \textbf{COP-GEN} & \textbf{Terra.} & \textbf{COP-GEN} & \textbf{Terra.} \\
\midrule
\multicolumn{9}{l}{\textit{\textbf{Target: Geolocation (Lat/Lon)}}} \\
DEM & 544 {\scriptsize $\,\pm$ 29} & \textbf{57} {\scriptsize $\,\pm$ 2} & 1402 {\scriptsize $\,\pm$ 65} & \textbf{474} {\scriptsize $\,\pm$ 59} & 2623 {\scriptsize $\,\pm$ 161} & 1686 {\scriptsize $\,\pm$ 259} & 2974 {\scriptsize $\,\pm$ 169} & \textbf{1752} {\scriptsize $\,\pm$ 265} \\
LULC & 453 {\scriptsize $\,\pm$ 19} & \textbf{38} {\scriptsize $\,\pm$ 1} & 1034 {\scriptsize $\,\pm$ 65} & \textbf{461} {\scriptsize $\,\pm$ 59} & 2167 {\scriptsize $\,\pm$ 241} & 1733 {\scriptsize $\,\pm$ 268} & 2402 {\scriptsize $\,\pm$ 241} & \textbf{1793} {\scriptsize $\,\pm$ 273} \\
S1RTC & 449 {\scriptsize $\,\pm$ 20} & \textbf{51} {\scriptsize $\,\pm$ 6} & 1061 {\scriptsize $\,\pm$ 83} & \textbf{482} {\scriptsize $\,\pm$ 61} & 2243 {\scriptsize $\,\pm$ 249} & 1725 {\scriptsize $\,\pm$ 217} & 2482 {\scriptsize $\,\pm$ 259} & \textbf{1791} {\scriptsize $\,\pm$ 225} \\
S2L1C & 456 {\scriptsize $\,\pm$ 19} & \textbf{41} {\scriptsize $\,\pm$ 5} & 1123 {\scriptsize $\,\pm$ 73} & \textbf{480} {\scriptsize $\,\pm$ 51} & 2393 {\scriptsize $\,\pm$ 209} & 1806 {\scriptsize $\,\pm$ 169} & 2643 {\scriptsize $\,\pm$ 217} & \textbf{1868} {\scriptsize $\,\pm$ 175} \\
S2L2A & 443 {\scriptsize $\,\pm$ 14} & \textbf{41} {\scriptsize $\,\pm$ 4} & 1001 {\scriptsize $\,\pm$ 73} & \textbf{385} {\scriptsize $\,\pm$ 40} & 2112 {\scriptsize $\,\pm$ 250} & 1500 {\scriptsize $\,\pm$ 178} & 2337 {\scriptsize $\,\pm$ 257} & \textbf{1549} {\scriptsize $\,\pm$ 182} \\
\bottomrule
\end{tabular}
}
\caption{\textbf{Leave-One-Out Ablation Performance. Experiment-level.} Comparison of COP-GEN vs. TerraMind (Terra.) when removing a single input modality. We report Mean $\pm$ Std. \textbf{Bold} indicates the best result between the two models.}
\label{tab:leave_one_out_full_experiment_level}
\end{table*}
}

{\begin{table*}[t]
\centering
\small
\begin{tabular}{lll}
\toprule
& \textbf{TerraMind (Base)} & \textbf{COP-GEN (Base)} \\
\midrule
\textbf{Backbone Parameters} & 237.62 M & 286.99 M \\
\bottomrule
\end{tabular}
\caption{\textbf{Backbone parameter comparison.} Number of trainable parameters in the base backbone architectures of TerraMind and COP-GEN. COP-GEN achieves multimodal stochastic generation and any-to-any conditional modeling with a parameter count comparable to existing Earth observation foundation models.}
\label{tab:model_parameters}
\end{table*}
}

{\begin{figure*}
    \centering
    \includegraphics[width=\linewidth]{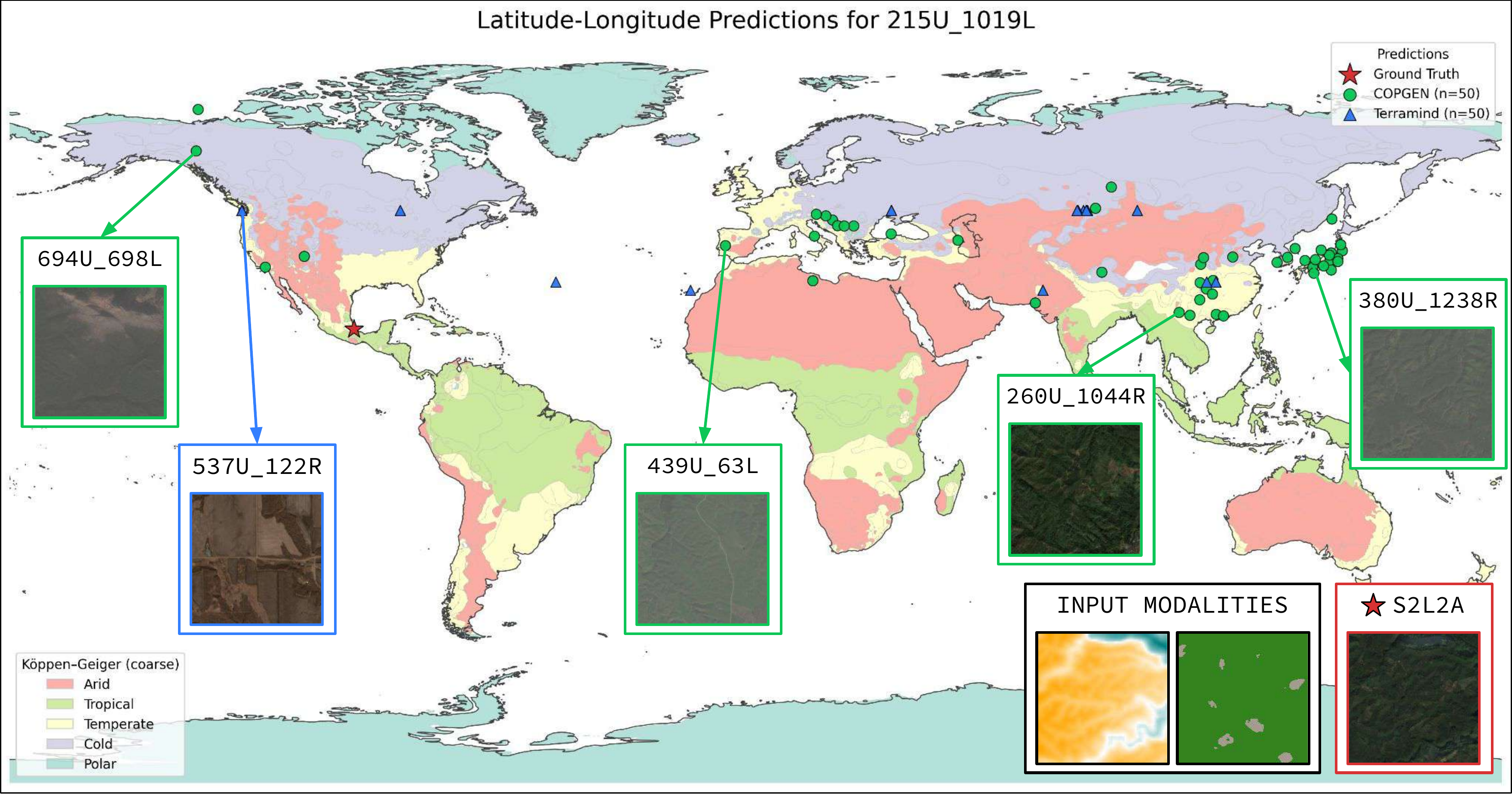}
    \caption{\textbf{Geospatial distribution analysis.} Latitude–longitude coordinates are predicted from DEM and LULC inputs over $n=50$ runs. TerraMind (\textcolor{terramindblue}{blue}) collapses to a small set of locations, whereas COP-GEN (\textcolor{copgengreen}{green}) produces a broader distribution of plausible sites sharing similar topographic and biome characteristics, reflecting the non-injective nature of the mapping. Real thumbnail visualisations of predicted locations are shown for reference.}
    \label{fig:lat-lon-215U_1019L}
\end{figure*}
}
{\begin{figure*}
    \centering
    \includegraphics[width=\linewidth]{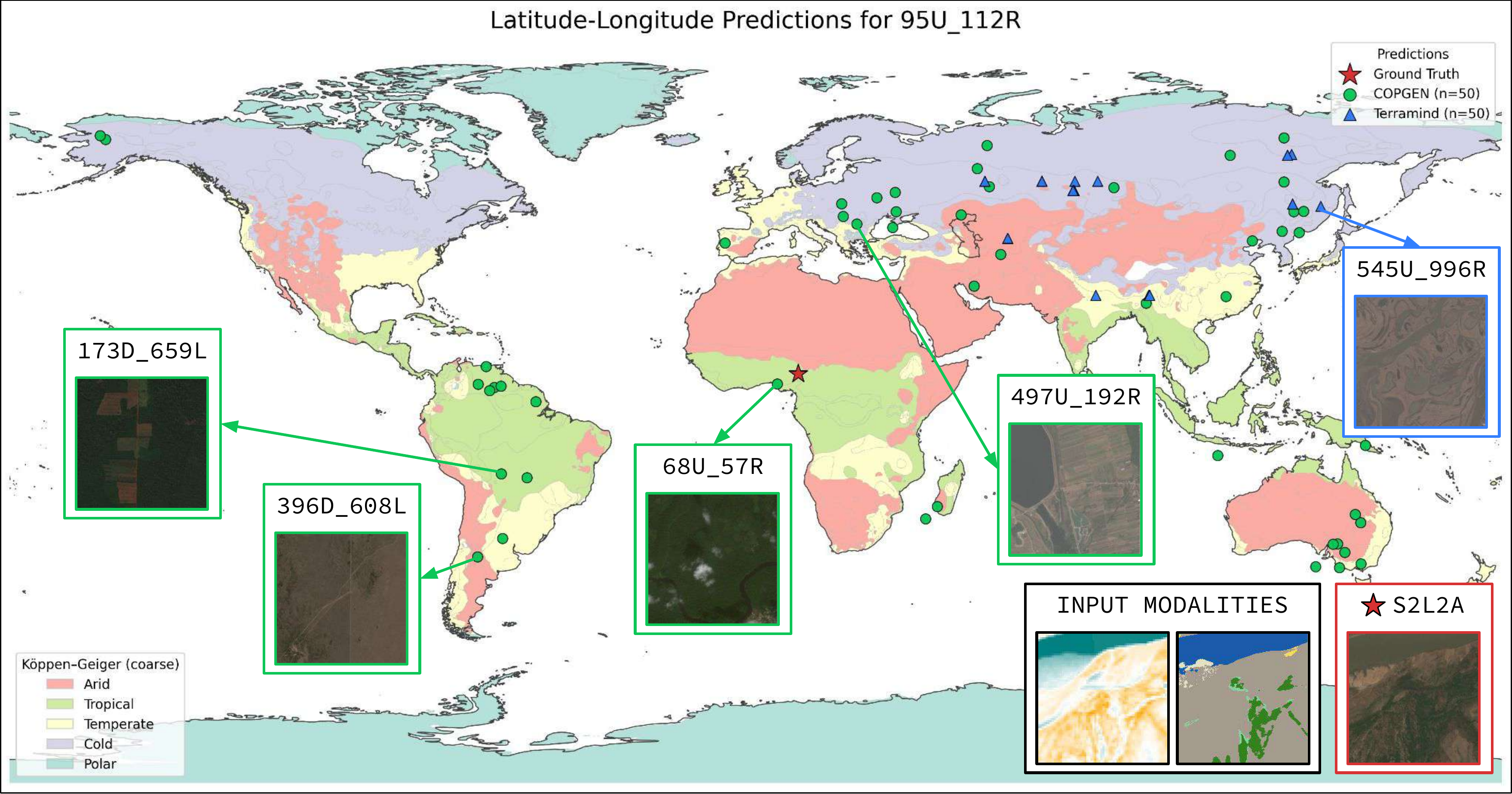}
    \caption{\textbf{Geospatial distribution analysis.} Conditional latitude–longitude predictions from DEM and LULC inputs ($n=50$) reveal that TerraMind (\textcolor{terramindblue}{blue}) concentrates on a few modes, while COP-GEN (\textcolor{copgengreen}{green}) captures multiple plausible geographic locations with similar terrain and biome properties, consistent with a non-injective mapping. Example real-image thumbnails are provided for comparison.}
    \label{fig:lat-lon-95U_112R}
\end{figure*}
}
{\begin{figure*}
    \centering
    \includegraphics[width=\linewidth]{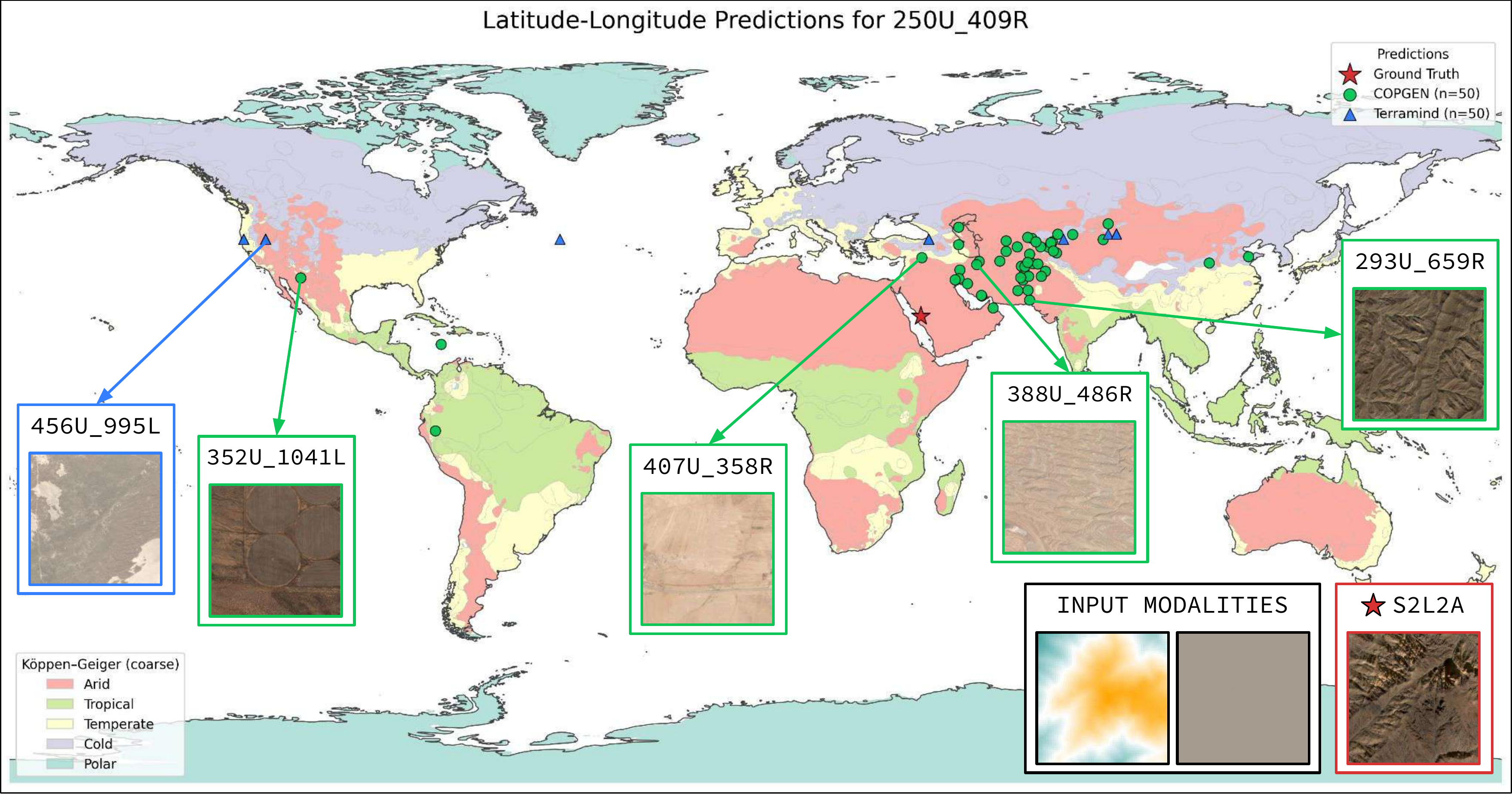}
    \caption{\textbf{Geospatial distribution analysis.} Given DEM and LULC inputs, latitude–longitude predictions over $n=50$ runs show TerraMind (\textcolor{terramindblue}{blue}) collapsing to a limited set of locations, while COP-GEN (\textcolor{copgengreen}{green}) captures a diverse distribution of plausible sites with similar topographic and biome characteristics, reflecting the non-injective nature of the task. This behavior is further supported by the fact that most COP-GEN predictions fall within the \textit{Arid} Köppen–Geiger climate class, consistent with the bare-ground input. Real location thumbnails are shown for reference.}
    \label{fig:lat-lon-250U_409R}
\end{figure*}
}
{\begin{figure*}
    \centering
    \includegraphics[width=\linewidth]{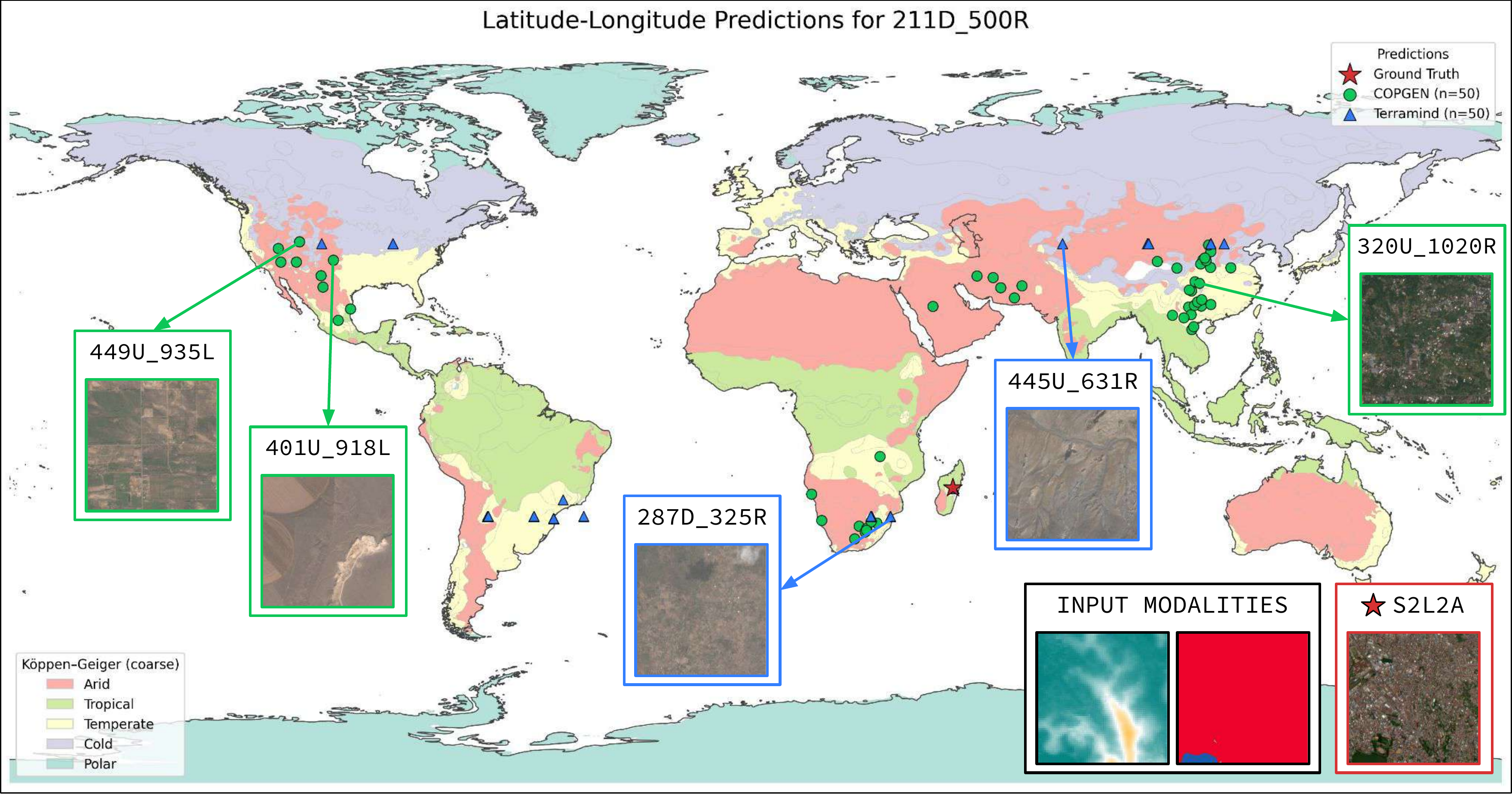}
    \caption{\textbf{Geospatial distribution analysis.} We evaluate conditional latitude–longitude prediction from DEM and LULC inputs across $n=50$ samples. TerraMind (\textcolor{terramindblue}{blue}) tends to collapse to a few specific locations, whereas COP-GEN (\textcolor{copgengreen}{green}) generates a spatially distributed set of plausible locations with similar terrain and biome attributes, consistent with a non-injective mapping. Real thumbnail visualizations of predicted locations are included.}
    \label{fig:lat-lon-211D_500R}
\end{figure*}
}

{\begin{figure*}
    \centering
    \includegraphics[width=\linewidth]{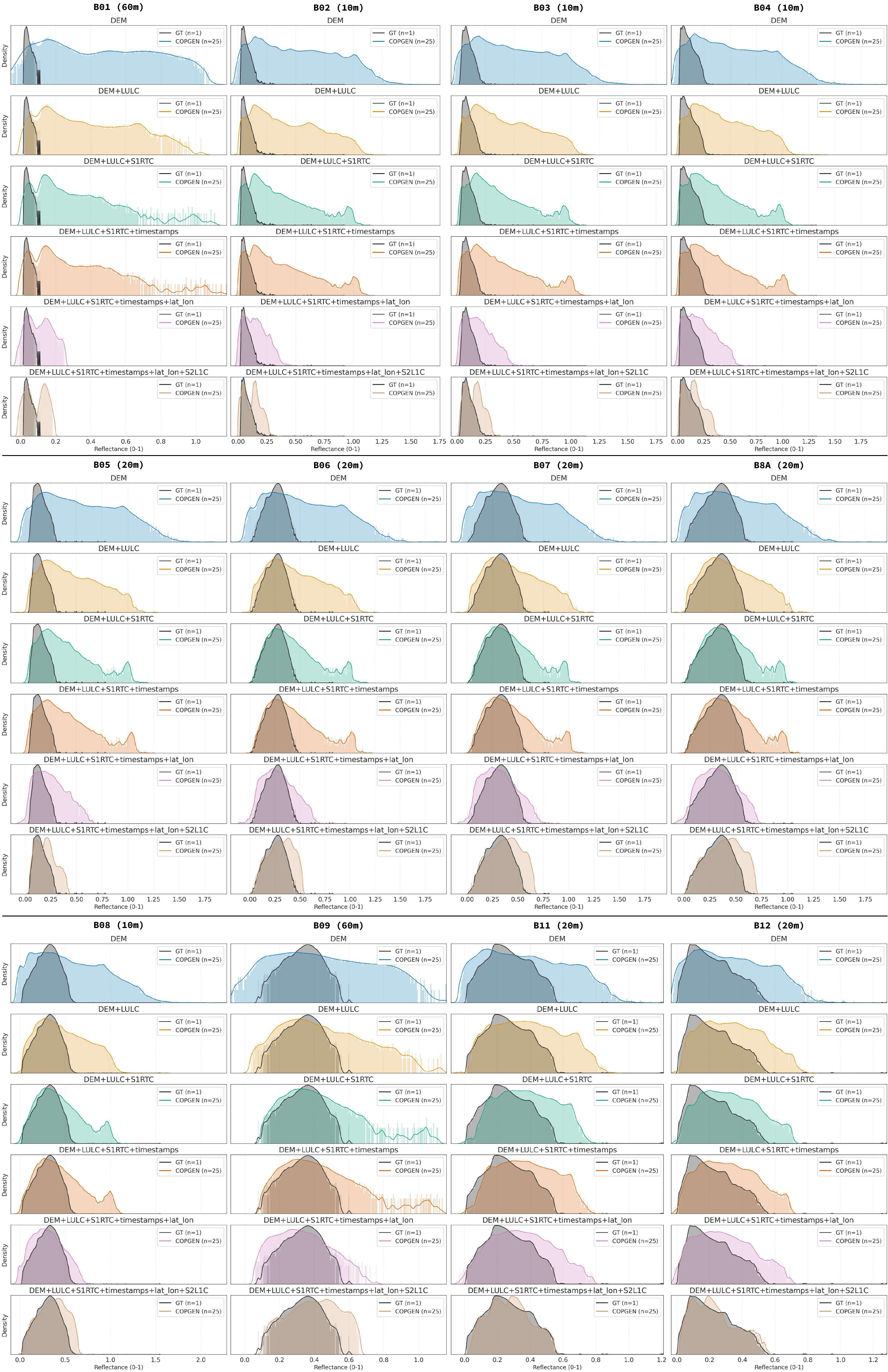}
    \caption{\textbf{Effect of input conditioning on spectral distribution spread (195D\_669L).} As additional input modalities are incorporated, generated samples increasingly align with the ground-truth (GT) distribution. For each conditioning configuration, 25 stochastic Sentinel-2 L2A (S2L2A) samples are generated, and predicted band distributions are summarized using histograms and kernel density estimates (KDEs).}
    \label{fig:195D_669L_distribution}
\end{figure*}
}
{\begin{figure*}
    \centering
    \includegraphics[width=\linewidth]{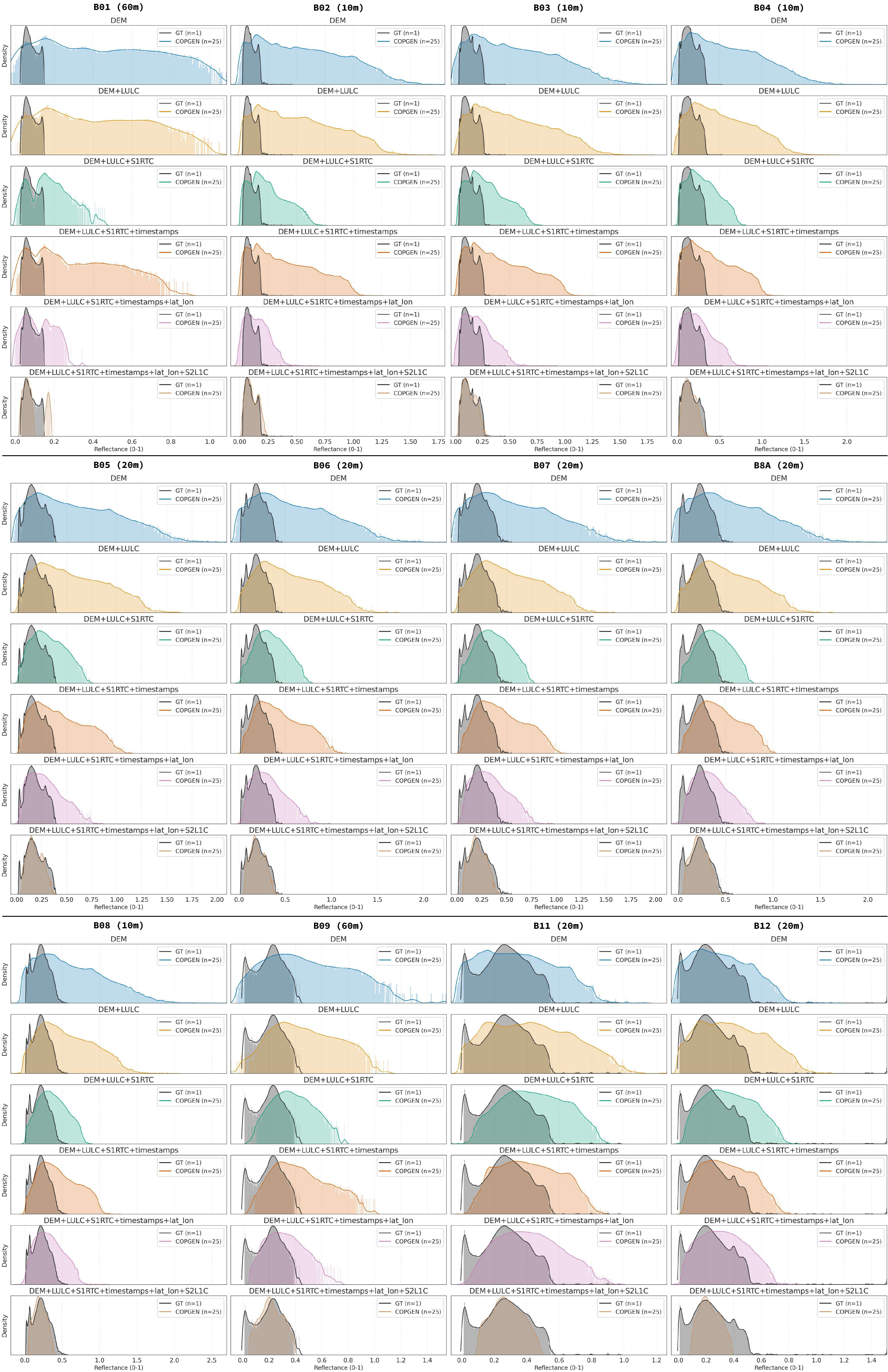}
    \caption{\textbf{Effect of input conditioning on spectral distribution spread (248U\_978R).} Increasing the number of conditioning modalities progressively constrains the generated Sentinel-2 L2A (S2L2A) outputs toward the ground-truth (GT) distribution. For each setting, 25 stochastic samples are generated and evaluated using histograms and kernel density estimates (KDEs).}
    \label{fig:248U_978R_distribution}
\end{figure*}
}
{\begin{figure*}
    \centering
    \includegraphics[width=\linewidth]{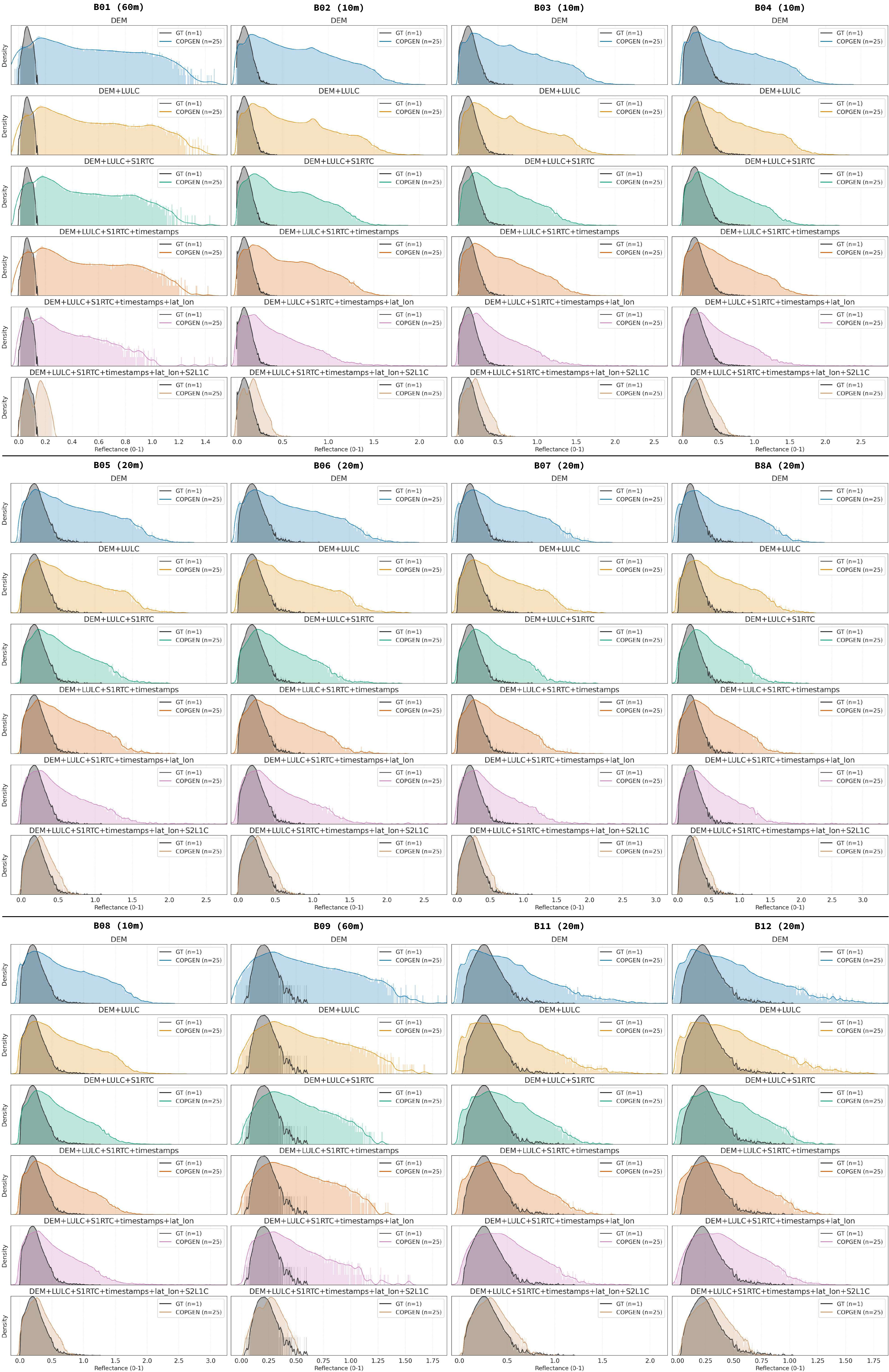}
    \caption{\textbf{Effect of input conditioning on spectral distribution spread (250U\_409R).} As additional modalities are used for conditioning, the variability of generated Sentinel-2 L2A (S2L2A) samples decreases and better matches the ground-truth (GT) distribution. For each configuration, 25 stochastic generations are summarized via histograms and kernel density estimates (KDEs).}
    \label{fig:250U_409R_distribution}
\end{figure*}
}
{\begin{figure*}
    \centering
    \includegraphics[width=\linewidth]{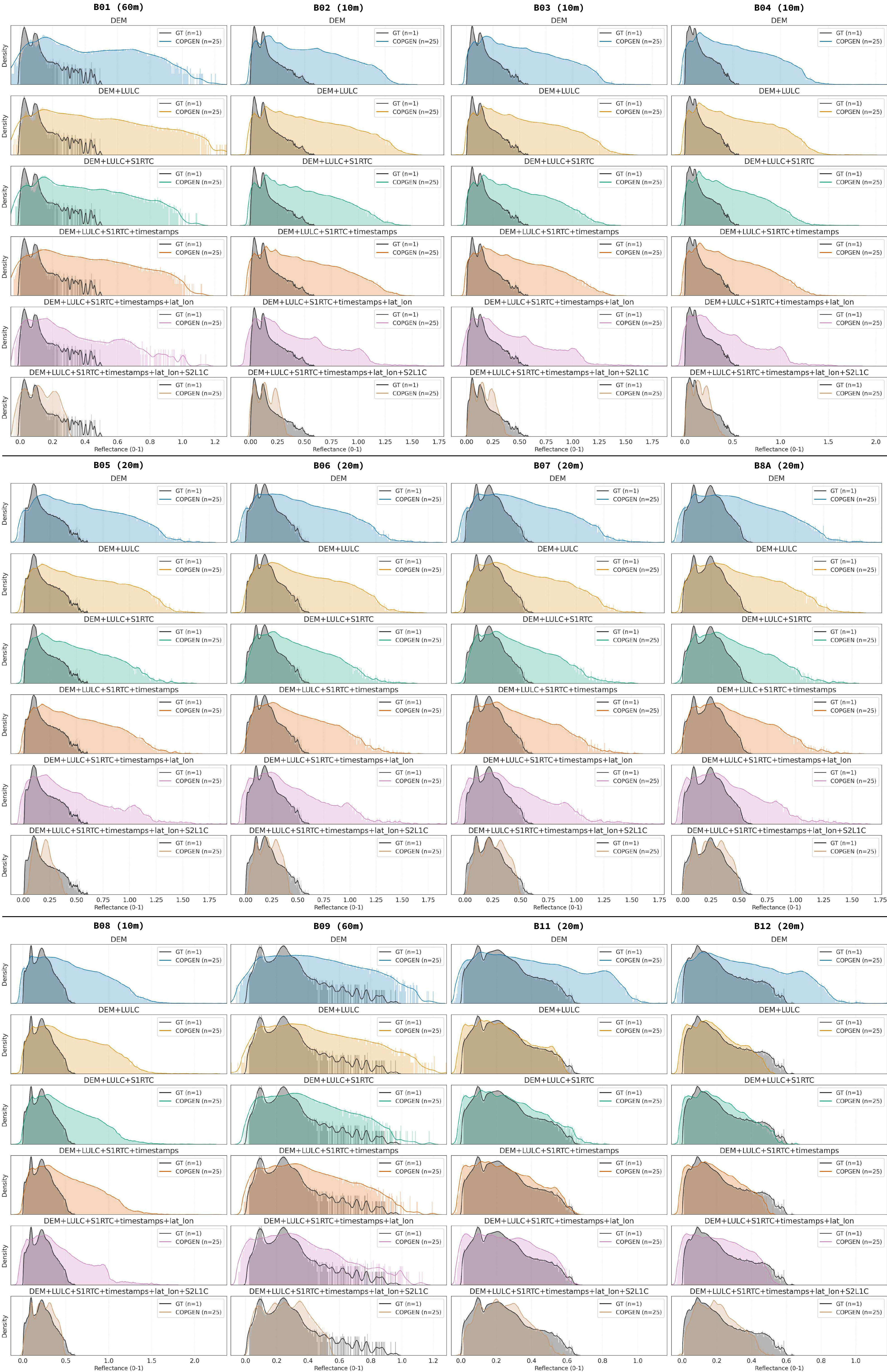}
    \caption{\textbf{Effect of input conditioning on spectral distribution spread (143D\_1481R).} Providing additional conditioning modalities narrows the distribution of generated Sentinel-2 L2A (S2L2A) samples toward the ground truth. For each setup, 25 stochastic samples are evaluated using histograms and KDEs.}
    \label{fig:143D_1481R_distribution}
\end{figure*}
}

{\begin{figure*}
    \centering
    \includegraphics[width=\linewidth]{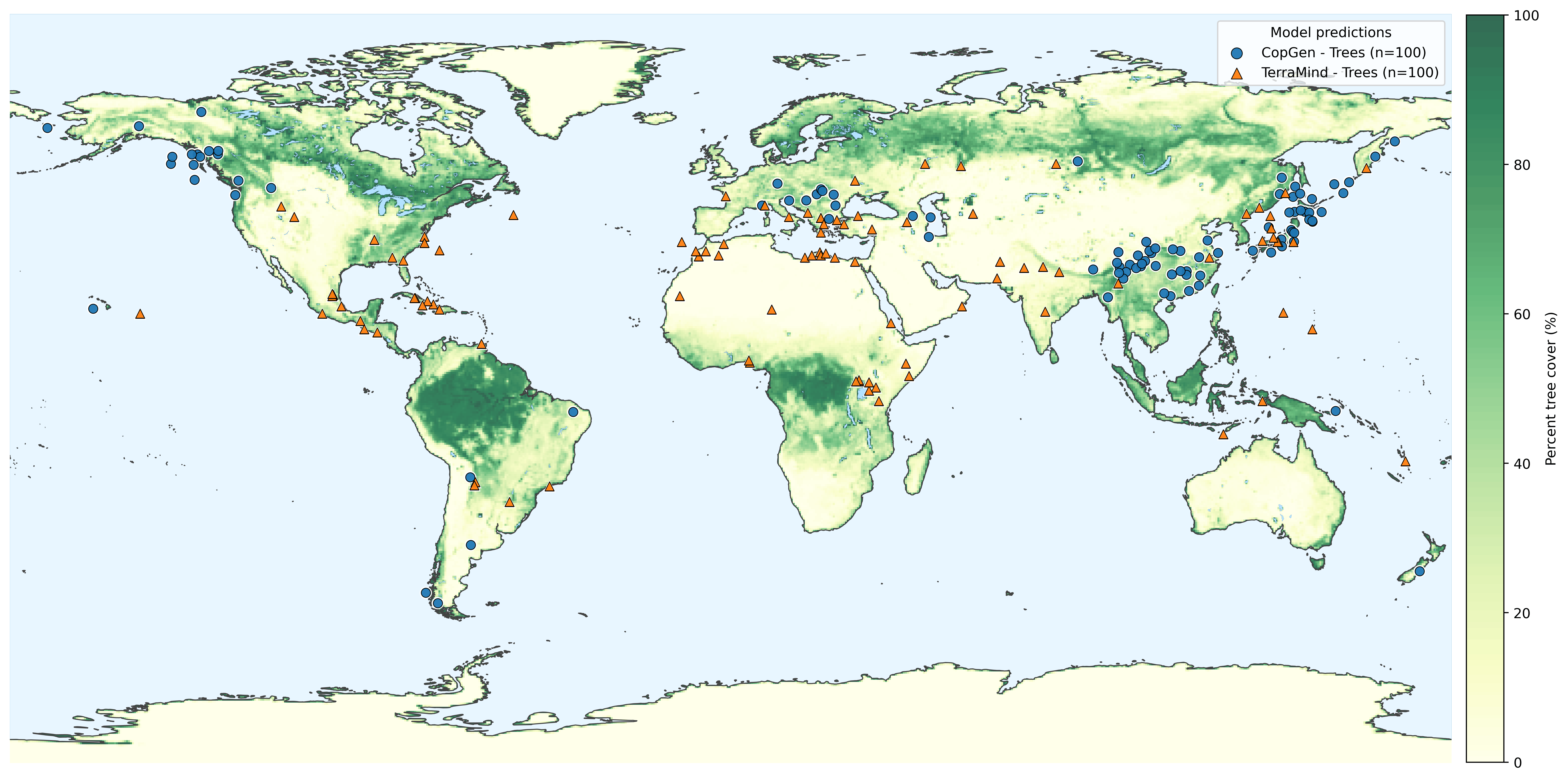}
    \caption{\textbf{Class-conditional geolocation priors for trees.} 
    COP-GEN is conditioned solely on a land-cover tile that is entirely labeled as \emph{trees}. The model generates multiple plausible latitude--longitude samples, which are plotted as points on the map. The background shows global tree cover, normalised by percentage, where darker green indicates higher tree density. Predicted locations concentrate in regions with substantial forest cover across multiple continents, indicating that COP-GEN has learned meaningful global geographic priors associated with tree-dominated landscapes, with limited dispersion into implausible regions.}
    \label{fig:lat_lon_comparison_trees}
\end{figure*}
}
{\begin{figure*}
    \centering
    \includegraphics[width=\linewidth]{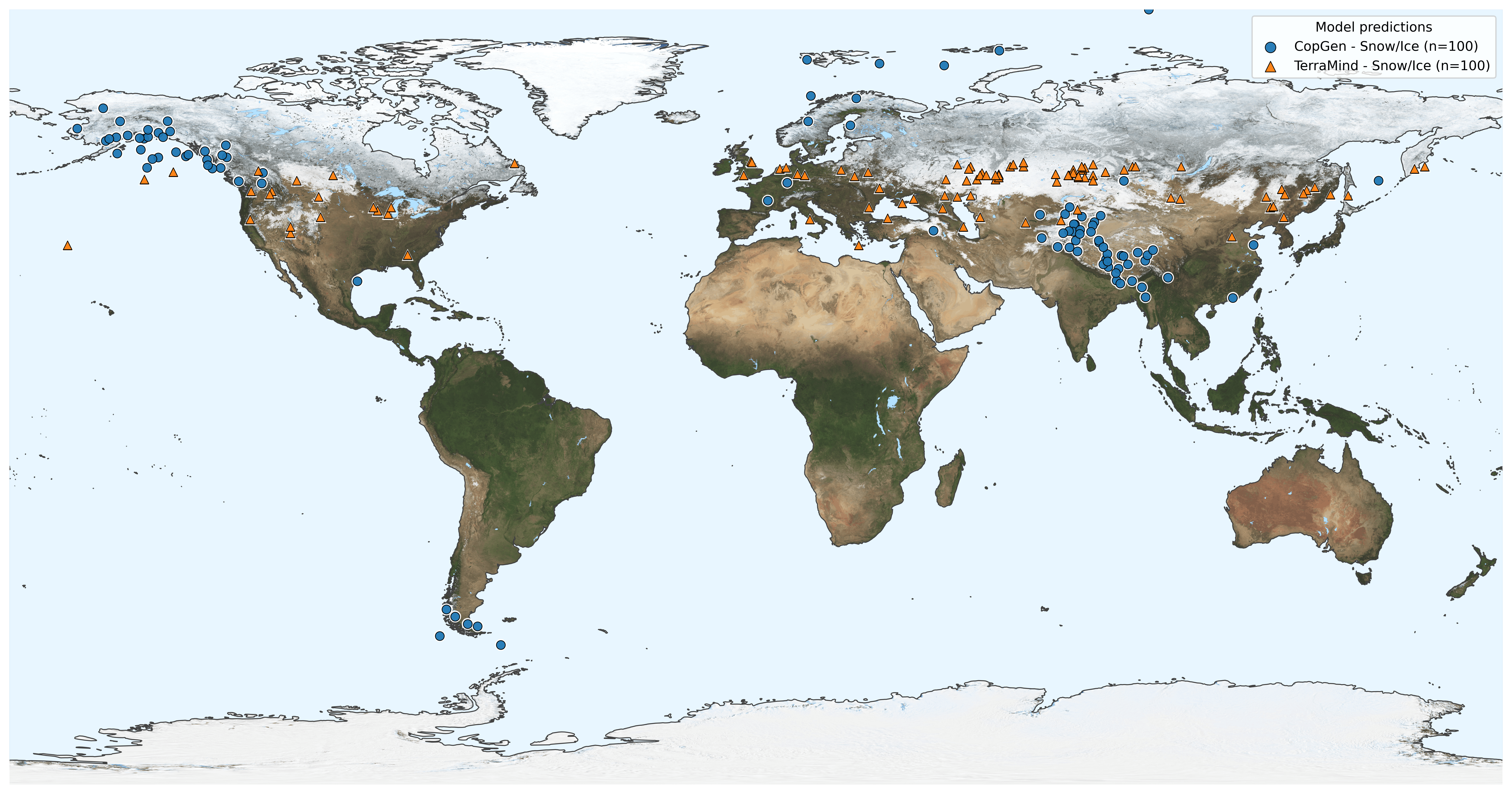}
    \caption{\textbf{Class-conditional geolocation priors for snow and ice.} 
    The model is conditioned on a tile fully assigned to the \emph{snow/ice} land-cover class and outputs multiple plausible latitude--longitude samples. Predictions are overlaid on a global mountain-range basemap to provide geographic context. The majority of samples align with high-latitude and high-elevation regions, including the Himalayas, Alaska, Patagonia, and other major mountain systems, demonstrating that COP-GEN captures realistic geographic and topographic priors associated with persistent snow and ice cover.}
    \label{fig:lat_lon_comparison_snow_ice}
\end{figure*}
}
{\begin{figure*}
    \centering
    \includegraphics[width=\linewidth]{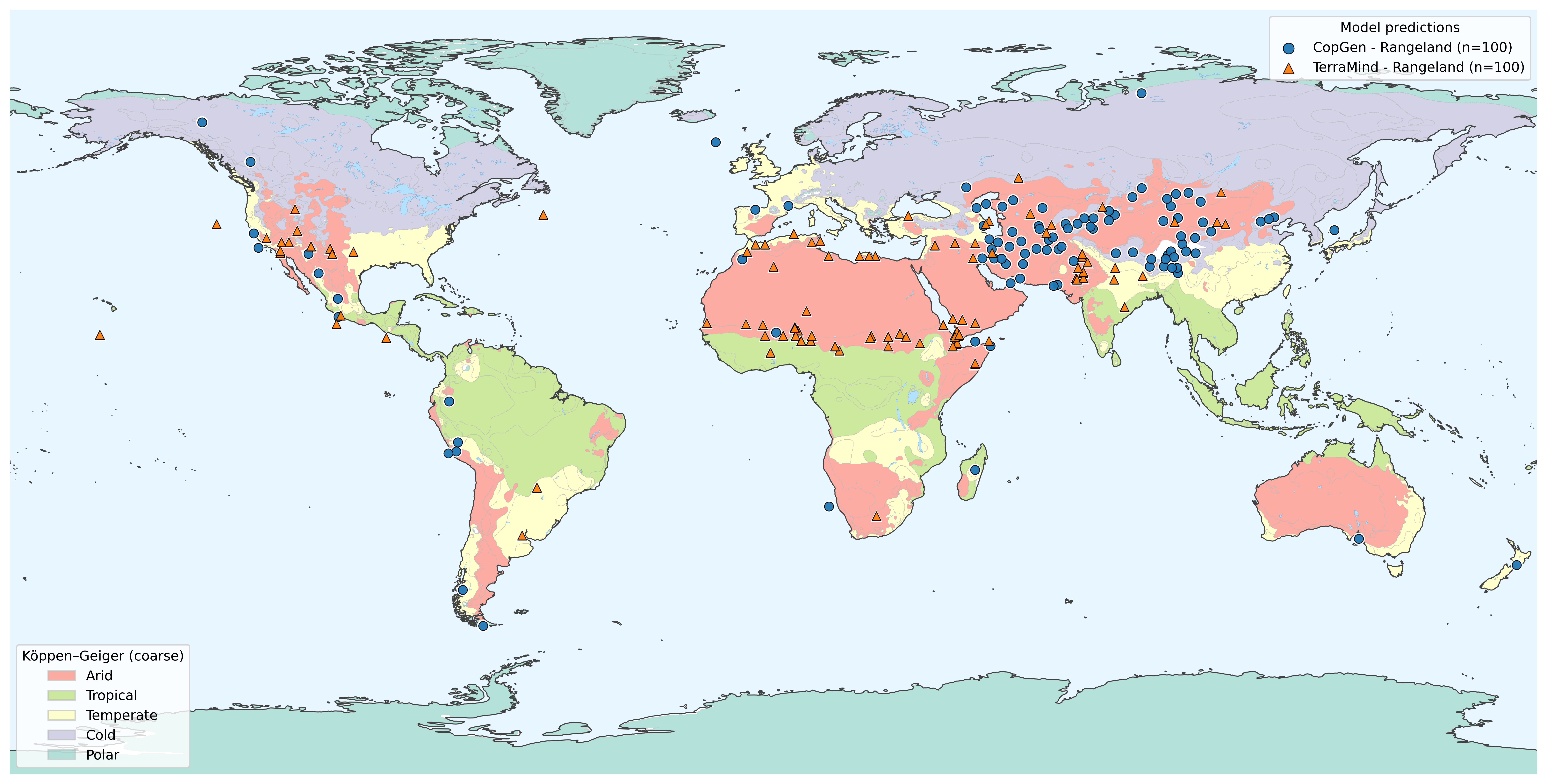}
    \caption{\textbf{Class-conditional geolocation priors for rangeland.} 
    COP-GEN is conditioned on a homogeneous \emph{rangeland} tile and generates a distribution of plausible geographic locations. The predictions are visualised on a Köppen--Geiger climate classification basemap. Generated locations predominantly fall within arid climate zone and exhibit broad global coverage, indicating that the model has learned large-scale climatic associations characteristic of rangeland ecosystems.}
    \label{fig:lat_lon_comparison_rangeland}
\end{figure*}
}
{\begin{figure*}
    \centering
    \includegraphics[width=\linewidth]{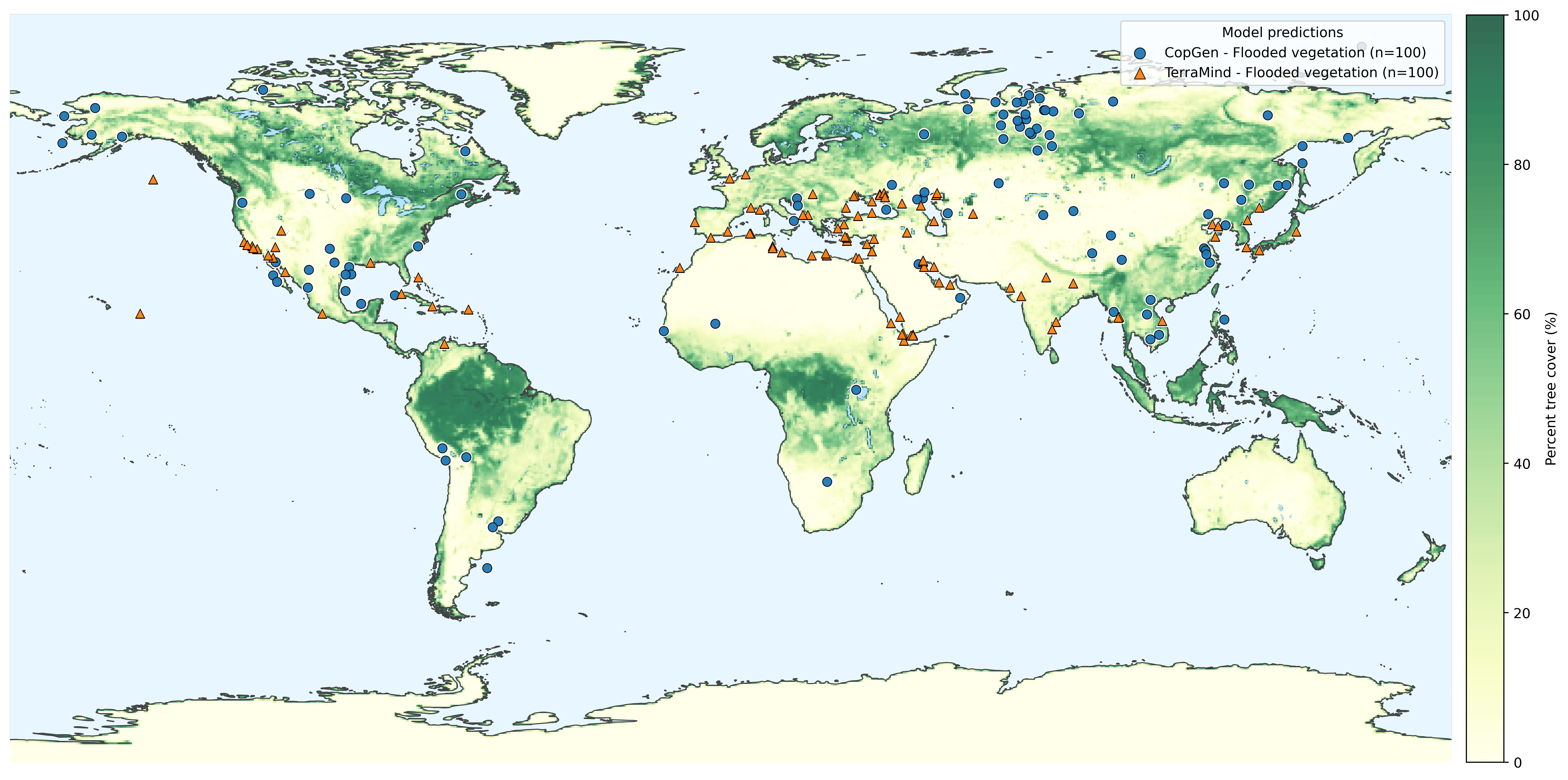}
    \caption{\textbf{Class-conditional geolocation priors for flooded vegetation.} 
    Given a tile entirely labeled as \emph{flooded vegetation}, COP-GEN outputs multiple candidate latitude--longitude samples. Predictions are plotted on a global tree cover basemap (normalised by percentage) to contextualise vegetation density. The predicted locations tend to cluster in forested and wetland-rich regions, consistent with the ecological conditions under which flooded vegetation typically occurs, while maintaining reasonable geographic diversity.}
\label{fig:lat_lon_comparison_flooded_vegetation}
\end{figure*}
}
{\begin{figure*}
    \centering
    \includegraphics[width=\linewidth]{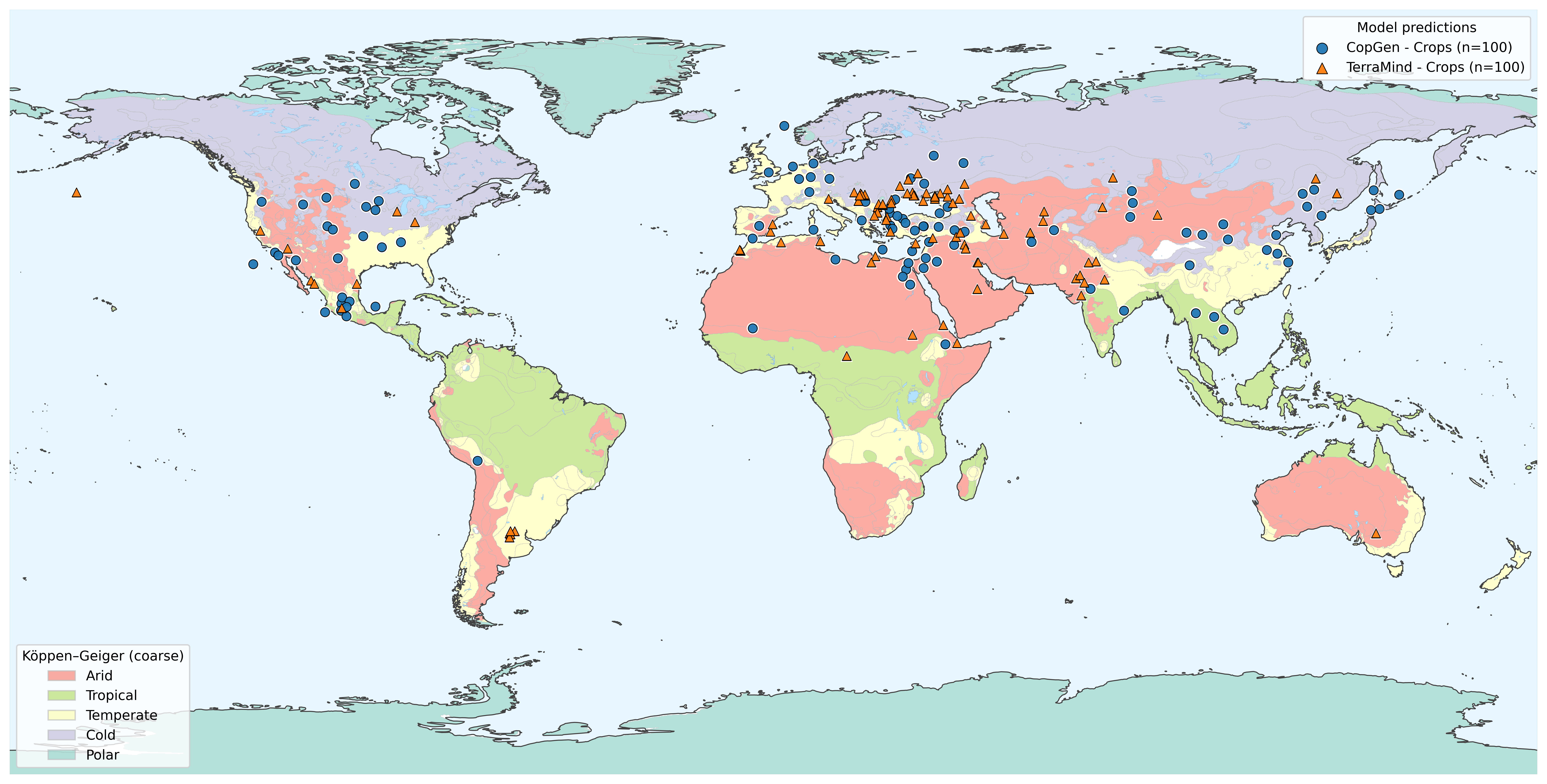}
    \caption{\textbf{Class-conditional geolocation priors for crops.} 
    The model is conditioned on a land-cover tile fully classified as \emph{crops} and produces multiple plausible geographic locations. Predictions are overlaid on a Köppen--Geiger climate basemap. Samples concentrate in temperate and continental climate zones, with notable densities over Central Europe, North America, and Central Asia. The inferred latitude distribution aligns well with known climatic requirements for large-scale agriculture, indicating that COP-GEN captures meaningful agro-climatic priors.}
    \label{fig:lat_lon_comparison_crops}
\end{figure*}
}
{\begin{figure*}
    \centering
    \includegraphics[width=\linewidth]{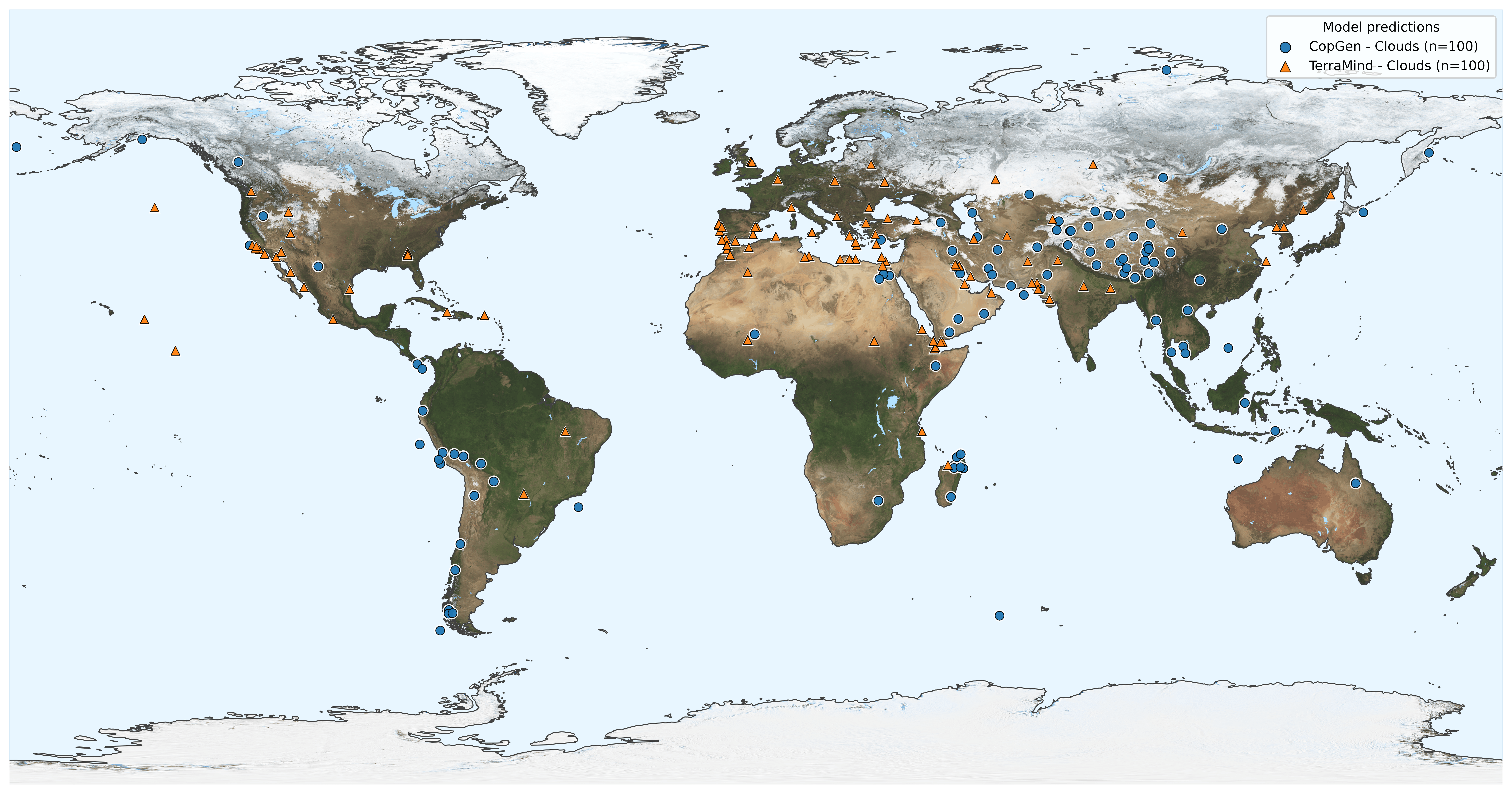}
    \caption{\textbf{Class-conditional geolocation priors for clouds.} 
    COP-GEN is conditioned on a tile entirely labeled as \emph{clouds} and generates multiple latitude--longitude predictions. The points are visualised on a global mountain-range basemap. Many samples align with major orographic regions such as the Himalayas and the Andes, where cloud formation is frequent due to topographic lifting, suggesting that the model has learned physically plausible geographic patterns associated with persistent cloud cover.}

    \label{fig:lat_lon_comparison_clouds}
\end{figure*}
}
{\begin{figure*}
    \centering
    \includegraphics[width=\linewidth]{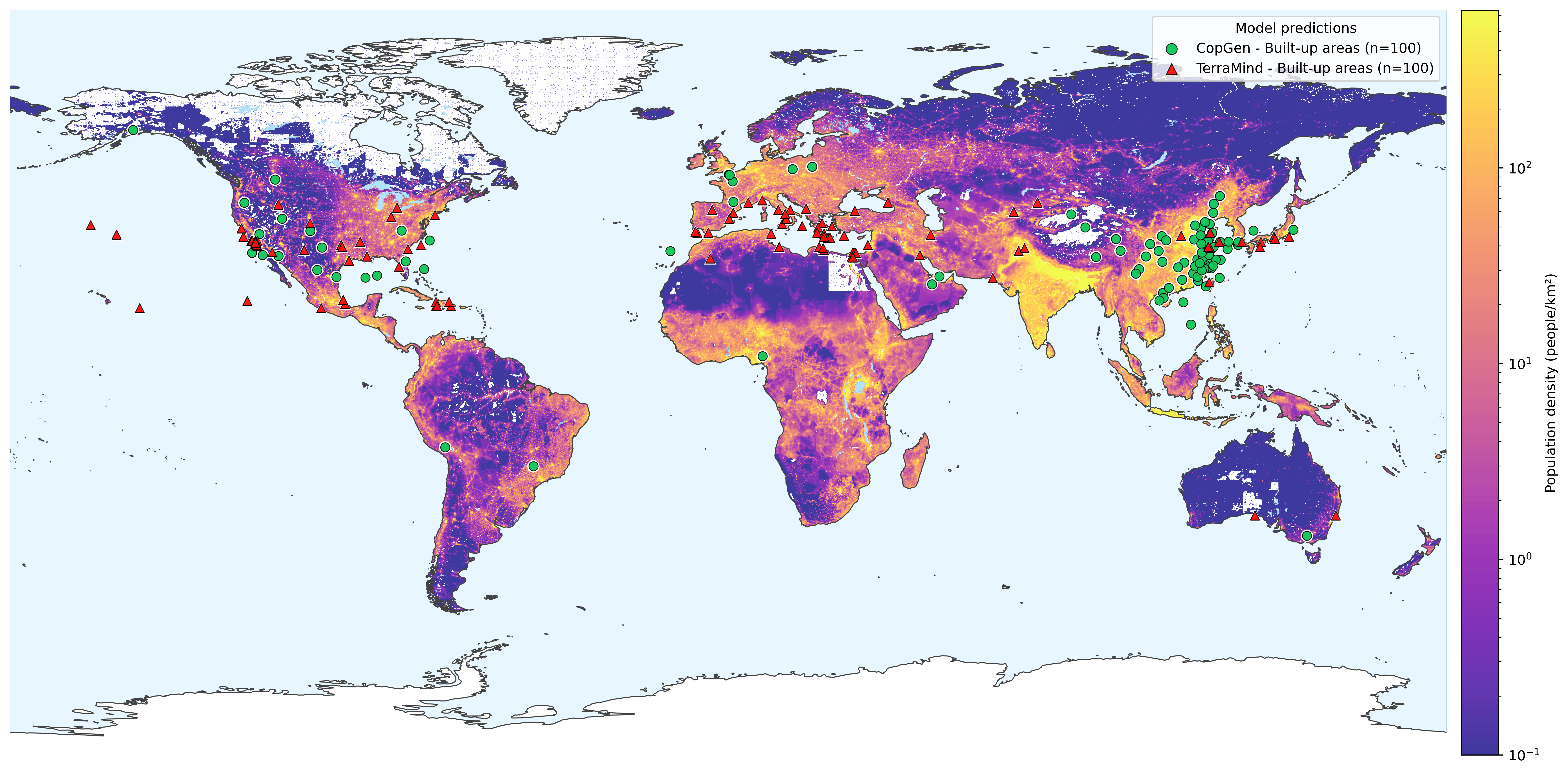}
    \caption{\textbf{Class-conditional geolocation priors for built-up areas.} 
    Conditioned on a homogeneous \emph{built-up} land-cover tile, COP-GEN generates a distribution of plausible geographic locations. Predictions are overlaid on a global population density basemap. The model places most samples in densely populated regions, particularly across Asia, with additional concentrations in Europe and North America, reflecting realistic global settlement patterns and urbanisation priors.}
    \label{fig:lat_lon_comparison_built_area}
\end{figure*}
}
{\begin{figure*}
    \centering
    \includegraphics[width=\linewidth]{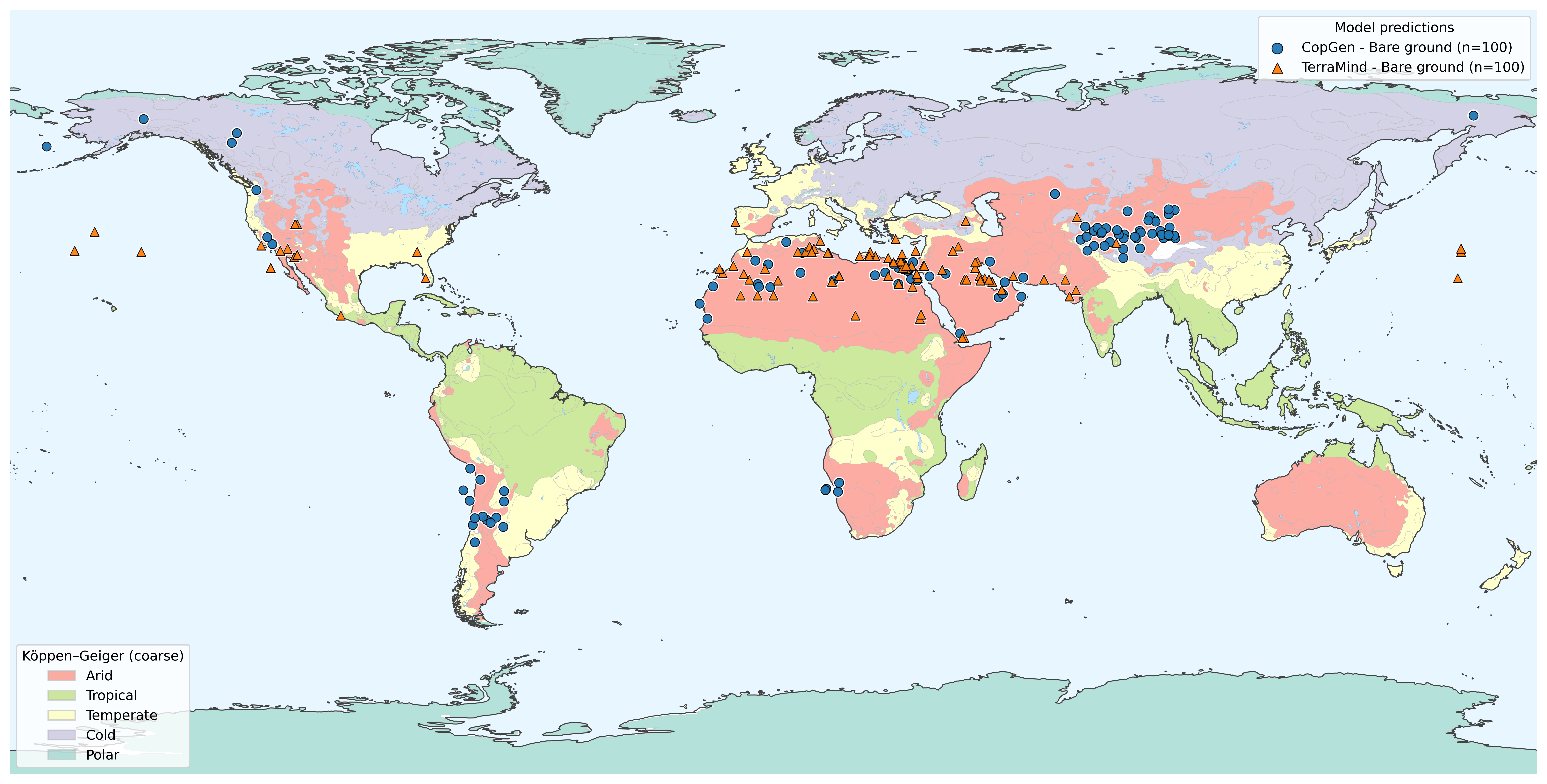}
    \caption{\textbf{Class-conditional geolocation priors for bare ground.} 
    Given a tile fully classified as \emph{bare ground}, COP-GEN outputs multiple candidate latitude--longitude locations. Predictions are shown on a Köppen--Geiger climate basemap. The majority of samples fall within arid climate zones and are broadly distributed across global desert regions, aligning well with the expected geographic distribution of bare soil and sparsely vegetated landscapes.}
    \label{fig:lat_lon_comparison_bare_ground}
\end{figure*}
}

\end{document}